\documentclass[phd,11pt]{psuthesis}
\usepackage[table, rgb]{xcolor}
\usepackage{caption}
\usepackage{subcaption}
\usepackage{tikz, pgfplots}
\usepackage{amsthm}
\usepackage{graphicx}
\usepackage{tikz}
\usetikzlibrary{arrows,decorations.pathmorphing,backgrounds,positioning,fit,petri}
\usepackage{tablefootnote, threeparttable, booktabs, multirow}
\usepackage{algorithmicx}
\usepackage[ruled]{algorithm}
\usepackage{algpseudocode}

\usepackage{cite,url,bm,hyperref}
\usepackage[thinlines]{easytable}
\usetikzlibrary{shapes,arrows}

\DeclareGraphicsExtensions{.pdf, .jpg}
\usepackage{dutchcal}
\def\x{{\mathbf x}}

\newcommand{\norm}[1]{\left\|#1\right\|}

\newcommand{\tb}{\textbf}

\def\bmt{\left[\begin{matrix}}
\def\emt{\end{matrix}\right]}

\def\imply{\Rightarrow}

\def\tb{\textbf}

\def\bx{\mathbf{x}}

\def\bb{\mathbf{b}}

\def\bd{\mathbf{d}}
\def\be{\mathbf{e}}
\def\fb{\mathbf{f}}

\def\bm{\mathbf{m}}
\def\bq{\mathbf{q}}
\def\bs{\mathbf{x}}
\def\bu{\mathbf{u}}
\def\by{\mathbf{y}}
\def\and{\text{~and~}}
\def\barN{\bar{N}}

\def\trace{\textrm{trace}}
\def\etal{\textit{et al.}}
\def\R{\mathbb{R}}

\def\bA{\mathbf{A}}
\def\bB{\mathbf{B}}
\def\bD{\mathbf{D}}
\def\bE{\mathbf{E}}
\def\Fb{\mathbf{F}}
\def\bG{\mathbold{G}}

\def\bI{\mathbf{I}}

\def\bM{\mathbf{M}}
\def\bN{\mathbf{N}}
\def\bP{\mathbf{P}}
\def\bQ{\mathbf{Q}}
\def\bU{\mathbf{U}}
\def\bW{\mathbf{W}}
\def\bX{\mathbf{X}}
\def\bY{\mathbf{Y}}
\def\bV{\mathbf{V}}
\def\bZ{\mathbf{Z}}

\def\barX{\bar{\mathbf{X}}}

\def\barX{\bar{\mathbf{X}}}

\def\bW{\mathbf{W}}
\def\bw{\mathbf{w}}

\def\bDc{\bD_{0}}
\def\bXc{\bX^{0}}
\def\bzeros{\mathbf{0}}
\def\diag{\text{diag}}

\newtheorem{theorem}{Lemma}
\makeatletter
\newsavebox\myboxA
\newsavebox\myboxB
\newlength\mylenA
\newcommand*\lbar[2][.75]{%
    \sbox{\myboxA}{$\m@th#2$}%
    \setbox\myboxB\null
    \ht\myboxB=\ht\myboxA%
    \dp\myboxB=\dp\myboxA%
    \wd\myboxB=#1\wd\myboxA
    \sbox\myboxB{$\m@th\overline{\copy\myboxB}$}
    \setlength\mylenA{\the\wd\myboxA}
    \addtolength\mylenA{-\the\wd\myboxB}%
    \ifdim\wd\myboxB<\wd\myboxA%
       \rlap{\hskip 0.5\mylenA\usebox\myboxB}{\usebox\myboxA}%
    \else
        \hskip -0.3\mylenA\rlap{\usebox\myboxA}{\hskip 0.3\mylenA\usebox\myboxB}%
    \fi}
\makeatother

\def\lbD{\lbar{\bD}}
\def\lbY{\lbar{\bY}}
\def\lbX{\lbar{\bX}}

\usepackage[mathscr]{eucal}
\DeclareMathAlphabet\mathbfcal{OMS}{cmsy}{b}{n}

\usepackage[english]{babel}

\usepackage{amsmath}
\usepackage{amssymb}
\usepackage{amsthm}
\usepackage{exscale}
\usepackage[mathscr]{eucal}
\usepackage{bm}
\usepackage{eqlist} 

\usepackage{caption}

\usepackage{epsf,psfig}
\usepackage{epsfig}
\usepackage{latexsym}
\usepackage{slashbox}
\usepackage{bbding}
\usepackage[Lenny]{fncychap}
\ChTitleVar{\Huge\sffamily\bfseries}

\usepackage{hyperref}
\hypersetup{%
    pdfborder = {0 0 0}
}
\title{Signal classification under \\structured sparsity constraints}

\author{Tiep Huu Vu}
\dept{Electrical Engineering}
\degreedate{May 2019}

\honorsdegreeinfo{for a baccalaureate degree \\ in Engineering Science \\ with honors in Engineering Science}

\documenttype{Dissertation}

\submittedto{The Graduate School}

\numberofreaders{4}

\honorsadviser{Honors P. Adviser}

\secondthesissupervisor{Second T. Supervisor}

\honorsdepthead{Department Q. Head}

\advisor[Dissertation Advisor, Chair of Committee]
        {Vishal Monga}
        {Associate Professor of Electrical Engineering}

\readerone[]
          {William E. Higgins}
          {Distinguished Professor of Electrical Engineering}

\readertwo[]
          {Kenneth Jenkins}
          {Professor of Electrical Engineering}

\readerthree[]
            {Robert T. Collins}
            {Associate Professor of Computer Science and Engineering}

\readerfour[]
          {Kultegin Aydin}
          {Professor of Electrical Engineering and Department Head}

\begin{document}
\frontmatter

%


\psutitlepage

\psucommitteepage

%
\pagestyle{plain}
\chapter*{Abstract}
    \begin{singlespace}

Object Classification is a key direction of research in signal and image
processing, computer vision and artificial intelligence. The goal is to come up
with algorithms that automatically analyze images and put them in predefined
categories. This dissertation focuses on the theory and application of sparse
signal processing and learning algorithms for image processing and computer
vision, especially object classification problems. A key emphasis of this work
is to formulate novel optimization problems for learning dictionary and
structured sparse representations. Tractable solutions are proposed subsequently
for the corresponding optimization problems.

Sparse signal processing has had remarkable recent success in problems such as
classification, object detection, image and audio recognition, image
super-resolution, etc. It essentially attempts to represent any signal (e.g.
image, video, audio, etc.) using only a few number of observations or features
from the same physical phenomenon. Based on this theory, a sparse
representation-based classifier (SRC)~\cite{Wright2009SRC} was initially
developed for robust face recognition, and thereafter was extended to several
other signal classification problems. It has been shown that SRC can be further
improved by learning a dictionary from the training samples instead of using all
of them as a dictionary. In the first part of the dissertation, we develop a
discriminative dictionary learning framework for histopathological image
analysis. Particularly, we propose an automatic feature discovery framework via
learning class-specific dictionaries and present a low-complexity method for
classification and disease grading in histopathology. Essentially, our
Discriminative Feature-oriented Dictionary Learning (DFDL)~\cite{vu2016tmi}
method learns class-specific dictionaries such that under a sparsity constraint,
the learned dictionaries allow representing a new image sample parsimoniously
via the dictionary corresponding to the class identity of the sample. At the
same time, the dictionary is designed to be poorly capable of representing
samples from other classes. 
\newpage

The next part of this dissertation exploits the observation {that although
different objects possess distinct characteristics}, they also usually share
common patterns. A recently proposed dictionary learning framework has shown the
benefit of separating the particularity and the commonality
(COPAR)~\cite{kong2012dictionary}. Inspired by this, we propose a novel method
to explicitly and simultaneously learn a set of common patterns as well as
class-specific features for classification with more intuitive constraints. Our
dictionary learning framework is hence characterized by both a shared dictionary
and particular (class-specific) dictionaries. The shared dictionary is
constrained to have low-rank, i.e. its spanning subspace should have low
dimension and the coefficients corresponding to this dictionary should be
similar. For the particular dictionaries, we impose on them the well-known
constraints stated in the Fisher discrimination dictionary learning
(FDDL)~\cite{yang2014sparse}. Further, new fast and accurate algorithms are
developed to solve the subproblems in the learning step, accelerating its
convergence. The said algorithms could also be applied to FDDL and its
extensions. The efficiency of these algorithms is theoretically and
experimentally verified by quantifying their complexity and running time with
other well-known dictionary learning methods. The work has culminated into a
dictionary learning toolbox called DICTOL~\cite{vu2016fast}
(\url{https://github.com/tiepvupsu/DICTOL}). The toolbox (in Matlab and Python)
implements numerous sparse coding algorithms as well as widely used generative
and discriminative dictionary learning methods. Many classical methods are sped
up via new numerical optimization innovations.

{We also extend sparsity} models to tensor sparsity models which significantly
enhance classification accuracy for signals with multi-channels and multi-views.
In this part, we present three novel sparsity-driven techniques, which not only
exploit the subtle features of raw captured data but also take advantage of the
polarization diversity and the aspect angle dependence information from
multi-channel signals. First, the traditional SRC is generalized to exploit
shared information of classes and various sparsity structures of tensor
coefficients for multi-channel data. Corresponding tensor dictionary learning
models are consequently proposed to enhance classification accuracy. Lastly, a
new tensor sparsity model is proposed to model responses from multiple
consecutive looks of objects, which is a unique characteristic of the dataset.



An important goal of this dissertation is to demonstrate the wide applications
of these algorithmic tools for real-world applications. To that end, we explored
important problems in the areas of: 
\begin{enumerate}
    \item Medical imaging: histopathological images acquired from mammalian
    tissues, human breast tissues, and human brain tissues.

    \item Low-frequency (UHF to L-band) ultra-wideband (UWB) synthetic aperture
    radar: detecting bombs and mines buried under rough surfaces.
    \item General object classification: face, flowers, objects, dogs, indoor scenes, etc. 
\end{enumerate}

    \end{singlespace}
\newpage

\thesistableofcontents

\thesislistoffigures

\thesislistoftables


%
\chapter{Acknowledgments}
\begin{singlespace}
I am grateful to my parents, who provided me through moral and emotional support in my life. Without their support I would not be here. I am also grateful to my siblings, especially my brother, Long Vu, who have supported me along the way since my undergraduate at Hanoi. 

With a special gratitude to my advisor, Dr. Vishal Monga, who always help me through my PhD. It was fantastic to have the opportunity to work with him. 

And finally, last but not least, also to everyone in the iPAL. Especially, Hojjat, Tiantong and John whom I worked closely with and enjoyed every moment of it. It was great sharing laboratory with all of you during my last five years. 
Thanks for all your encouragement!
\end{singlespace}


\thesismainmatter

\allowdisplaybreaks{
%

\chapter{Introduction}
\label{chapter:introduction}
\section{Motivation} 
\label{sec:motivation}

\begin{figure*}[t]
\centering
  \includegraphics[width=.9\textwidth]{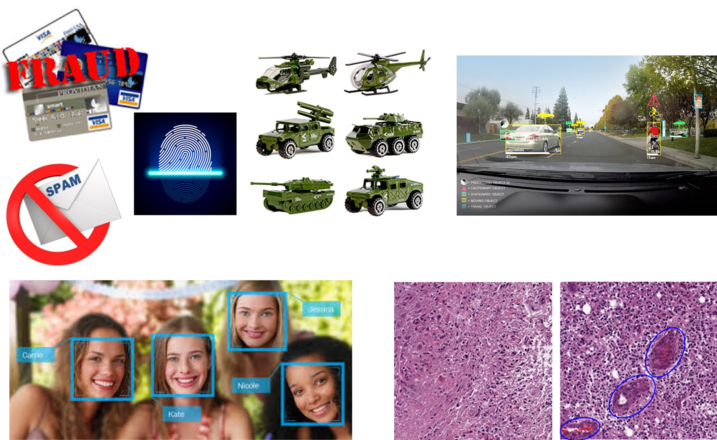}
  \vspace{-0.1in}
 \caption{\small Real-world classification problems}
  \label{fig:classificationproblems}
\end{figure*}

Object classification is one of the most important problems in machine
intelligence systems. This is a well-studied problem in many important domains
of data processing today. Typically, we are given images belonging to two or
more classes, and the challenge is to develop an effective procedure of
determining identity of a new sample. Application of object classification
ranges broadly from spam email detection, fraud transaction detection to
fingerprint verification, cancer detection, face identification, military
vehicle recognition, self-driving cars, etc (see
Fig.~\ref{fig:classificationproblems}).

\par 

In general, a classification system comprises of two main components: a feature
extraction tool and a classifier. The \textit{feature extraction} part plays a
role in generating discriminative characteristics of each class. Traditional
techniques in extracting features are hand-designed, such as Haar wavelets,
Difference of Gaussians (DOG) filter, Gabor filters, histogram of oriented
gradient (HOG) descriptors, SIFT descriptors\cite{lowe1999object}, spatial
pyramid matching\cite{lazebnik2006beyond} and many others. In some particular
cases, the feature extraction also includes data dimension reduction step, which
reduces the complexity of the classifier and hence, maximizes speed of the
system. The \textit{classifier}, which is often a result of a learning scheme,
takes the extracted features as inputs and predicts identity of signals. Several
classifier design techniques have been applied in practical applications, such
as Naive Bayes \cite{lewis1998naive}, Decision Trees
\cite{quinlan1986induction}, Logistic Regression \cite{hosmer2004applied},
Neural Networks \cite{hornik1989multilayer}, Support Vector Machines
\cite{hearst1998support}, and Deep Learning \cite{hinton2006fast,raina2007self}
recently. 

One classifier can be seen as a function $f$ that takes an input signal $\bx$
(discrete or continuous) and generates an discrete output $y \in \mathcal{C}$
representing
category/class of $\bx$. Based on several training pairs $\{(\bx_1, y_1),
(\bx_2, y_2), \dots, (\bx_N, y_N)\}$, a classification algorithm tries to
`learn' the mapping $f$ such that $y_n \approx f(\bx_n)$ for as many as possible
$n \in \{1, 2, \dots, N\}$. A good classifier is one that not only could well
capture relationship of training pairs, i.e. what it has seen, but also need to
well represent new samples. This property, called \textit{generalization}, is
one of the most important factors in a classification system. However,
attaining this property is usually a challenging task, as real-world
classification problems exhibit difficulties. \textit{In particular, this dissertation emphasizes on addressing the following challenges:}

\begin{itemize}
    \item Many practical problems encounter the  issue of \textit{insufficient
training}. From probability perspective, the identity of a signal can be
formulated as a maximum likelihood estimation problem, i.e. estimating $p
(y|\bx)$ for each possible category $y \in \mathcal{C}$. Class decisions are
made using empirical estimates of the true densities learned from available
training data. Without prior knowledge of the (limited) training data, a system
can severely \textit{misestimate} the density, resulting in a poor classifier
with lack of \textit{generalization} property. {This problem becomes more severe
when data dimensionality increases}. In this case, the volume of the space
drastically increases. In order to achieve a stable system, the amount of data
required to support the result often grows exponentially with the
dimensionality. Unfortunately, this \textit{high-dimensional data} phenomenon is
commonly encountered in signal classification problems. For instance, a small
gray image of size $200\times 200$ has dimension of 40000. Particular images,
such as hyperspectral or medical, have even more dimensions with million of
pixels per channel and multiple channels per image. The problem persists even
when low-dimensional image features are considered instead of entire images.
\textit{Training insufficiency} and \textit{high dimensionality} are thus
important concerns in signal classification.





\item Another challenge is that in practical problems, generous training is
often not
only limited but also taken under \textit{presence of noise or occlusion}. This
might be caused by limitations of devices, e.g. optical sensors, microscopes, or
radar transmitter/receiver, that capture signals. Some unexpected sources of
noise or distortions are often incorporated into the classification frameworks.
In a face identification system, training faces might be captured under a good
light conditions and/or well-aligned; a test face image, however, is usually
captured under bad conditions with low light, occlusion by other objects, or
at different angles. In a radar system, captured signals are commonly heavily
interfered by other signals in the field.

\item Additionally, classification systems often face the problem of 
\textit{high within-class variance} and \textit{small between-class variance}.
In a face identity problem, an image could include the face of the same
individual wearing sunglasses, with hat, or at different angles. In a flower
classification problem, a flower may be captured with different background, at
different times of a day, or at different stages of the blooming process. This
variability of samples from the same class is also called \textit{high
within-class variance}. Another challenge could happen in a classification
problem is that different classes could share common patterns. Objects of
different classes might be captured under the same background conditions. In
histopathological images, an image in \textit{disease} class could contain a
large portion of \textit{healthy} nuclei. This \textit{small between-class
variance} issue is another aspect we should consider.

\end{itemize}

Sparse representations have emerged as a powerful tool for a range of signal
processing applications. Applications include {compressed sensing
\cite{donoho2006compressed}, signal denoising, sparse signal recovery
\cite{mousavi2015iterative}, image inpainting \cite{Aharon2006KSVD}, image
segmentation \cite{spratling2013image}}, and more recently, signal
classification. In such representations, most of signals can be expressed by a
linear combination of few bases taken from a ``dictionary''. Based on this
theory, a sparse representation-based classifier (SRC) \cite{Wright2009SRC} was
initially developed for robust face recognition, and thereafter adapted to
numerous signal/image classification problems, ranging from medical image
classification \cite{vu2015dfdl,vu2016tmi,Srinivas2014SHIRC}, hyperspectral
image classification \cite{sun2015task,sun2014structured,chen2013hyperspectral},
synthetic aperture radar (SAR) image classification \cite{zhang2012multi},
recaptured image recognition \cite{thongkamwitoon2015image}, video anomaly
detection \cite{mo2014adaptive}, and several others~\cite{zhang2012joint,
dao2014structured, dao2016collaborative, van2013design, yang2010metaface,
vu2016amp}.

The success of SRC-related methods firstly comes from the fact that sparsity
models are often robust to noise and occlusions~\cite{Wright2009SRC}. In
addition, the insufficient training problem can be mitigated by incorporating
prior knowledge of signals into the optimization problem during the inference
process as regularization terms or sparsity constraints that capture signal
relationships~\cite{zhang2012joint, zhou2014jointly, Srinivas2014SHIRC}. 
It has been shown that learning a dictionary from the training samples instead
of using all of them as a dictionary can further enhance the performance of SRC.

\textit{This dissertation proposes novel structured sparsity frameworks for
different classification applications}. These applications include but are not
limited to disease diagnosis and cancer detection, ultra-wide band synthetic
aperture radar signal classification, face identification, flower
classification, and general object classification. The works in this
dissertation have culminated into one software and two toolboxes which are
widely used by peer researchers. 

The next section of this chapter provides overview of sparse
representation-based classification and fundamental dictionary learning methods.
The last section will present contributions and organization of this
dissertation.

\section{Sparse Representation-based Classification} 
\label{sec:sparse_representation}
\subsection{Sparse Representation-based Classification} 
\label{sub:sparse_representation_based_classification}

A significant contribution to the development of algorithms for image
classification that addressed some of aforementioned challenges up to some
extent is a recent sparse representation-based classification (SRC) framework
\cite{Wright2009SRC}, which exploits the discriminative capability of sparse
representation. Given a sufficiently diverse collection of training images from
each class, any image from a specific class can be approximately represented as
a linear combination of training images from the same class. Therefore, if we
have training images of all classes and form a basis or dictionary based on
that, any new and unseen test image has a sparse representation with respect to
such over-complete dictionary. It is worth to mention that sparsity assumption
holds due to the class-specific design of dictionaries as well as the assumption
of the linear representation model.

\par
Concretely, given a collection $\bD \in \R^{d\times k_i}$ of objects from one
class, in which each column of $\bD$ is the vectorized version of one signal
(image in this case), a new signal $\by \in \R^{d}$ from the same class can be
approximately expressed as $\by \approx \bD\bs$. In addition, this assumption is
not true when $\bD$ comprising images coming from other class. Now suppose that
we have $C$ classes of subject, and let $\bD = [\bD_1, \bD_2, \dots, \bD_C]$ be
the set of original training samples, where $\bD_i$ is the sub-set of training
samples from class $i$. Denote by $\by$ a testing sample. The procedures of SRC
are as follows:
 \begin{enumerate}
   \item Sparsely code $\by$ on $\bD$ via $l_1$-norm minimization:
   \begin{equation}
       \hat{\bs} = \arg\min_{\bs} \{\|\by - \bD\bs\| + \lambda\|\bs\|_1\}
       \label{eqn: src}
   \end{equation}
   where $\lambda$ is a scalar constant.
   \item Do classification via:
   \begin{equation}
       \text{identity}(\by) = \arg\min_{i}\{e_i\}
   \end{equation}
   where $e_i = \|\by - \bD_i\delta_i{(\hat{\bs})}\|$ and $\delta_i(\hat{\bs})$ is the part of $\hat{\bs}$ associated with class $i$.   
 \end{enumerate}

Although SRC scheme shows interesting results, the dictionary used in it may not
be effective enough to represent the query images due to the uncertain and noisy
information in the original training images. The coding complexity increases as
more training data {is} involved in building the dictionary. In addition, using the
original training samples as the dictionary could not fully exploit the
discriminative information hidden in the training samples. On the other hand,
using analytically designed off-the-shelf bases as dictionary (e.g.
\cite{huang2006sparse} uses Haar wavelets and Gabor wavelets as the dictionary)
might be universal to all types of images but will not be effective enough for
specific type of images such as face, digit and texture images. In fact, all the
above mentioned problems of predefined dictionary can be addressed, at least to
some extent, by learning properly a dictionary from the original training
samples. The next subsections will review the Online Dictionary
Learning Method for compact purpose and two well-known Discriminative Dictionary
Learning methods used for classification purpose.
\subsection{Online Dictionary Learning} 
\label{sub:online_dictionary_learning_method}

{When number of training images increases}, concatenating all of them
into one ``fat'' matrix $\bD$ and then solving the sparse coding problem
(\ref{eqn: src}) would be a time-consuming task. Additionally, it would be
extremely redundant if some images in one class look very similar. To address
this problem, several dictionary learning methods have been proposed for the
problem of reconstruction. Concretely, from the big training set $\bY$ of one
class, the dictionary learning algorithm tries to find a comprehensive set of
bases which have ability to sparsely represent all element in the set. This task
could be done by solving the following optimization problem:
\begin{equation}
    \label{eqn: dl0}
    (\bD^*, \bX^* ) = \arg \min_{\bD, \bX} \frac{1}{N}\sum_{i=1}^N \Big\{\norm{\by_i - \bD\bx_i}_2^2 + \lambda\norm{\bx_i}_1\Big\}
\end{equation}
where columns of $\bD$ are constrained by $\norm{\bd_j}_2^2 \leq 1$ to avoid trivial solutions. Problem (\ref{eqn: dl0}) could be rewritten in the matrix form:
\begin{equation}
    \label{eqn: dl}
    (\bD^*, \bX^* ) = \arg \min_{\bD, \bX} \Big\{\norm{\bY - \bD\bX}_F^2 + \lambda\norm{\bX}_1\Big\}
\end{equation}
where $\norm{\bX}_1$ is the sum of absolute values of all elements in $\bX$.
\par
Problem (\ref{eqn: dl}) is not simultaneously convex with respect to both $\bD$ and $\bX$ but convex with each variable if the other variable is fixed. The typical approach is as follows. First, suppose that $\bD$ is fixed, then $\bX$ could be found using LASSO\cite{tibshirani1996regression}:
\begin{equation}
    \bX^* = \arg\min_{\bX} \{\|\bY - \bD\bX\|_F^2 + \lambda\|\bX\|_1\}
\end{equation}
Second, fixing $\bX = \bX^*$ and compute $\bD$ by:
\begin{eqnarray}
    \bD^* &= &\arg\min_{\bD} \{\|\bY - \bD\bX\|_F^2 + \lambda\|\bX\|_1\} \\
    & = & \arg\min_{\bD} \{ \trace(\bD\Fb\bD^T) - 2\trace(\bD^T\bE) \}
    \label{eqn: findDODL}
\end{eqnarray}
Mairal \etal\cite{mairal2010online} propose an algorithm for computing $\bD$ by updating column by column until convergence:
  \begin{eqnarray}
    \bu_j &\leftarrow &\frac{1}{\Fb(j,j)}(\be - \bD\fb_j) + \bd_j \\
    \label{eqn: updateuj}
    \bd_j & \leftarrow & \frac{1}{\max(\norm{\bu_j}, 1)}\bu_j
    \label{eqn: updatedj}
  \end{eqnarray}
where $\Fb(j,j)$ is the value of $\Fb$ at coordinate $(j,j)$ and $\fb_j$ denotes the $j$-th column of $\Fb$.
\par 

The dictionary learned from this method has been shown to be comprehensive in terms of sparsely expressing in-class samples. Because ODL is trained using in-class samples only, it might or might not well present complementary samples. However, from classification view point, this fact could be useful in terms of discriminability. Several dictionary learning methods have been proposed for the classification purpose. Two well-known dictionary learning methods demonstrating impressive results in object recognition are LC-KSVD\cite{Zhuolin2013LCKSVD} and FDDL\cite{Meng2011FDDL}, which are also reviewed in the following sections of this chapter.
\subsection{Label Consistent K-SVD (LC-KSVD)} 
\label{sec:label_consistent_k_svdcite_zhuolin2013lcksvd}
Z. Jiang \etal\cite{Zhuolin2013LCKSVD} propose another dictionary learning method which learns a single over-complete dictionary and an optimal linear classifier simultaneously. It yields dictionaries so that feature points with the same class labels have similar sparse codes. {Each dictionary item is chosen so that it can be associated with a particular label}. It is claimed that the performance of the linear classifier depends on the discriminability  of the input sparse codes $\bs$. For obtaining discriminative sparse codes $\bs$ with the learned $\bD$, an objective function for dictionary construction is defined as:
\begin{eqnarray}
    (\hat{\bD}, \hat{\bX}, \hat{\bA}) &=& \arg\min_{\bD, \bX, \bA}\|\bY - \bD\bX\|_F^2 + \alpha \|\bQ - \bA\bX\|_F^2 \\
    \text{subject to:} && \|\bx_i\|_0 \leq L
\end{eqnarray}
where $\alpha$ controls the relative contribution between reconstruction and label consistent regularization, and $\bQ = [\bq_1, \bq_2, \dots, \bq_N] \in \R^{d \times N}$ are the 'discriminative' sparse codes of input signals $\bY$ for classification. They say that $\bq_i = [\bq_i^1, \bq_i^2, \dots, \bq_i^N]^T = [0,\dots, 1, 1, \dots, 0]^T \in \R^d$ is a 'discriminative' sparse code corresponding to an input signal $\by_i$, if the non-zero values of $\bq_i$ occur at those indicates where the input signal $\by_i$ and the dictionary item $\bd_k$ share the same label. 
\par 
\textbf{Classification scheme}: After the desired dictionary $\hat{\bD}$ has been learned, for a test image $\by$, they first compute its sparse representation $\bs$ by solving the optimization problem:
\begin{equation}
    \bs = \arg\min_{\bx}\|\by - \hat{\bD}\bs\|_2^2 ~~\text{subject to } ~ \|\bs\|_0 \leq L
\end{equation}
Then label of the image $\by$ is estimated by using the linear predictive classifier $\hat{\bW}$:
\begin{equation}
    \text{identity}(\by) = \arg\max_{j}(l = \hat{\bW}\bs)
\end{equation}
where $l \in \R^m$ is the class label vector. 

\subsection{Fisher Discrimination Dictionary Learning (FDDL)} 
\label{sec:fisher_discrimination_dic}
M. Yang \etal\cite{Meng2011FDDL} proposed another dictionary learning method to improve the pattern classification performance compared to SRC\cite{Wright2009SRC}. A structured dictionary, whose bases have correspondence to the class labels, is learned so that the reconstruction error after sparse coding can be used for pattern classification. Meanwhile, the Fisher discrimination criterion is imposed on the coding coefficients so that they have small within-class scatter but big between-class scatter.
\par 
Specifically, the ``total'' dictionary $\bD$ consisting of $c$ class-specific
dictionaries $\bD = [\bD_1, \bD_2, \dots, \bD_c]$ is learned via the following
optimization problem:
\begin{equation} J_{\bD, \bX} = \arg\min_{\bD, \bX} \Big\{\sum_{i=1}^c
    r(\bY_i,\bD, \bX_i) + \lambda_1\|\bX\|_1 + \lambda_2\big(\trace(S_W(\bX) -
    S_B(\bX)) + \eta\|\bX\|_F^2\big)\Big\}
\end{equation}
where:
\begin{itemize}
  \item $\displaystyle r(\bY_i,\bD, \bX_i) = \|\bY_i - \bD\bX_i\|_F^2 + \|\bY_i - \bD_i\bX_i^i\|_F^2 + \sum_{j = 1, j \neq i}^c\|\bD_j\bX_i^j\|_F^2$: the discriminative fidelity term.
  \item $\bX_i^j$: the coding coefficient of $\bY_i$ over the sub-dictionary $\bD_j$.
  \item $f(\bX) = \trace(S_W(\bX) - S_B(\bX)) + \eta\|\bX\|_F^2$: the discriminative coefficient term. 
\end{itemize}
More details about the optimization of FDDL and its classification scheme could be found at \cite{Meng2011FDDL}. 

\newpage
\section{Dissertation Contributions and Organization} 
\label{sec:dissertation_contributions_and_organization}

A snapshot of the main contributions of this dissertation is presented next.
Publications related to the contribution in each chapter are also listed where
applicable. 

In \textbf{Chapter 2}, the primary contribution is the proposal of a new
Discriminative Feature-oriented Dictionary Learning (DFDL) method for automatic
feature discovery in histopathological images. This proposal mitigates the
generally difficulty of feature extraction in histopathological images. Our
\emph{discriminative} framework learns dictionaries that emphasize inter-class
differences while keeping intra-class differences small, resulting in {enhanced}
classification performance. The design is based on solving a sparsity
constrained optimization problem, for which we develop a tractable algorithmic
solution.

\begin{figure*}[t]
\centering
  \includegraphics[width=.9\textwidth]{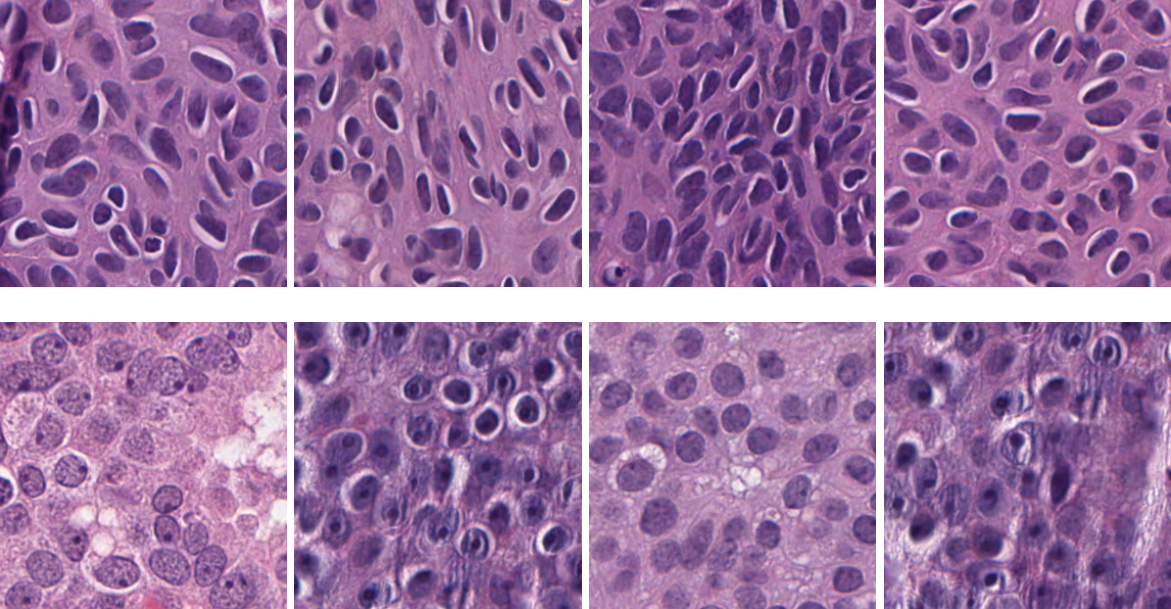}
  \vspace{-0.1in}
\caption{Samples form IBL dataset. First row: UDH. Second row: DCIS}
\label{fig:iblsamples}
\end{figure*}
\begin{figure*}[t]
\centering
  \includegraphics[width=.9\textwidth]{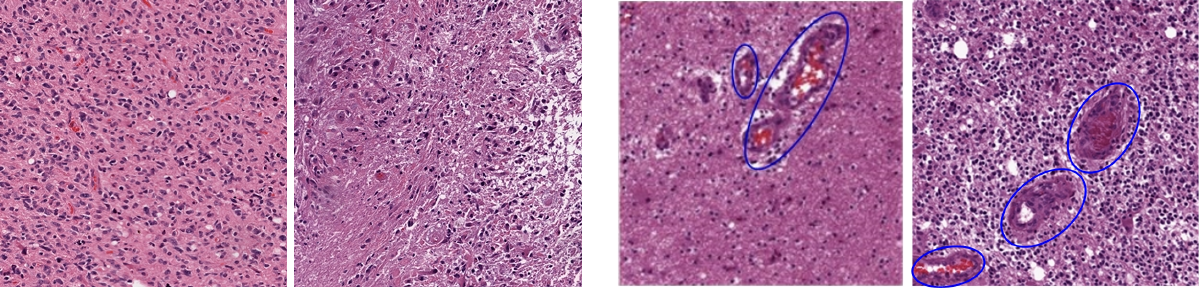}
  \vspace{-0.1in}
\caption{\small Samples {from} TCGA dataset. Left two images: regions without MVP. Right two images: regions with MVP are inside blue ovals.}
\label{fig:tcgasamples}
\end{figure*}


\begin{figure}[t]
\centering
  \includegraphics[width=\textwidth]{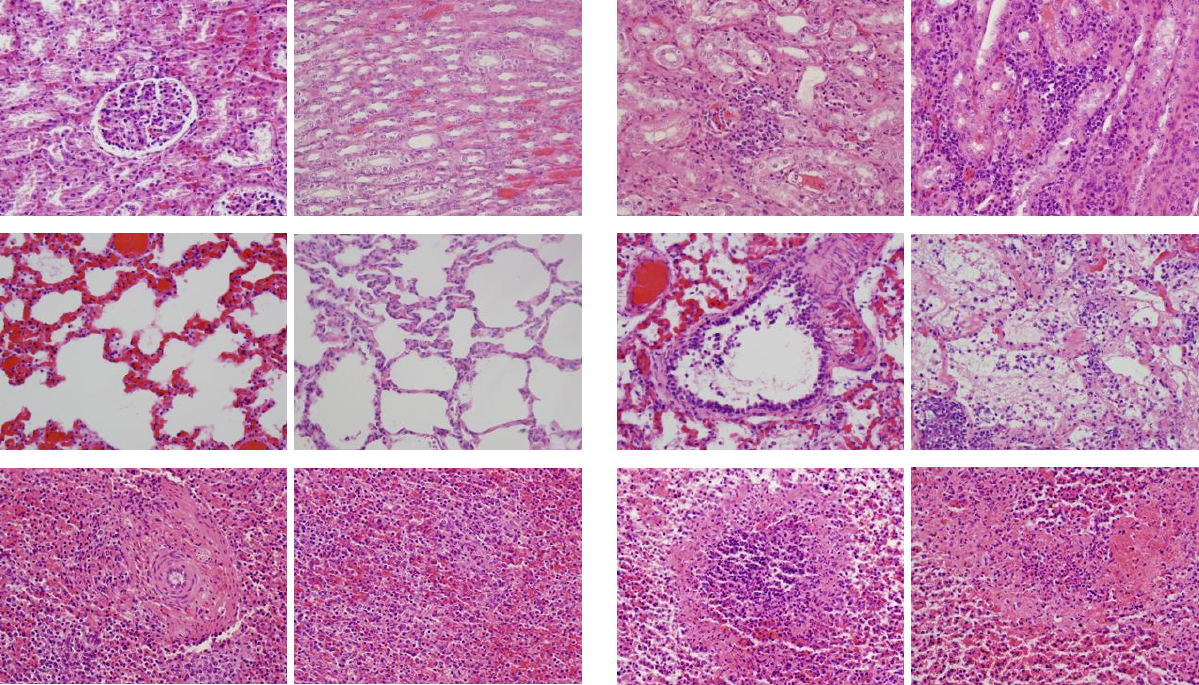}
  \vspace{-0.1in}
\caption{Samples form ADL dataset. First row: kidney. Second row: lung. Last row: spleen. First two columns: healthy. Last two columns: inflammatory.}
\label{fig:adlsamples}
\end{figure}

Experimental validation of DFDL is carried out on three diverse
histopathological datasets to show its broad applicability. \textit{The first
dataset} is courtesy of the Clarian Pathology Lab and Computer and Information
Science Dept., Indiana University-Purdue University Indianapolis (IUPUI). The
images acquired by the process described in \cite{Dundar2011} correspond to
human Intraductal Breast Lesions (IBL). Two well-defined categories will be
classified: Usual Ductal Hyperplasia (UDH)--benign, and Ductal Carcinoma In Situ
(DCIS)--actionable. \textit{The second dataset} contains images of brain cancer
{(glioblastoma or GBM)} obtaind from The Cancer Genome Atlas (TCGA)~\cite{TCGA}
provided by the National Institute of Health, and will henceforth be referred as
the TCGA dataset. For this dataset, we address the problem of detecting
MicroVascular Proliferation (MVP) regions, which is an important indicator of a
high grade glioma (HGG)~\cite{Mousavi2015JPI}. \textit{The third dataset} is
provided by the Animal Diagnostics Lab (ADL), The Pennsylvania State University.
It contains tissue images from three mammalian organs - kidney, lung and spleen.
For each organ, images will be assigned into one of two categories--healthy or
inflammatory.  
The samples of these three datasets are given in Figs.~\ref{fig:iblsamples},
~\ref{fig:tcgasamples}, and~\ref{fig:adlsamples}, respectively. Extensive
experimental results show that our method outperforms many competing methods,
particularly in low training scenarios. In addition, Receiver Operating
Characteristic (ROC) curves are provided that facilitate a trade-off between
false alarm and miss rates. This work was done in collaboration with Prof.
Ganesh Rao at the Department of Neurosurgery, and Prof. UK Arvind Rao
at the Department of Bionformatics and Computational Biology, both at
the University of Texas MD Anderson Cancer Center, Houston, TX, USA.

We derive the computational complexity of DFDL as well as competing dictionary
learning methods in terms of approximate number of operations needed. We also
report experimental running time of DFDL and three other dictionary learning
methods. All results in the manuscript are reproducible via a user-friendly
software\footnote{The software can be downloaded at
\url{http://signal.ee.psu.edu/dfdl.html}}. The software (MATLAB toolbox) is also
provided with the hope of usage in future research and comparisons via peer
researchers.

This material was presented at the 2015 IEEE International Symposium on
Biomedical Imaging and appeared in the IEEE Transactions of Medical Imaging in
March 2016.


\textbf{Chapter 3} proposes a new low-rank shared dictionary learning framework
(LRSDL) for automatically extracting both discriminative and shared bases in
several widely used image datasets to enhance the classification
performance of dictionary learning methods. Our framework simultaneously learns
each class-dictionary per class to extract discriminative features and the
shared features that all classes contain. For the shared part, we impose two
intuitive constraints. First, the shared dictionary must have a low-rank
structure. Otherwise, the shared dictionary may also expand to contain
discriminative features. Second, we contend that the sparse coefficients
corresponding to the shared dictionary should be almost similar. In other words,
the contribution of the shared dictionary to reconstruct every signal should be
close together. We will experimentally show that both of these constraints are
crucial for the shared dictionary.

Significantly, new accurate and efficient algorithms for selected existing and
proposed dictionary learning methods are proposed. We present three effective
algorithms for dictionary learning: i) sparse coefficient update in FDDL
\cite{yang2014sparse} by using FISTA \cite{beck2009fast}. We address the main
challenge in this algorithm -- how to calculate the gradient of a complicated
function effectively -- by introducing a new simple function
$\mathcal{M}(\bullet)$ on block matrices and a lemma to support the result. ii)
Dictionary update in FDDL \cite{yang2014sparse} by a simple ODL
\cite{mairal2010online} procedure using $\mathcal{M}(\bullet)$ and another
lemma. {Because it is an extension of FDDL, the proposed LRSDL also benefits
from the aforementioned efficient procedures.} iii) Dictionary update in DLSI
\cite{ramirez2010classification} by a simple ADMM \cite{boyd2011distributed}
procedure which requires only one matrix inversion instead of several matrix
inversions as originally proposed in \cite{ramirez2010classification}. We
subsequently show the proposed algorithms have both performance and
computational benefits.

We derive the computational complexity of numerous dictionary learning methods
in terms of approximate number of operations (multiplications) needed. We also
report complexities and experimental running time of aforementioned efficient
algorithms and their original counterparts. Numerous sparse coding and
dictionary learning algorithms in the manuscript are reproducible via a
user-friendly toolbox. The toolbox includes implementations of SRC
\cite{Wright2009SRC}, ODL \cite{mairal2010online}, LC-KSVD
\cite{Zhuolin2013LCKSVD}\footnote{Source code for LC-KSVD is directly taken from
the paper at:\\
\texttt{http://www.umiacs.umd.edu/$\sim$zhuolin/projectlcksvd.html}.}, efficient
DLSI \cite{ramirez2010classification}, efficient COPAR
\cite{kong2012dictionary}, efficient FDDL \cite{yang2014sparse}, $D^2L^2R^2$
\cite{li2014learning} and the proposed LRSDL. The toolbox (a MATLAB version and
a Python version) is provided\footnote{The toolbox can be downloaded at:
{\texttt{https://github.com/tiepvupsu/DICTOL}}.} with the hope of usage in
future research and comparisons via peer researchers.

The material in Chapter 3 was presented at the 2016 IEEE International
Conference on Image Processing and appeared in the IEEE Transactions on Image
Processing in November 2017. 


In \textbf{Chapter 4}, a framework for simultaneously denoising and classifying 2-D
UWB SAR imagery is introduced. Subtle features from targets of interest are
directly learned from their SAR imagery. The classification also exploits
polarization diversity and consecutive aspect angle dependence information of
targets. A generalized tensor discriminative dictionary learning (TensorDL) is
also proposed when more training data involved. These dictionary learning
frameworks are shown to be robust even with high levels of noise.

Additionally, {a} relative SRC framework (ShiftSRC) is proposed to deal
with multi-look data. Low-frequency UWB SAR signals are often captured at
different views of objects, depending on the movement of the radar carriers.
These signals contain uniquely important information of consecutive views.
With ShiftSRC, this information will be comprehensively exploited.
Importantly, a solution to the ShiftSRC framework can be obtained by an
elegant modification on the training dictionary, resulting in a tensor
sparse coding problem, which is similar to a problem proposed in 
the last paragraph. This work was done under the mentorship of Dr. Lam Nguyen at
the U.S. Army Research Laboratory, Adelphi, MD.  
All tensor sparsity algorithms in this chapter are reproducible via a
user-friendly toolbox. The toolbox written in Matlab is provided\footnote{
\href{https://github.com/tiepvupsu/tensorsparsity}{
\texttt{https://github.com/tiepvupsu/tensorsparsity}}} with the hope of usage in future research via peer
researchers.

The material in this chapter was presented at the 2017 IEEE Radar Conference and
is under review at the IEEE Transactions on Aerospace and Electronic Systems. On
a related note, we employed a deep learning framework for simultaneously
denoising and classifying UWB SAR imagery. This work is however not included in
this dissertation. This work was recently presented as an invited talk at the
2018 IEEE Radar Conference.

In \textbf{Chapter 5}, the main contributions of this dissertation are
summarized.

As a side note, apart from building discriminative models for signal
classification, our research also solves other long-standing open problems
in sparse signal and image processing. Using sparsity as a prior is tremendously
interesting in a wide variety of applications; however, existing solutions to
address this issue are sub-optimal and often fail to capture the intrinsic
sparse structure of physical phenomenon. We have been trying to address a very
fundamental question in this area of how to efficiently and effectively capture
sparsity in natural signals by using the Spike and Slab priors. These priors
have been of much recent interest in signal processing as a means of inducing
sparsity in Bayesian inference. It is well-known that solving for the sparse
coefficient vector to maximize these priors results in a hard non-convex and
mixed integer programming problem. Most existing solutions to this optimization
problem either involve simplifying assumptions/relaxations or are
computationally expensive. We propose a new greedy and adaptive matching
pursuit (AMP) algorithm to directly solve this hard problem. Essentially, in
each step of the algorithm, the set of active elements would be updated by
either adding or removing one index, whichever results in better improvement. In
addition, the intermediate steps of the algorithm are calculated via an
inexpensive Cholesky decomposition which makes the algorithm much faster.
Results on simulated data sets as well as real-world image recovery challenges
confirm the benefits of the proposed AMP, particularly in providing a superior
cost-quality trade-off over existing alternatives. These findings recently
appeared in 2017 at the IEEE International Conference on Acoustics, Speech,
and Signal Processing (ICASSP), and was nominated as a \textbf{Finalist for the
Best Student Paper Award}. This work is nevertheless not included in this
dissertation.

\chapter{Discriminative Feature-oriented Dictionary Learning for Histopathological Image Classification}
\label{chapter:contrib1}
\section{Introduction}
\label{sec:intro}
 Automated histopathological image analysis has recently become a significant research problem in medical imaging and there is an increasing need for developing quantitative image analysis methods as a complement to the effort of pathologists in diagnosis process. Consequently, an emerging class of problems in medical imaging focuses on the the development of computerized frameworks to classify histopathological images \cite{Gurcan2009,Srinivas2013,Srinivas2014SHIRC,Nandita2013,Mousavi2015JPI}. These advanced image analysis methods have been developed with {three main purposes of (i) relieving the workload on pathologists by sieving out obviously diseased and also healthy cases, which allows specialists to spend more time on more sophisticated cases; (ii) {reducing inter-expert variability}; and (iii) understanding the underlying reasons for a specific diagnosis that pathologists might not realize.}
\par
In the diagnosis process, pathologists often look for problem-specific visual
cues, or features, in histopathological images in order to categorize a tissue
image as one of the possible categories. These features might come from the
distinguishable characteristics of cells or nuclei, for example, size, shape or
texture\cite{Dundar2011,Gurcan2009}. They could also come from spatially related
structures of cells~\cite{Mousavi2015JPI,doyle2008automated,
tosun2011graph,Srinivas2014SHIRC}. In some cancer grading
problems, features might include the presence of particular regions
\cite{Mousavi2015JPI, hou2015efficient}. Consequently, different customized
feature extraction techniques for a variety of problems have been developed
based on these observed
features\cite{Orlov2008,Shamir2008,Gultekin2014,Shi2014,Minaee2013prediction}.
Morphological image features have been utilized in medical image
segmentation\cite{zana2001segmentation} for detection of vessel-like patterns.
Wavelet features and histograms are also a popular choice of features for
medical imaging\cite{chapelle1999support,unser2003guest}. Graph-based features
such as Delaunay triangulation, Vonoroi diagram, minimum spanning
tree\cite{doyle2008automated}, query graphs\cite{ozdemir2013hybrid} have been
also used to exploit spatial structures. Orlov \etal \cite{Orlov2008,Shamir2008}
have proposed a multi-purpose framework that collects texture information, image
statistics and transforms domain coefficients to be set of features. For
classification purposes, these features are combined with powerful classifiers
such as neural networks or support vector machines (SVMs). Gurcan \etal
\cite{Gurcan2009} provided detailed discussion of feature and classifier
selection for histopathological analysis.
\par Sparse representation frameworks have also been proposed for medical applications recently \cite{Srinivas2014SHIRC, Nandita2013, kopriva2015offset}. Specifically, Srinivas \etal \cite{Srinivas2013,Srinivas2014SHIRC} presented a multi-channel histopathological image as a sparse linear combination of training examples under channel-wise constraints and proposed a residual-based classification technique. Yu \etal\cite{Yu2011} proposed a method for cervigram segmentation based on sparsity and group clustering priors. Song \etal\cite{song2015locality,song2015large}  proposed a locality-constrained and a  large-margin representation method for medical image classification. In addition, Parvin \etal \cite{Nandita2013} combined a dictionary learning framework with an autoencoder to learn sparse features for classification. Chang \etal \cite{chang2013characterization} extended this work by adding a spatial pyramid matching to enhance the performance.
\subsection{Challenges and Motivation} 
\label{sub:motivation}
While histopathological analysis shares some traits with other image classification problems, there are also principally distinct challenges specific to histopathology.  The central challenge comes from the geometric richness of tissue images, resulting in the difficulty of obtaining reliable discriminative features for classification. Tissues from different organs have structural and morphological diversity which often leads to highly customized feature extraction solutions for each problem and hence the techniques lack broad applicability.


\par

\par

Being mindful of the aforementioned challenges, we design via optimization, a discriminative dictionary for each class by imposing sparsity constraints that minimizes intra-class differences, {while simultaneously} emphasizing inter-class differences. On one hand, small intra-class differences encourage the comprehensibility of the set of learned bases, which has ability of representing in-class samples with only few bases (intra class sparsity). This encouragement forces the model to find the representative bases in that class. On the other hand, large inter-class differences prevent bases of a class from sparsely representing samples from other classes.
Concretely, given a dictionary from a particular class $\bD$ with $k$ bases and a certain sparsity level $L\ll k$, we define an \emph{$L$-subspace} of $\bD$ as a span of a subset of $L$ bases from $\bD$. Our proposed Discriminative Feature-oriented Dictionary Learning (DFDL) aims to build dictionaries with this key property: any sample from a class is reasonably close to \emph{an} $L$-subspace of the associated dictionary while a complementary sample is far from \emph{any} $L$-subspace of that dictionary. Illustration of the proposed idea is shown in Fig. \ref{fig: idea}.

\begin{figure}[t]
\centering
  \includegraphics[width=0.87\textwidth]{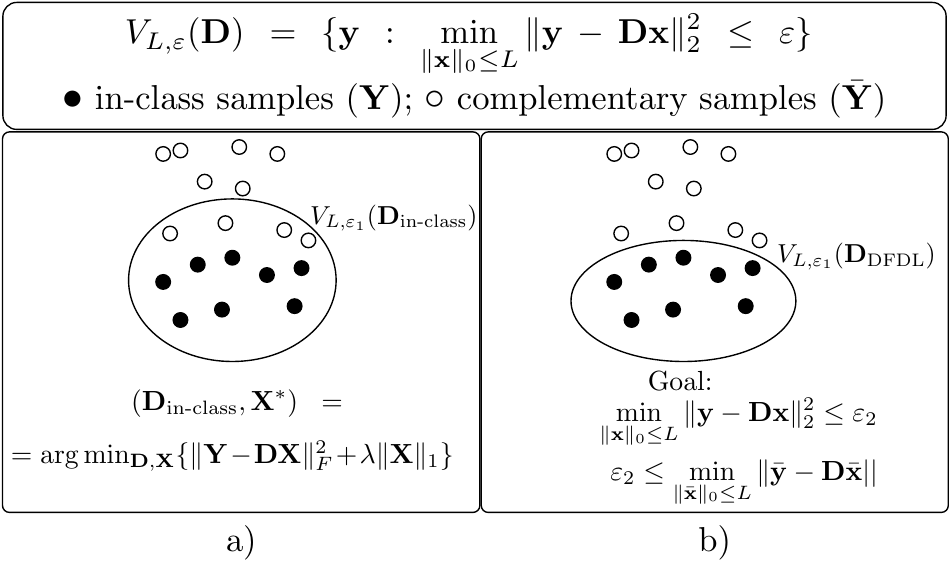}
  \vspace{-0.1in}
  \caption[DFDL: main idea]{\small Main idea: a) The sparse representation space of learned dictionary using in-class samples only, e.g. KSVD\cite{Aharon2006KSVD} or ODL\cite{mairal2010online}($V_{L,\varepsilon_1}(\bD_{\text{in-class}})$ may also cover some complementary samples), and b) desired DFDL ($V_{L,\varepsilon_2}(\bD_{\text{DFDL}})$ cover in-class samples only).}
  \label{fig: idea}
\end{figure}

\section{Contributions}
\label{sec:contributions}
\subsection{Notation} 
\vspace{-0.05in}
\label{sub:notaions}
The vectorization of a small block (or patch)\footnote{In our work, a training vector is obtained by vectorizing all three RGB channels followed by concatenating them together to have a long vector.} extracted from an image is denoted as a column vector $\by \in \R^d$ which will be referred as a sample.
In a classification problem where we have $c$ different categories, collection of all data samples from class $i$ ($i$ can vary between $1$ to $c$) forms the matrix $\bX_i \in \R^{d\times N_i}$ and let $\lbX_i \in \R^{d \times \bar{N}_i}$ be the matrix containing all complementary data samples i.e. those that are not in class $i$. We denote by $\bD_i \in \R^{d \times k_i}$ the dictionary of class $i$ that is desired to be learned through our DFDL method. 
\par
{For a vector $\bx \in \R^k$, we denote by $\|\bx\|_0$ the number of its non-zero elements. The sparsity constraint of $\bx$ can be formulated as $\|\bx\|_0 \le L$. For a matrix $\bX$ , $\|\bX\|_0 \le L$ means that \emph{each} column of $\bX$ has no more than $L$ non-zero elements.}
\subsection{Discriminative Feature-oriented Dictionary Learning} 
\label{sec:DFDL}
We aim to build {class-specific} dictionaries $\bD_i$ such that each $\bD_i$ can sparsely represent samples from class $i$ but is \emph{poorly} capable of representing its complementary samples with small number of bases. Concretely, for the learned dictionaries we need:
\begin{eqnarray*}
  &\displaystyle \min_{\|\bx_l\|_0 \leq L_i }\|\by_l - \bD_i \bx_l\|_2^2,~~ \forall l = 1, 2, \dots, N_i & \text{to be small}\\
 \text{and } &\displaystyle \min_{\|\bar{\bx}_m\|_0 \leq L_i}\|\bar{\by}_m - \bD_i \bar{\bx}_m\|_2^2,~~ \forall m = 1,2,\dots, \bar{N}_i & \text{to be large.}
\end{eqnarray*}
 where $L_i$ controls the sparsity level. These two sets of conditions could be simplified in the matrix form:
 \begin{eqnarray}
  \label{eqn:intra}
   \text{intra-class differences: }\displaystyle\frac{ 1  } { N_i } \min_{   \|\bX_i\|_0 \leq L_i }\|\bY_i - \bD_i \bX_i\|_F^2 & \text{small,}\\
  \label{eqn:inter}
 \text{inter-class differences: } \displaystyle \frac{ 1  } { \bar{N_i} } \min_{\|\lbY_i\|_0 \leq L_i}\|\lbX_i - \bD_i \lbX_i\|_F^2 & \text{large.}
 \end{eqnarray}

The averaging operations $\left(\displaystyle\frac{ 1  } { N_i } \text{~and~} \displaystyle\frac{1}{\bar{N_i}}\right)$  are taken here  for avoiding the case where the largeness of inter-class differences is solely resulting from $\bar{N_i} \gg N_i$.
 \par For simplicity, from now on, we consider only one class and drop the class index in each notion, i.e., using $\bY, \bD, \bX, \lbX, N, \bar{N}, L$ instead of $\bY_i, \bD_i, \bX_i, \lbX_i, N_i, \bar{N}_i$ and $L_i$.
 Based on the argument above, we formulate the optimization problem for each dictionary:
 \begin{equation}
   \bD^*= \arg \min _{\bD} \Big(\frac{ 1  } { N }\min_{\|\bX\|_0 \le L}\|\bY - \bD \bX\|_F^2 - \frac{ \rho  } { \bar{N} } \min_{\|\lbX\|_0 \le L}\|\lbY - \bD \lbX\|_F^2 \Big),
 \label{eqn:findDopt}
 \end{equation}
 where $\rho$ is a positive regularization parameter. The first term in the above optimization problem encourages intra-class differences to be small, while the second term, with minus sign, emphasizes inter-class differences. By solving the above problem, we can jointly find the appropriate dictionaries as we desire in (\ref{eqn:intra}) and (\ref{eqn:inter}).
 \par
 \textbf{How to choose $L$:} The sparsity level $L$ for classes might be different. For one class, if $L$ is too small, the dictionary might not appropriately express in-class samples, while if it is too large, the dictionary might be able to represent complementary samples as well. In both cases, the classifier might fail to determine identity of one new test sample. We propose a method for estimating $L$ as follows. First, a dictionary is learned using ODL\cite{mairal2010online} using in-class samples $\bY$ only:
 \begin{equation}
     (\bD^0, \bX^0) = \arg\min_{\bD, \bX}\{\|\mathbf{Y} - \bD \bX\|_F^2 + \lambda\|\bX\|_1\},
     \label{eqn:findD0}
 \end{equation}
 where $\lambda$ is a positive regularization parameter controlling the sparsity level. {Note that the same $\lambda$ can still lead to different $L$ for different classes, depending on the intra-class variablity of each class. Without prior knowledge of those variablities, we choose the same $\lambda$ for every class.} After $\bD^0$ and $\bX^0$ have been computed, $\bD^0$ could be utilized as a warm initialization of $\bD$ in our algorithm, $\bX^0$ could be used to estimate the sparsity level $L$:
 \begin{equation}
    L \approx \frac{1}{N}\sum_{i=1}^{N}\|\bx_i^0\|_0.
     \label{eqn:findL}
 \end{equation}

\textbf{Classification scheme:} In the same manner with SRC \cite{Wright2009SRC}, a new patch $\by$ is classified as follows. Firstly, the sparse codes $\hat{\bx}$ are calculated via $l_1$-norm minimization:
\begin{equation}
    \hat{\bx} = \arg \min_{\bx} \big\{ \|\by - \bD_{total}\bx\|_2^2 + \gamma \|\bx\|_1 \big\},
    \label{eqn:class1}
\end{equation}
    where $\bD_{total} = [\bD_1, \bD_2, \dots, \bD_c]$ is the collection of all dictionaries and $\gamma$ is a scalar constant.
    Secondly, the identity of $\by$ is determined as: $\displaystyle \arg \min_{i \in \{1, \dots, c\}}\{r_i(\by)\} $ where
\begin{equation}
  r_i(\by) = \|\by-\bD_i \delta_i(\hat{\bx})\|_2
  \label{eqn:class2}
\end{equation}
    and $\delta_i(\hat{\bx})$ is part of $\hat{\bx}$ associated with class $i$.


\subsection{Proposed solution}
\label{subsec: solution}
We use an iterative method to find the optimal solution for the problem in (\ref{eqn:findDopt}). Specifically, the process is iterative by fixing $\bD$ while optimizing $\bX$ and $\lbX$ and vice versa. 
\par
In the sparse coding step, with fixed $\bD$, optimal sparse codes $\bX^*, \lbX^*$ can be found by solving:
$$
       \bX^* = \arg \min_{\|\bX\|_0 \le L}\|\bY - \bD\bX\|_F^2;~~
       \lbX^* = \arg \min_{\|\lbX\|_0 \le L} \|\lbY - \bD\lbX\|_F^2.
       $$
\par  With the same dictionary $\bD$, these two sparse coding problems can be combined into the following one:
  \begin{equation}\label{eqn:findS}
    \hat{\bX}^* = \arg\min_{\|\hat{\bX}\|_0 \le L}\norm{\hat{\bY} - \bD\hat{\bX}}_F^2.
  \end{equation}
  with $\hat{\bY} = [\bY, \lbY]$ being the matrix of all training samples and $\hat{\bX} = [\bX, \lbX]$. This sparse coding problem can be solved effectively by OMP\cite{tropp2007signal} using SPAMS toolbox\cite{SPAMS}.

  \par
For the bases update stage, $\bD^*$ is found by solving: 
\begin{align}
  \label{subeqn:findD1}
  \bD^* &= \arg\min_{\bD} \Big\{ \frac{ 1  } { N } \|\bY - \bD\bX\|_F^2 - \frac{ \rho } { \barN } \|\lbY - \bD\lbX\|_F^2 \Big\}, \\
  \label{subeqn:findD2}
  &=\arg\min_{\bD} \big\{-2\trace(\bE \bD^T) + \trace(\bD \Fb \bD^T) \big\}.
\end{align}

\noindent
We have used the equation $\|\bM\|_F^2 = \trace(\bM\bM^T)$ for any matrix $\bM$ to derive (\ref{subeqn:findD2}) from (\ref{subeqn:findD1}) and denoted:
\begin{equation}
\label{eqn:defineEF}
        \bE = \frac{1}{N} \bY \bX^T - \frac{ \rho } { \barN } \barX\barX^T; \quad
        \Fb = \frac{ 1  } { N } \bX \bX^T - \frac{ \rho } { \barN } \barX \barX^T.
\end{equation}

\par

{The objective function in (\ref{subeqn:findD2}) is very similar to the objective function in the dictionary update stage problem in \cite{mairal2010online} except that it is not guaranteed to be convex. It is convex if and only if $\Fb$ is positive semidefinite. For the discriminative dictionary learning problem, the symmetric matrix $\Fb$ is {\em not} guaranteed to be positive semidefinite, even all of its eigenvalues are real. In the worst case, where $\Fb$ is negative semidefinite, the objective function in (\ref{subeqn:findD2}) becomes concave; if we apply the same dictionary update algorithm as in \cite{mairal2010online}, we will obtain its maximum solution instead of the minimum.}
  \par
{To deal with this situation, we propose a technique which convexifies the objective function based on the following observation.
}
  \par {If we look back to the main optimization problem stated in (\ref{eqn:findDopt}):
    \begin{equation*} 
        \bD^* = \arg\min_{\bD} \left(\frac{1}{N} \min_{\|\bX\|_0 \leq L} \|\bY - \bD\bX\|_F^2 - \frac{\rho}{\bar{N}} \min_{\|\lbX\|_0 \leq L} \|\lbY - \bD \lbX\|_F^2\right),
    \end{equation*}
    we can see that if $\bD = \bmt \bd_1 & \bd_2 & \dots & \bd_k \emt $ is an optimal solution, then $\displaystyle \bD = \bmt \frac{\bd_1}{a_1} & \frac{\bd_2}{a_2} & \dots & \frac{\bd_k}{a_k} \emt$ is also an optimal solution as we multiply $j$-th rows of optimal $\bX$ and $ \lbX$ by $a_j$, where $a_j, j = 1, 2, \dots, k,$ are arbitrary nonzero scalars. Consequently, we can introduce constraints: $\|\bd_i\|_2^2 = 1, j = 1, 2, \dots, k$, without affecting optimal value of (\ref{subeqn:findD2}). With these constraints, $\trace(\bD \lambda_{\min} (\Fb)\mathbf{I}_k \bD^{T}) = \lambda_{\min}(\Fb)\trace(\bD^{T}\bD) = \lambda_{\min}(\Fb)\sum_{i = 1}^k \bd_i^{T} \bd_i = k\lambda_{\min}(\Fb)$, where $\lambda_{\min}(\Fb)$ is the minimum eigenvalue of $\Fb$ and $\mathbf{I}_k$ denotes the identity matrix, is a constant. Substracting this constant from the objective function will not change the optimal solution to (\ref{subeqn:findD2}).}

\begin{algorithm}[t]
    \caption{Discriminative Feature-oriented Dictionary Learning}\label{alg:DFDL}
    \begin{algorithmic}
    \Function {$\bD^*$ = DFDL}{$\bY, \lbY, k, \rho$}
    \State \tb{INPUT:} {$\bY, \lbY$: collection of all in-class samples and complementary samples. $k$: number of bases in the dictionary. $\rho$: the regularization parameter.}
    \State 1. Choose initial $\bD^*$ and $L$ as in (\ref{eqn:findD0}) and (\ref{eqn:findL}).
    \While{not converged}
        \State 2. Fix $\bD = \bD^*$ and update $\bX, \lbX$ by solving (\ref{eqn:findS});
        \State 3. Fix $\bX, \lbX$, calculate:
        \begin{equation*}
            \bE = \frac{ 1 } { N } \bY \bX^T - \frac{ \rho } { \bar{N} } \barX\barX^T; \quad
             \Fb = \frac{ 1  } { N } \bX \bX^T - \frac{ \rho } { \bar{N} } \barX \barX^T.
        \end{equation*}
         \State 4. {Update $\bD$ from:
         \begin{equation*}
           \bD^*= \arg \min_{\bD} \Big\{-2\trace(\bE \bD^T) + \trace\Big(\bD \big(\Fb - \lambda_{\min}(\Fb) \mathbf{I}\big) \bD^T\Big)  \Big\}
         \end{equation*}
         \begin{equation*}
           \text{subject to:}  \|\bd_i\|_2^2 = 1, i = 1, 2, \dots, k.
         \end{equation*}
           }
      \EndWhile
      \State \tb{RETURN:} $\bD^*$
    \EndFunction
    \end{algorithmic}
\end{algorithm}
{Essentially, the following problem in (\ref{eq:newProb}) is equivalent to (\ref{subeqn:findD2}):
\begin{equation} 
  \bD^* = \arg\min_{\bD}\{-2\trace(\bE\bD^{T}) +  \trace\big(\bD(\Fb - \lambda_{\min}(\Fb)\mathbf{I}_k)\bD^{T}\big)\}
  \label{eq:newProb}
\end{equation}
  $$\text{subject to:}  \|\bd_i\|_2^2 = 1, i = 1, 2, \dots, k.$$
The matrix $\hat{\Fb} = \Fb - \lambda_{\min}(\Fb)\mathbf{I}_k$ is guaranteed to be positive semidefinite since all of its eignenvalues now are nonnegative, and hence the objective function in (\ref{eq:newProb}) is convex. Now, this optimization problem is very similar to the dictionary update problem in \cite{mairal2010online}. Then, $\bD^*$ could be updated by the following iterations until convergence:
\begin{eqnarray}
    \label{eqn: updateuj}
    \bu_j &\leftarrow &\frac{1}{\hat{\Fb}_{j,j}}(\be_j - \bD\hat{\fb}_j) + \bd_j. \\
    \bd_j & \leftarrow & \frac{\bu_j}{\norm{\bu_j}_2}.
    \label{eqn: updatedj}
  \end{eqnarray}
where $\hat{\Fb}_{j,j}$ is the value of $\hat{\Fb}$ at coordinate $(j,j)$ and $\hat{\fb}_j$ denotes the $j$-th column of $\hat{\Fb}$.
}
\par
Our DFDL algorithm is summarized in Algorithm \ref{alg:DFDL}.
\subsection{Overall classification procedures for three datasets} 
\label{sec:overallclassification}
In this section, we propose a DFDL-based procedure for classifying images in three datasets.

\subsubsection{IBL and ADL datasets} 
\label{sub:ibladlprocedure}
\begin{figure*}[t]
\centering
  \includegraphics[width=\textwidth]{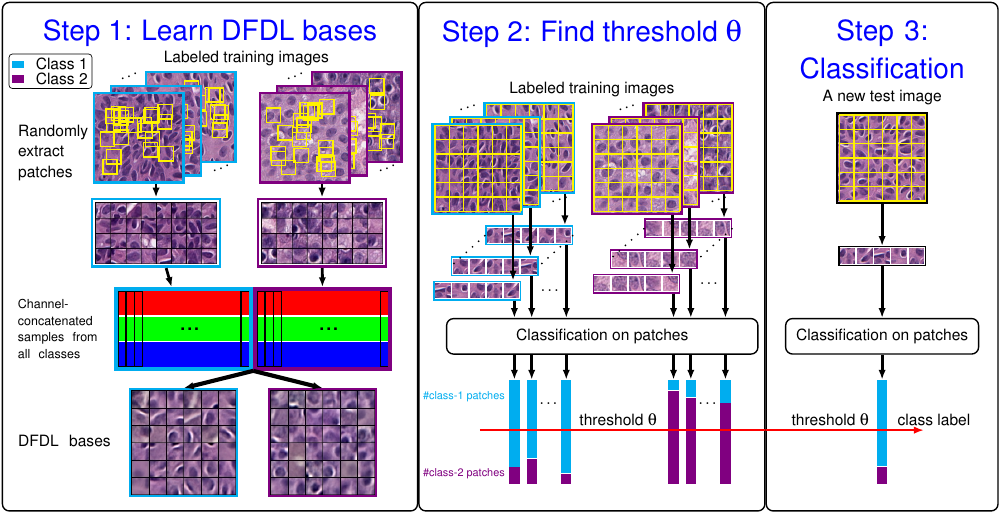}
  \vspace{-0.1in}
 \caption{\small IBL/ADL classification procedure }
  \label{fig: ibladlprocedure}
\end{figure*}
The key idea in this procedure is that a healthy tissue image largely consists of healthy patches which cover a dominant portion of the tissue. This procedure is shown in Fig. \ref{fig: ibladlprocedure} and consists of the following three steps:
\par \noindent \textbf{Step 1}: \textit{Training DFDL bases for each class}. From labeled training images, training patches are randomly extracted (they might be overlapping). The size of these patches is picked based on pathologist input and/or chosen by cross validation\cite{Kohavi95astudy}. After we have a set of \textit{healthy} patches and a set of \textit{diseased} patches for training, class-specific DFDL dictionaries and the associated classifier are trained by using Algorithm \ref{alg:DFDL}.
\par\noindent \textbf{Step 2:}\label{} \textit{Learning a threshold $\theta$ for proportion of \textit{healthy} patches in one \textit{healthy} image}. Labeled training images are now divided into non-overlapping patches. Each of these patches is then classified using the DFDL classifier as described in Eq. (\ref{eqn:class1}) and (\ref{eqn:class2}). The main purpose of this step is to find the threshold $\theta$ such that healthy images have proportion of \textit{healthy} patches greater or equal to $\theta$ and diseased ones have proportion of \textit{diseased} patches less than $\theta$. {We can consider the proportion of healthy patches in one training image as its one-dimension feature. This feature is then put into a simple SVM to learn the threshold $\theta.$}


\par \noindent \textbf{Step 3:} \textit{Classifying test images}. For an unseen test image, we calculate the proportion $\tau$ of \textit{healthy} patches in the same way described in Step 2. Now, the identity of the image is determined by comparing the proportion $\tau$ to $\theta$. It is categorized as healthy (diseased) if $\tau \geq (<) \theta$.
The procedure readily generalizes to multi-class problems.

\label{sec:implementation_pipeline_for_}
\subsubsection{MVP detection problem in TCGA dataset} 
\label{sec:tcgaprocedure}
\begin{figure}
  \centering
  \includegraphics[width = 0.7\textwidth]{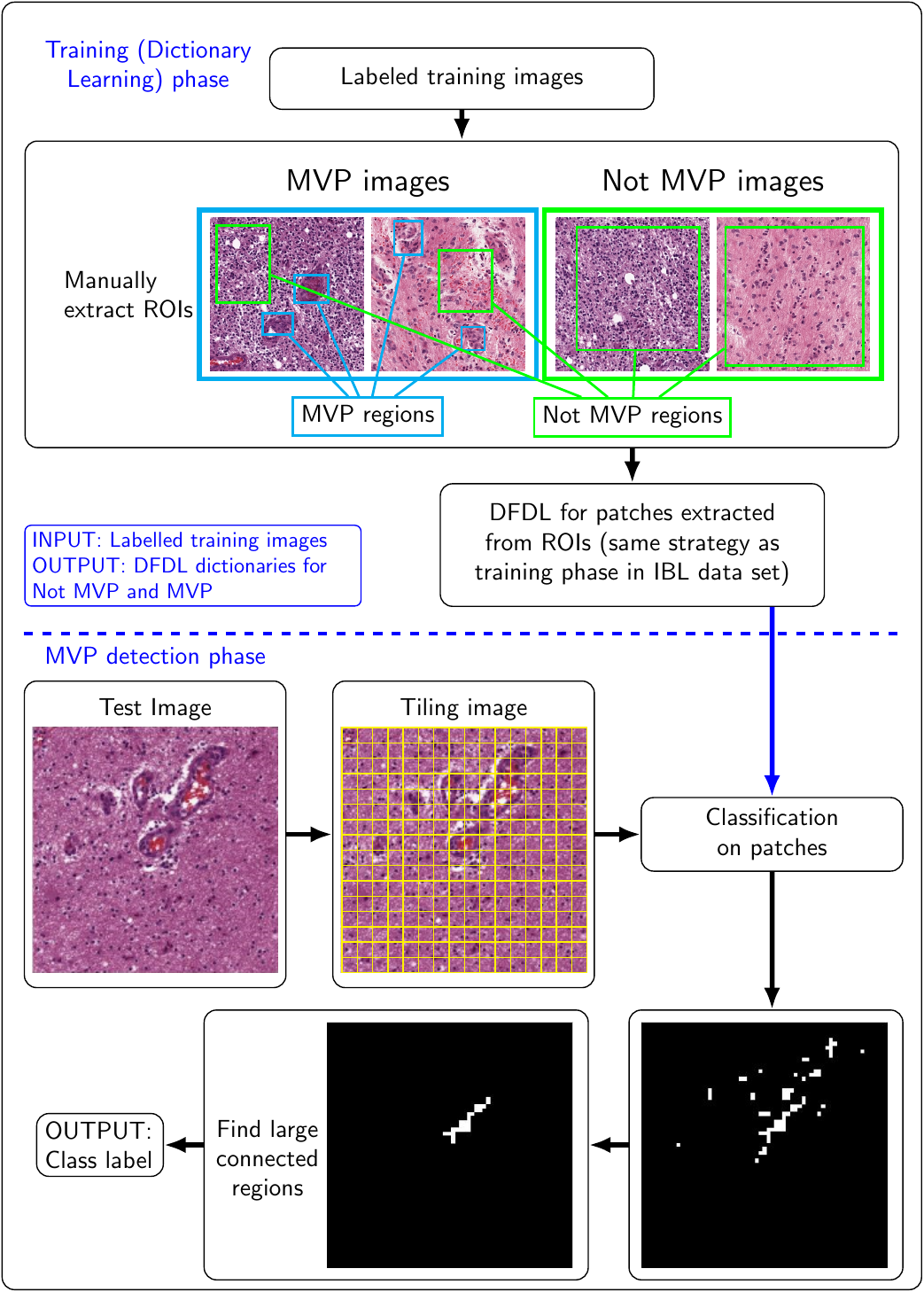}
  \caption{\small  MVP detection procedure}
  \label{fig:mvpdetection}
\end{figure}
\par As described earlier, MicroVascular Proliferation (MVP) is the presence of blood vessels in a tissue and it is an important indicator of a high-grade tumor in brain glioma. Essentially presence of one such region in the tissue image indicates the high-grade tumor. Detection of such regions in TCGA dataset is an inherently hard problem  and unlike classifying images in IBL and ADL datasets which are distinguishable by researching small regions, it requires more {effort} and investigation on larger connected regions. This is due to the fact that an MVP region may significantly vary in size and is usually surrounded by tumor cells which are actually benign or low grade. In addition, an MVP region is characterized by the presence of enlarged vessels in the tissue with different color shading and thick layers of cell rings inside the vessel (see Fig. \ref{fig:tcgasamples}).
We define a patch as \emph{MVP} if it lies entirely within an MVP region and as \emph{Not MVP} otherwise. We also define a region as Not MVP if it does not contain any MVP patch.
The procedure consists of two steps:
  \par \noindent\tb{Step 1:} \emph{Training phase}. From training data, MVP regions and Not MVP regions are manually extracted. Note that while MVP regions come from MVP images only, Not MVP regions might appear in all images. From these extracted regions, DFDL dictionaries are obtained in the same way as in step 1 of IBL/ADL classification procedure described in section \ref{sub:ibladlprocedure} and Fig. \ref{fig: ibladlprocedure}.
  \par{\noindent \tb{Step 2: } \emph{MVP detection phase:} A new unknown image is decomposed into non-overlapping patches. These patches are then classified using DFDL model learned before. After this step, we have a collection of patches classified as MVP. A region with large number of connected classified-as-MVP patches could be considered as an MVP region. If the final image does not contain any MVP region, we categorize the image as a Not MVP; otherwise, it is classified as MVP. {The definition of connected regions contains a parameter $m$, which is the number of connected patches. Depending on $m$, positive patches might or might not appear in the final step. Specifically, if $m$ is small, false positives tend to be determined as MVP patches; if $m$ is large, true positives are highly likely eliminated. To determine $m$, we vary it from 1 to 20 and compute its ROC curve for training images and then simply pick the point which is closest to the origin and find the {\em optimal} $m$. }}\label{par:choosem}
This procedure is visualized in Fig. \ref{fig:mvpdetection}. 
\par

\section{Validation and Experimental Results}
\label{sec:results} 

\label{sec:experiment_results}
In this section, we present the experimental results of applying DFDL to three diverse histopathological image datasets and compare our results with different competing methods:
  \par$\bullet$ WND-CHARM\cite{Orlov2008,Shamir2008} in conjunction with SVM: this method combines state-of-the-art feature extraction and classification methods. We use the collection of features from WND-CHARM, which is known to be a powerful toolkit of features for medical images. While the original paper used weighted nearest neighbor as a classifier, we use a more powerful classifier (SVM \cite{CC01a}) to further enhance classification accuracy. We pick the most relevant features for histopathology\cite{Gurcan2009}, including but not limited to (color channel-wise) histogram information, image statistics, morphological features and wavelet coefficients from each color channel. The source code for WND-CHARM is made available by the National Institute of Health online at \url{http://ome.grc.nia.nih.gov/}.
  \par$\bullet$ SRC\cite{Wright2009SRC}: We apply SRC on the vectorization of the luminance channel of the histopathological images, as proposed initially for face recognition and applied widely thereafter.
  \par$\bullet$ SHIRC\cite{Srinivas2014SHIRC}: Srinivas \etal\cite{Srinivas2013,Srinivas2014SHIRC} presented a simultaneous sparsity model for multi-channel histopathology image representation and classification which extends the standard SRC\cite{Wright2009SRC} approach by designing three color dictionaries corresponding to the RGB channels. The MATLAB code for the algorithms is posted online at: \url{http://signal.ee.psu.edu/histimg.html}.
  \par$\bullet$ LC-KSVD\cite{Zhuolin2013LCKSVD} and FDDL\cite{Meng2011FDDL}: These are two well-known dictionary learning methods which were applied to object recognition such as face, digit, gender, vehicle, animal, etc, but to our knowledge, have not been applied to histopathological image classification. To obtain a fair comparison, dictionaries are learned on the same training patches. Classification is then carried out using the learned dictionaries on non-overlapping patches in the same way described in Section \ref{sec:overallclassification}.

  \par$\bullet$ Nayak's: In recent relevant work, Nayak \etal\cite{Nandita2013} proposed a patch-based method to solve the problem of classification of tumor histopathology via sparse feature learning. The feature vectors are then fed into  SVM to find the class label of each patch. 
\par
\subsection{Experimental Set-Up: Image Datasets} 
\label{subsec:experimental_setup_image_data_sets}
\par
\textbf{IBL dataset:} Each image contains a number of regions of interest (RoIs), and we have chosen a total of 120 images (RoIs), consisting of a randomly selected set of 20 images for training and the remaining 100 RoIs for test. Images are downsampled for computational  purposes such that size of a cell is around 20-by-20 (pixels). Examples of images from this dataset are shown in Fig. \ref{fig:iblsamples}.
Experiments in section \ref{subsec:validation_of_central_idea_overall_classification_accuracy} below are conducted with 10 training images per class, 10000 patches of size 20-by-20 for training per class, $k=500$ bases for each dictionary, $\lambda = 0.1$ and $\rho = 0.001$. {These parameters are chosen using cross-validation \cite{Kohavi95astudy}.}


\par \textbf{ADL dataset:} This dataset contains bovine histopathology images from three sub-datasets of kidney, lung and  spleen. Each sub-dataset consists of images of size $4000\times 3000$ pixels from two classes: healthy and inflammatory. Each class has around 150 images from which 40 images are chosen for training, the remaining ones are used for testing. Number of training patches, bases, $\lambda$ and $\rho$ are the same as in the IBL dataset. The classification procedure for IBL and ADL datasets is described in Section \ref{sub:ibladlprocedure}.

\textbf{TCGA dataset:} We use a total of 190 images (RoIs) (resolution {$3000\times 3000$}) from the TCGA, in which 57 images contain MVP regions and 133 ones have no traces of MVP. From each class, 20 images are randomly selected for training. The classification procedure for this dataset is described in Section \ref{sec:tcgaprocedure}. 
\par {Each tissue specimen in these datasets is fixed on a scanning bed and digitized using a digitizer at 40$\times$ magnification.}
\subsection{Validation of Central Idea: Visualization of Discovered Features} 
\label{subsec:validation_of_central_idea_overall_classification_accuracy}
\begin{figure*}[t]
\centering
\includegraphics[width = 0.98\textwidth]{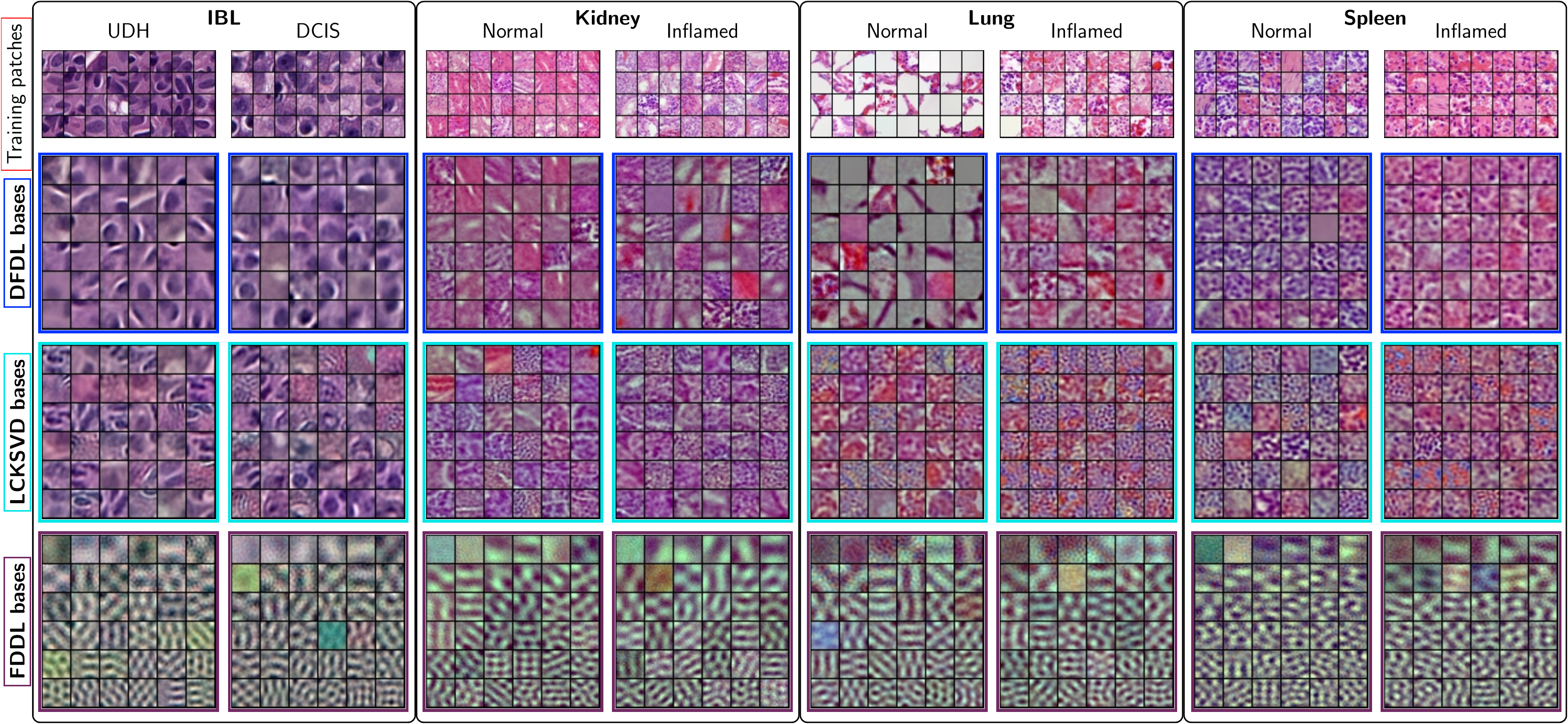}
\caption[Example bases learned from different methods on different datasets.]{\small  Example bases learned from different methods on different datasets. DFDL, LC-KSVD\cite{Zhuolin2013LCKSVD}, FDDL\cite{Meng2011FDDL} in IBL and ADL datasets. }
\label{fig:dictIKL}
\end{figure*}
\begin{figure}[t]
\centering
  \includegraphics[width=0.7\textwidth]{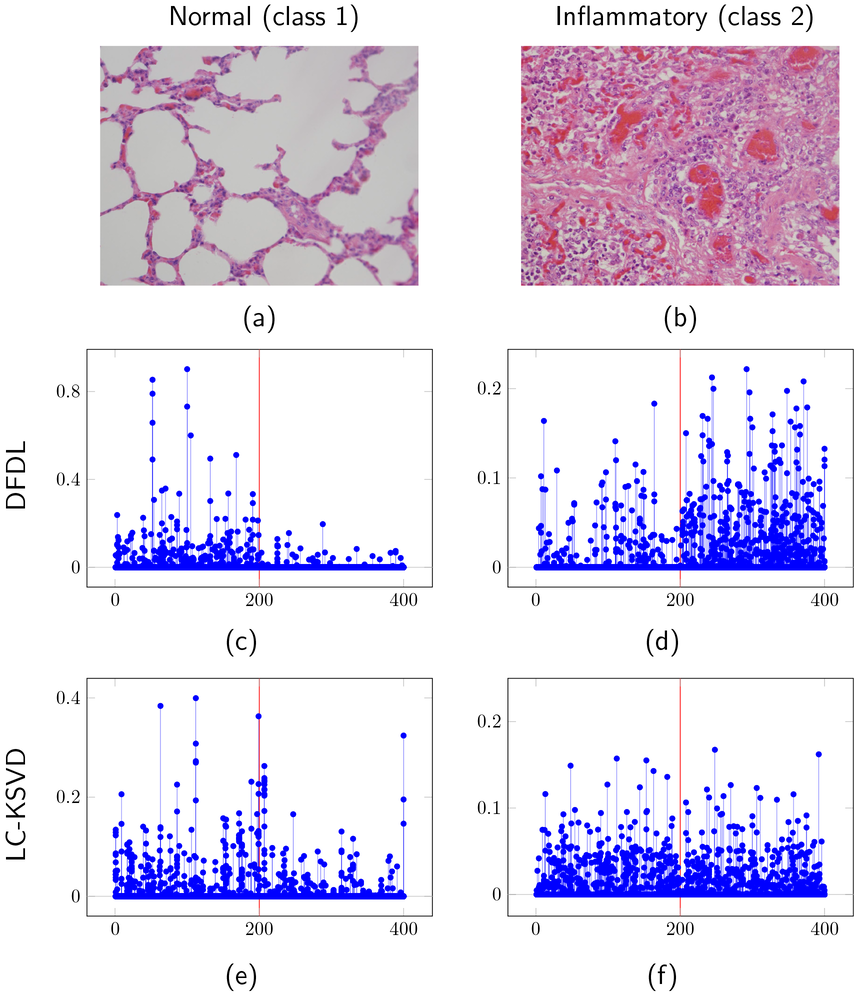}
 \caption[Example of sparse codes using DFDL and LC-KSVD approaches on lung dataset]{\small {Example of sparse codes using DFDL and LC-KSVD approaches on lung dataset. Left: normal lung (class 1). Right: inflammatory lung (class 2). Row 1: test images. Row 2: Sparse codes visualization using DFDL. Row 3: Sparse codes visualization using LC-KSVD. $x$ axis indicates the dimensions of sparse codes with codes on the left of red lines corresponding to bases of class 1, those on the right are in class 2. $y$ axis demonstrates values of those codes. In one vertical line, different dots represent values of non-zeros coefficients of different patches.}
 \vspace{0in}}
  \label{fig: visual_coef }
\end{figure}    

This section provides experimental validation of the central hypothesis of this chapter: by imposing sparsity constraint on forcing intra-class differences to be small, while simultaneously emphasizing inter-class differences, the class-specific bases obtained are discriminative.

\par {Example bases obtained by different dictionary learning methods are visualized in Fig. \ref{fig:dictIKL}. By visualizing these bases, we emphasize that our DFDL is able to look for discriminative visual features from which pathologists could understand the reasons behind diseases. In the spleen dataset for example, it is really difficult to realize the differences between two classes by human eyes. However, by looking at DFDL learned bases, we can see that the distribution of cells in two classes are different such that a larger number of cells appears in a normal patch. These differences may provide pathologists one visual cue to classify these images without advanced tools. Moreover, for IBL dataset, UDH bases visualize elongated cells with sharp edges while DCIS bases present more rounded cells with blurry boundaries, which is consistent with their descriptions in \cite{Srinivas2014SHIRC} and \cite{Dundar2011}; for ADL-Lung, we observe that a healthy lung is characterized by large clear openings of the alveoli, while in the inflamed lung, the alveoli are filled with bluish-purple inflammatory cells. This distinction is very clear in the bases learned from DFDL where white regions appear more in normal bases than in inflammatory bases and no such information can be deduced from LC-KSVD or FDDL bases. In comparison, FDDL fails to discover discriminative visual features that are interpretable and LC-KSVD learns bases with the inter-class differences being less significant than DFDL bases. Furthermore, these LC-KSVD bases do not present key properties of each class, especially in lung dataset.} 

\par
{To understand more about the significance of discriminative bases for classification, let us first go back to SRC~\cite{Wright2009SRC}. For simplicity, let us consider a problem with two classes with corresponding dictionaries $\bD_1$ and $\bD_2$. The identity of a new patch $\by$, which, for instance, comes from class 1, is determined by equations (\ref{eqn:class1}) and (\ref{eqn:class2}). In order to obtain good results, we expect most of active coefficients to be present in $\delta_1(\hat{\bx})$. For $\delta_2(\hat{\bx})$, its non-zeros, if they exists should have small magnitude. Now, suppose that one basis, $\bd_1$, in $\bD_1$ looks very similar to another basis, $\bd_2$, in $\bD_2$. When doing sparse coding, if one patch in class 1 uses $\bd_1$ for reconstruction, it is highly likely that a similar patch $\by$ in the same class uses $\bd_2$ for reconstruction instead. This misusage may lead to the case $\|\by -\bD_1 \delta_1(\hat{\bx})\| > \|\by - \bD_2 \delta_2(\hat{\bx})\|$, resulting in a misclassified patch. For this reason, the more discriminative bases are, the better the performance. }

\par {To formally verify this argument, we do one experiment on one normal and one inflammatory image from lung dataset in which the differences of DFDL bases and LCKSVD bases are most significant. From these images, patches are extracted, then their sparse codes are calculated using two dictionaries formed by DFDL bases and LC-KSVD bases. Fig. \ref{fig: visual_coef } demonstrates our results. Note that the plots in Figs.\ \ref{fig: visual_coef }c) and d) are corresponding to DFDL while those in Figs. \ref{fig: visual_coef }e) and f) are for LC-KSVD. Most of active coefficients in Fig. \ref{fig: visual_coef }c) are gathered on the left of the red line, and their values are also greater than values on the right. This means that $\bD_1$ contributes more to reconstructing the lung-normal image in Fig. \ref{fig: visual_coef }a) than $\bD_2$ does. Similarly, most of active coefficients in Fig. \ref{fig: visual_coef }d) locate on the right of the vertical line. This agrees with what we expect since the image in Fig. \ref{fig: visual_coef }a) belongs to class 1 and the one in Fig. \ref{fig: visual_coef }b) belongs to class 2. On the contrary, for LC-KSVD, active coefficients in Fig. \ref{fig: visual_coef }f) are more uniformly distributed on both sides of the red line, which adversely affects classification. In Fig. \ref{fig: visual_coef }e), although active coefficients are strongly concentrated to the left of the red line, this effect is even more pronounced with DFDL, i.e.\ in  Fig. \ref{fig: visual_coef }c).
}

\begin{figure}[t]
\centering
  \includegraphics[width=0.6\textwidth]{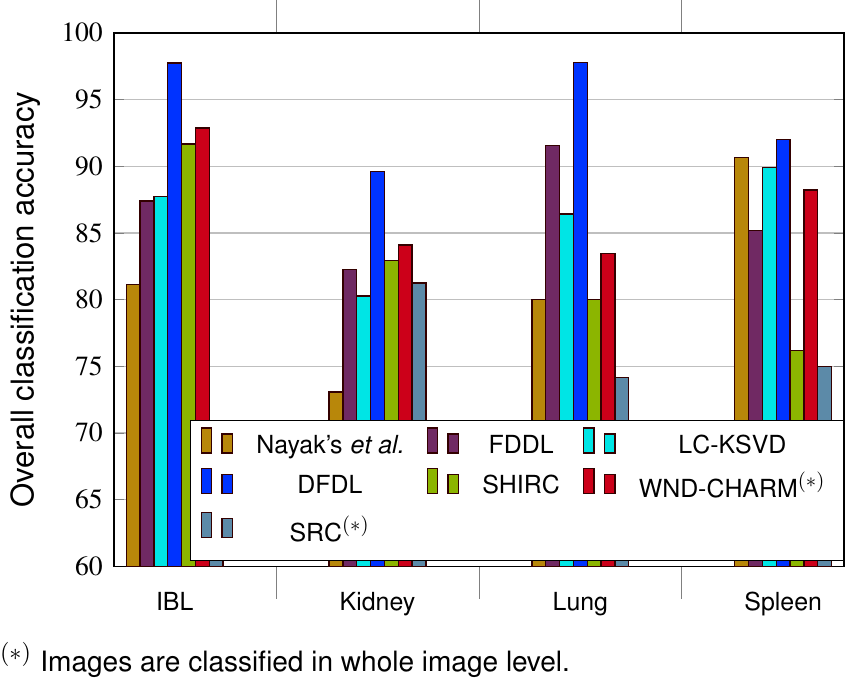}
 \caption{\small Bar graphs indicating the overall classification accuracies ($\%$) of the competing methods.}
\vspace{-0.15in}
  \label{fig: OverallAcc}
\end{figure}

\begin{figure}[t]
\centering
  \includegraphics[width=0.68\textwidth]{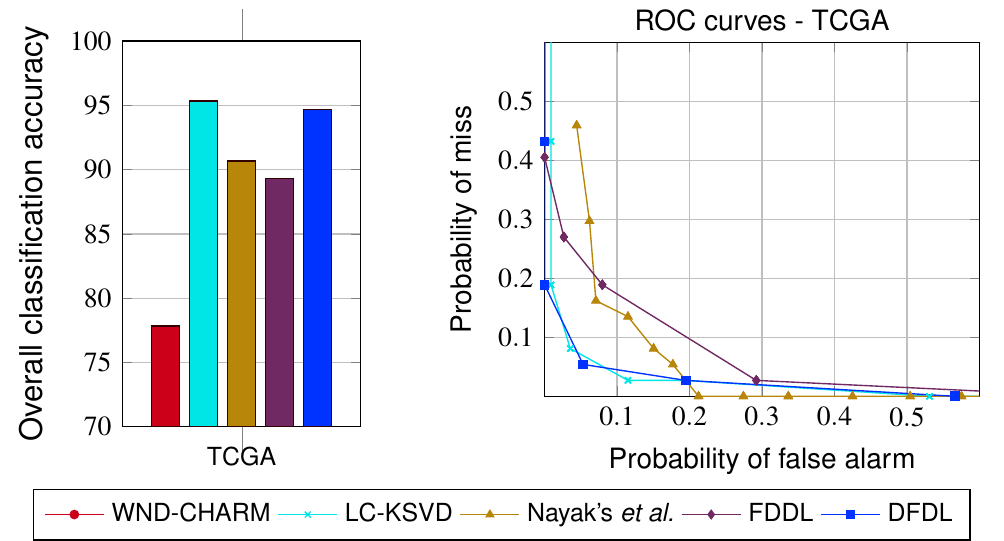}
 \caption{\small Bar graphs (left) indicating the overall classification accuracies ($\%$) and the receiver operating characteristic (right) of the competing methods for TCGA dataset.}
\vspace{-0.15in}
  \label{fig: OverallAccTCGA}
\end{figure}
\par

\subsection{Overall Classification Accuracy} 
\label{sub:overall_classification_accuracy}

 To verify the performance of our idea, for IBL and ADL datasets, we present overall classification accuracies in the form of bar graphs in Fig. \ref{fig: OverallAcc}. It is evident that DFDL outperforms other methods in both  datasets. Specifically, in IBL and ADL Lung, the overall classification accuracies of DFDL are over 97.75$\%$, the next best rates come from WND-CHARM (92.85$\%$ in IBL) and FDDL (91.56$\%$ in ADL-Lung), respectively, and much higher than those reported in \cite{Srinivas2014SHIRC} and our own previous results in \cite{vu2015dfdl}. It is noteworthy to mention here that the overall classification accuracy of MIL\cite{Dundar2011} applied to IBL is under 90$\%$. In addition, for ADL-Kidney and ADL-Spleen, our DFDL also provides the best result with accuracy rates being nearly 90$\%$ and over 92$\%$, respectively.


 For the TCGA dataset, overall accuracy of competing methods are shown in Fig. \ref{fig: OverallAccTCGA}, which reveals that DFDL performance is the second best, bettered only by LC-KSVD and by less than 0.67$\%$ (i.e.\ one more misclassified image for DFDL). 

\subsection{Complexity analysis} 
\label{sec:complexity_analysis}

\begin{table}[t]
\centering
\caption{ {Complexity analysis for different dictionary learning methods.}}
\label{tab:complexity}
  \begin{tabular}{|l|l|l|}
  \hline
  Method & Complexity & Running time \\ \hline
  DFDL &$c^2kN(2d + L^2)$  & $\sim$ 0.5 hours\\ \hline
  LC-KSVD\cite{Zhuolin2013LCKSVD} & $c^2kN(2d + 2ck + L^2)$& $\sim$ 3 hours \\ \hline
  Nayak's \etal\cite{Nandita2013}$^{(*)}$ &$c^2kN(2d + 2qck) + c^2dk^2$& $\sim$ 8 hours \\ \hline
  FDDL\cite{Meng2011FDDL}$^{(*)}$ & $c^2kN(2d + 2qck) + c^3dk^2$& $>$ 40 hours\\ \hline
  \end{tabular}
  \begin{tablenotes}
      \small
      \item $^{(*)}$$q$ is the number of iterations required for $l_1$-minimization in sparse coding step.
    \end{tablenotes}
\end{table}
\begin{table}[t]
\centering
\caption{Estimated number of operations required in different dictionary learning methods.}
\label{tab:operations}
  \begin{tabular}{|l|l|l|l|}
  \hline
  Method &  $q = 1$ &  $q = 3$        &              $q = 10$\\ \hline
  DFDL      & $6.6 \times 10^{10}$    & $6.6 \times 10^{10}$    & $6.6 \times 10^{10}$    \\ \hline
  LC-KSVD\cite{Zhuolin2013LCKSVD}  & $1.06 \times 10^{11}$   & $1.06 \times 10^{11}$   & $1.06 \times 10^{11}$   \\ \hline
  Nayak's \etal\cite{Nandita2013}     & $8.92\times 10^{10}$    & $1.692 \times 10^{11}$  & $4.492 \times 10^{11}$\\ \hline
  FDDL\cite{Meng2011FDDL}      & $9.04 \times 10^{10}$   & $1.704 \times 10^{11}$  & $4.504 \times 10^{11}$\\ \hline
  \end{tabular}
\end{table}
In this section, we compare the computational complexity for the proposed DFDL and competing dictionary learning methods: LC-KSVD\cite{Zhuolin2013LCKSVD}, FDDL\cite{Meng2011FDDL}, and Nayak's\cite{Nandita2013}. The complexity for each dictionary learning method is estimated as the (approximate) number of operations required by each method in learning the dictionary (see Appendix~\ref{Sec:appendixB} for details).
From Table \ref{tab:complexity}, it is clear that the proposed DFDL is the least expensive computationally. Note further, that the final column of Table \ref{tab:complexity} shows {\em actual run times} of each of the methods. The parameters were as follows: $c = 2$ (classes), $k = 500$ (bases per class), $N = 10,000$ (training patches per class), data dimension $d = 1200$ (3 channels $\times20\times20$), sparsity level $L = 30$. The run time numbers in the final column of Table \ref{tab:complexity} are in fact consistent with numbers provided in Table \ref{tab:operations}, which are calculated by plugging the above parameters into the second column of Table \ref{tab:complexity}.

\begin{table}[t]
\centering
\caption{ CONFUSION MATRIX: IBL. }
\label{tab: IBL_confusion}
\begin{tabular}{|c|c|c||l|}
\hline
Class                 & UDH            & DCIS           & Method  \\ \hline
\multirow{8}{*}{UDH}  & 91.75          & 8.25          & WND-CHARM$^{(*)}$ \cite{Shamir2008}   \\
                      & 68.00          & 32.00          & SRC$^{(*)}$     \cite{Wright2009SRC}     \\
                      & \textit{\textbf{93.33}}   & 6.67           & SHIRC    \cite{Srinivas2014SHIRC} \\
                      & 84.80          & 15.20          & FDDL    \cite{Meng2011FDDL}    \\
                      & 90.29          & 9.71           & LC-KSVD \cite{Zhuolin2013LCKSVD}     \\
                      & 85.71          & 14.29          & Nayak's \etal\cite{Nandita2013}   \\
                      & \textbf{96.00} & 4.00           & DFDL         \\ \hline
\multirow{8}{*}{DCIS} & 5.77           & \textit{\textbf{94.23}}          & WND-CHARM$^{(*)}$ \cite{Shamir2008}    \\
                      & 44.00          & 56.00          & SRC$^{(*)}$       \cite{Wright2009SRC}     \\
                      & 10.00          & 90.00          & SHIRC   \cite{Srinivas2014SHIRC} \\
                      & 10.00          & 90.00          & FDDL    \cite{Meng2011FDDL}        \\
                      & 14.86          & 85.14          & LC-KSVD   \cite{Zhuolin2013LCKSVD}      \\
                      & 23.43          & 76.57          & Nayak's \etal\cite{Nandita2013}      \\
                      & 0.50           & \textbf{99.50} & DFDL         \\ \hline
\end{tabular}
\begin{center}
  \begin{tablenotes}
        {\small      $^{(*)}$ Images are classified in whole image level.}
    \end{tablenotes}
\end{center}
\end{table}

\begin{table*}[ht!]

\centering
\caption{ CONFUSION MATRIX: ADL ($\%$).}
\label{tab: adl_confusion}
\begin{tabular}{|c||c|c||c|c||c|c||l|}
\hline
             & \multicolumn{2}{|c||}{Kidney}                       & \multicolumn{2}{|c||}{Lung}                        & \multicolumn{2}{|c||}{Spleen}                      &           \\ \hline
Class        & H.                 & I.            & H.                 & I.            & H.                 & I.            & \multicolumn{1}{c|}{Method}    \\ \hline \hline
\multirow{8}{*}{H.}
             & 83.27                   & 16.73                   & 83.20                   & 16.80                   & 87.23                   & 12.77                   & WND-CHARM$^{(*)}$ \cite{Shamir2008}   \\
             & \textit{\textbf{87.50}} & 12.50                   & 72.50                   & 27.50                   & 70.83                   & 29.17                   & SRC$^{(*)}$    \cite{Wright2009SRC}     \\
             & 82.50                   & 17.50                   & 75.00                   & 25.00                   & 65.00                   & 35.00                   & SHIRC    \cite{Srinivas2014SHIRC} \\
             & 83.26                   & 16.74                   & \textit{\textbf{93.15}}                   & 6.85                    & 86.94                   & 13.06                   & FDDL    \cite{Meng2011FDDL}    \\
             & 86.84                   & 13.16                   & 85.59                   & 15.41                   & \textit{\textbf{89.75}}                   & 10.25                   & LC-KSVD \cite{Zhuolin2013LCKSVD}     \\
             & 73.08                   & 26.92                   & 89.55                   & 10.45                   & 86.44                   & 13.56                   & Nayak's \etal \cite{Nandita2013}  \\
             & \textbf{88.21}          & 11.79                   & \textbf{96.52}          & 3.48                    & \textbf{92.88}          & 7.12                    & DFDL      \\ \hline
\multirow{8}{*}{I.}
             & 14.22                   & 85.78                   & 14.31                   & 83.69                   & 10.48                   & 89.52                   & WND-CHARM$^{(*)}$ \cite{Shamir2008}   \\
             & 25.00                   & 75.00                   & 24.17                   & 75.83                   & 20.83                   & 79.17                   & SRC$^{(*)}$    \cite{Wright2009SRC}     \\
             & 16.67                   & \textit{\textbf{83.33}}                   & 15.00                   & 85.00                   & 11.67                   & 88.33                   & SHIRC    \cite{Srinivas2014SHIRC} \\
             & 19.88                   & 80.12                   & 10.00                   & \textit{\textbf{90.00}}                   & 8.57                    & 91.43                        & FDDL    \cite{Meng2011FDDL}    \\
             & 19.25                   & 81.75                   & 10.89                   & 89.11                   & 8.57                    & 91.43                   & LC-KSVD \cite{Zhuolin2013LCKSVD}     \\
             & 26.92                   & 73.08                   & 25.90                   & 74.10                   & 6.05                    & \textbf{93.95}          & Nayak's \etal\cite{Nandita2013}   \\
             & 9.92                    &\textbf{90.02} & 2.57                    & \textbf{97.43}          & 7.89                    & \textit{\textbf{92.01}} & DFDL      \\ \hline
\end{tabular}
\begin{center}
\begin{tablenotes}
\centering
      {\small      $^{(*)}$ Images are classified in whole image level. H: Healthy, I: Inflammatory}
      \end{tablenotes}
\end{center}
\end{table*}

\subsection{Statistical Results: Confusion Matrices and ROC Curves} 
\label{subsec:detailed_results_confusion_matrices_and_roc_curves}



\begin{table}[t]
\centering
\caption{ CONFUSION MATRIX: TCGA ($\%$).}
\vspace{-0.1in}
\label{tab: tcga_confution}
\begin{tabular}{|c|c|c||l|}
\hline
     Class               & Not MVP & MVP   & Method    \\ \hline
\multirow{4}{*}{Not VMP} & 76.68   & 23.32 & WND-CHARM\cite{Shamir2008} \\
                         & 92.92   & 7.08  & Nayak's \etal\cite{Nandita2013}   \\
                         & \textbf{96.46}   & 3.54  & LC-KSVD\cite{Zhuolin2013LCKSVD}   \\
                         & {92.04}   & 7.96  & FDDL\cite{Meng2011FDDL}   \\
                         & \textit{\textbf{94.69}}   & 5.31  & DFDL      \\ \hline
\multirow{4}{*}{MVP}     & 21.62   & 78.38 & WND-CHARM\cite{Shamir2008} \\
                         & 16.22   & 83.78 & Nayak's \etal\cite{Nandita2013}   \\
                         & 8.10    & \textit{\textbf{91.90}} & LC-KSVD\cite{Zhuolin2013LCKSVD}   \\
                         & 18.92    & {81.08} & FDDL\cite{Meng2011FDDL}   \\
                         & 5.41   & {\textbf{94.59}} & DFDL      \\ \hline
\end{tabular}
\end{table}

\begin{figure*}[t]
\centering
  \includegraphics[width=\textwidth]{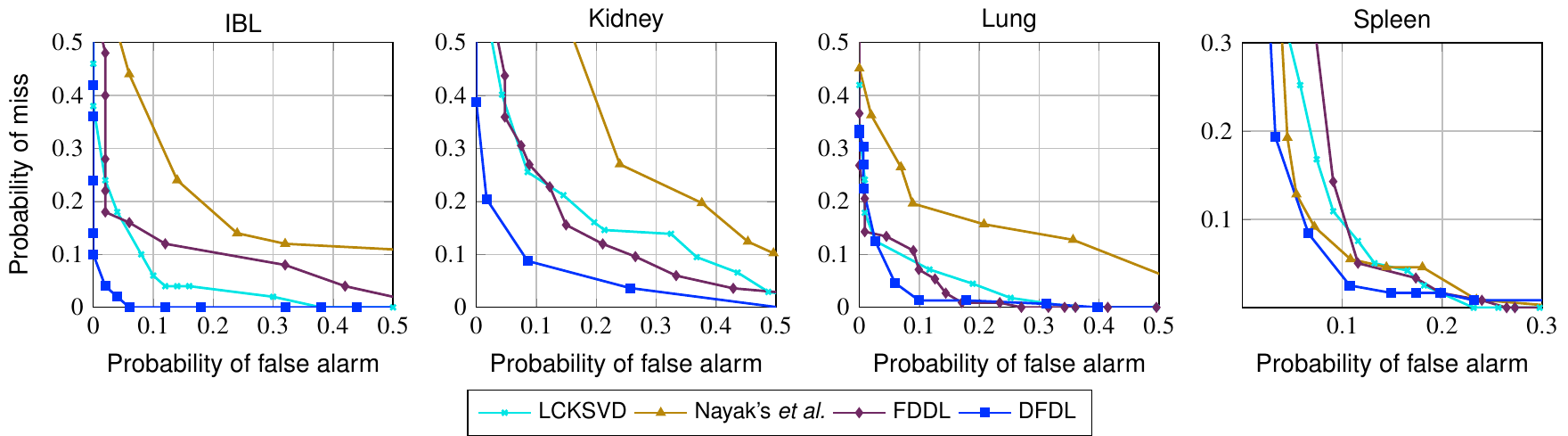}
 \caption{\small Receiver operating characteristic (ROC) curves for different organs, methods, and datasets (IBL and ADL).}
  \label{fig: ibladlroc}
\end{figure*}

\begin{figure*}[t]
\centering
  \includegraphics[width=\textwidth]{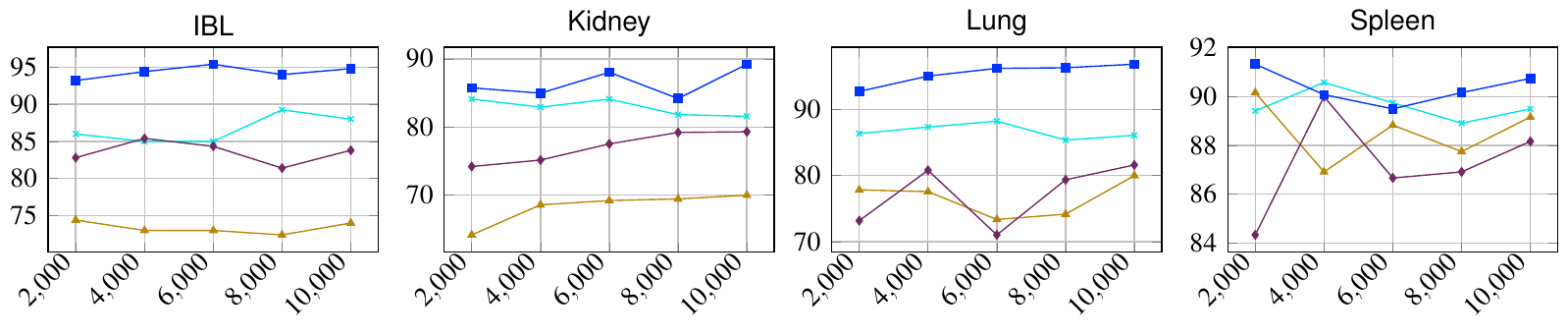}\\
  \includegraphics[width=\textwidth]{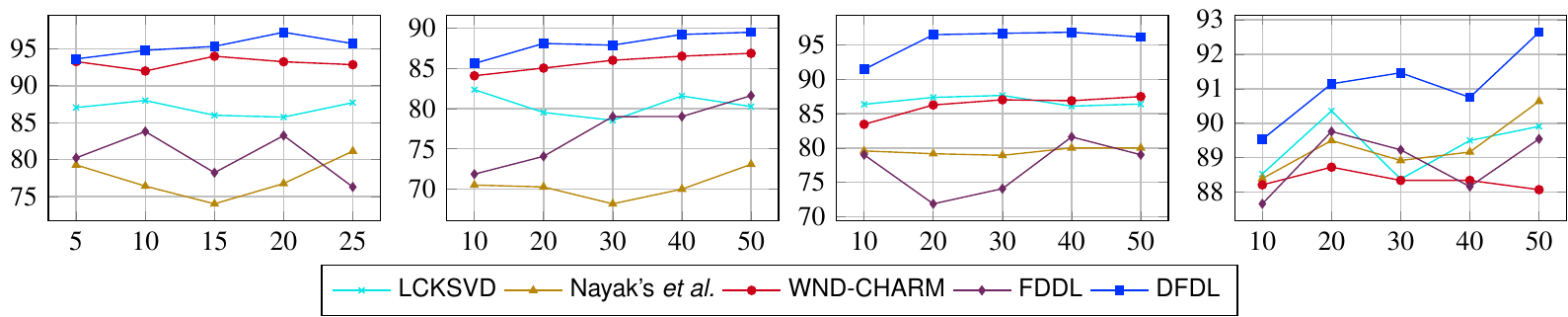}
 \caption{Overall classification accuracy ($\%$) as a function of training set size per class. Top row: number of training patches. Bottom row: number of training images.}
\vspace{-0.2in}
  \label{fig: CompareNimgs}
\end{figure*}
\begin{figure*}[t]
\centering
\includegraphics[width=\textwidth]{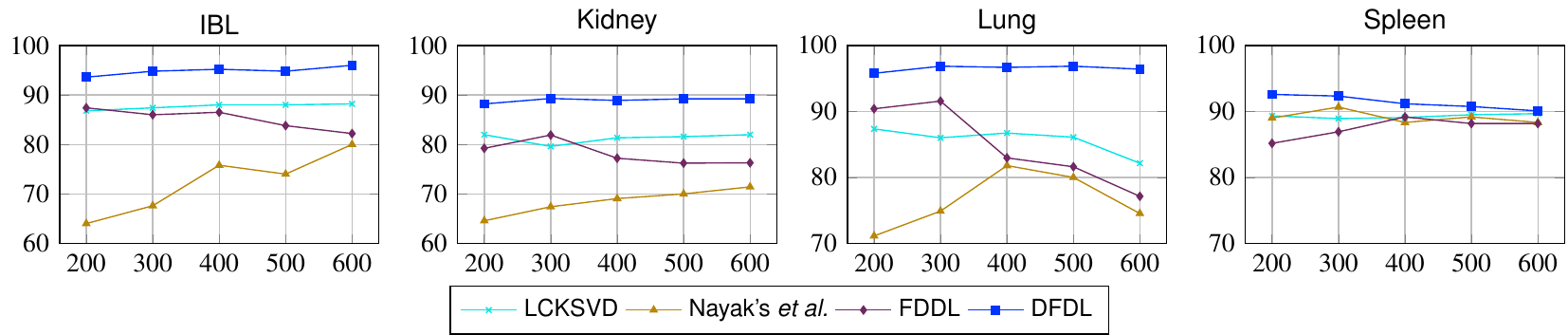}
 \caption{{Overall classification accuracy ($\%$) as a function of number of training bases.}}
  \label{fig: CompareNbases}
\end{figure*}

Next, we present a more elaborate interpretation of classification performance in the form of confusion matrices and ROC curves. Each row of a confusion matrix refers to the actual class identity of test images and each column indicates the classifier output. Table \ref{tab: IBL_confusion}, \ref{tab: adl_confusion} and \ref{tab: tcga_confution} show the mean confusion matrices for all of three dataset. In continuation of trends from Fig. \ref{fig: OverallAcc}, in Table \ref{tab: adl_confusion}, DFDL offers the best disease detection accuracy in almost all datasets for each organ, while maintaining high classification accuracy for healthy images.
\par
Typically in medical image classification problems, pathologists desire algorithms that reduce the probability of miss (diseased images are misclassified as healthy ones) while also ensuring that the false alarm rate remains low. However, there is a trade-off between these two quantities, conveniently described using receiver operating characteristic (ROC) curves. Fig. \ref{fig: ibladlroc} and Fig. \ref{fig: OverallAccTCGA} (right) show the ROC curves for all three datasets. The lowest curve (closest to the origin) has the best overall performance and the optimal operating point minimizes the sum of the miss and false alarm probabilities. It is evident that ROC curves for DFDL perform best in comparison to those of other state-of-the-art methods.
\par {\textbf{Remark:} Note for ROC comparisons, we compare the different flavors of dictionary learning methods (the proposed DFDL, LC-KSVD, FDDL and Nayak's), this is because as Table \ref{tab: tcga_confution} shows, they are the most competitive methods. Note for the IBL and ADL datasets, $\theta$, as defined in Fig. \ref{fig: ibladlprocedure}, is changed from 0 to 1 to acquire the curves; whereas for the TCGA dataset, number of connected classified-as-MVP patches, $m$, is changed from 1 to 20 to obtain the curves.} It is worth re-emphasizing that DFDL achieves these results even as its complexity is lower than competing methods.

\subsection{Performance vs. size of training set} 
\label{sec:performance_vs_size_of_training_set}
Real-world histopathological classification tasks must often contend with lack of availability of large training sets. To understand training dependence of the various techniques, we present a comparison of overall classification accuracy as a function of the training set size for the different methods. We also present a comparison of classification rates as a function of the number of training patches for different dictionary learning methods\footnote{Since WND-CHARM is applied in the whole image level, there is no result for it in comparison of training patches.}. In Fig. \ref{fig: CompareNimgs}, overall classification accuracy is reported for IBL and ADL datasets corresponding to five scenarios. It is readily apparent that DFDL exhibits the most graceful decline as training is reduced.
\subsection{{Performance vs. number of training bases}} 
\label{sub:performance_vs_number_of_training_bases}

{We now compare the behavior of each dictionary learning method as the number of bases in each dictionary varies from 200 to 600 (with patch size being fixed at $20\times20$ pixels). Results reported in Fig. \ref{fig: CompareNbases} confirm that DFDL again outperforms other methods. In general, overall accuracies of DFDL on different datasets remain high when we reduce number of training bases. Interpreted another way, these results illustrate that DFDL is fairly robust to changes in parameters, which is a highly desirable trait in practice.
}
\section{{Discussion and Conclusion}} 
\label{sec:Conclusion}
\par
{In this chapter, we address the histopathological image classification problem from a feature discovery and dictionary learning standpoint. This is a very important and challenging problem and the main challenge comes from the geometrical richness of tissue images, resulting in the difficulty of obtaining reliable discriminative features for classification. Therefore, developing a framework capable of capturing this structural richness and  being able to discriminate between different types is investigated and to this end, we propose the DFDL method which learns discriminative features for histopathology images. Our work aims to produce a more versatile histopathological image classification system through the design of discriminative, class-specific dictionaries which is hence capable of automatic feature discovery using example training image samples.
}\par
{Our DFDL algorithm learns these dictionaries by leveraging the idea of sparse representation of in-class and out-of-class samples. This idea leads to an optimization problem which encourages intra-class similarities and emphasizes the inter-class differences. Ultimately, the optimization in \eqref{subeqn:findD2} is done by solving the proposed equivalent optimization problem using a convexifying trick. Similar to other dictionary learning (machine learning approaches in general), DFDL also requires a set of regularization parameters. Our DFDL requires only one parameter, $\rho$, in its training process which is chosen by cross validation\cite{Kohavi95astudy} -- plugging different sets of parameters into the problem and selecting one which gives the best performance on the validation set. In the context of application of DFDL to real-world histopathological image slides, there are quite a few other settings should be carefully chosen, such as patch size, tiling method, number of connected components in the MVP detection etc. Of more importance is the patch size to be picked for each dataset which is mostly determined by consultation with the medical expert in the specific problem under investigation and the type of features that we should be looking for. For simplicity we employ regular tiling; however, using prior domain knowledge this may be improved. For instance in the context of MVP detection, informed selection of patch locations using existing disease detection and localization methods such as \cite{Mousavi2015JPI} can be used to further improve the detection of disease.
}\par
{Experiments are carried out on three diverse histopathological datasets to show the broad applicability of the proposed DFDL method. It is illustrated our method is competitive with or outperforms state of the art alternatives, particularly in the regime of realistic or limited training set size. It is also shown that with minimal parameter tuning and algorithmic changes, DFDL method can be easily applied on different problems with different natures which makes it a good candidate for automated medical diagnosis instead of using customized and problem specific frameworks for every single diagnosis task. We also make a software toolbox available to help deploy DFDL widely as a diagnostic tool in existing histopathological image analysis systems. Particular problems such as grading and detecting specific regions in histopathology may be investigated using our proposed techniques.
}\par
\chapter{Fast Low-rank Shared Dictionary Learning for Image Classification}
\label{chapter:contrib_lrsdl}
\def\M{\mathcal{M}}
\def\bG{\mathbf{G}}

\def\bbX{\lbar{\bX}}        
\def\bbx{\lbar{\bx}}        
\def\bbY{\lbar{\bY}}        
\def\bbD{\lbar{\bD}}  

\def\emt{\end{matrix}\right]}

\section{Introduction}
\label{sec:intro}

The central idea in SRC is to represent a test sample (e.g. a face) as a linear combination of samples from the available training set. Sparsity manifests because most of non-zeros correspond to bases whose memberships are the same as the test sample. Therefore, in the ideal case, each object is expected to lie in its own class subspace and all class subspaces are non-overlapping. Concretely, given $C$ classes and a dictionary $\bD = [\bD_1, \dots, \bD_C]$ with $\bD_c$ comprising training samples from class $c, c = 1, \dots, C$, a new sample $\by$ from class $c$ can be represented as $\by \approx \bD_c\bx^c$. Consequently, if we express $\by$ using the dictionary $\bD: \by \approx \bD\bx = \bD_1\bx^1 + \dots + \bD_c\bx^c + \dots + \bD_C\bx^C$, then most of active elements of $\bx$ should be located in $\bx^c$ and hence, the coefficient vector $\x$ is expected to be sparse. In matrix form, let $\bY = [\bY_1, \dots, \bY_c, \dots, \bY_C]$ be the set of all samples where $\bY_c$ comprises those in class $c$, the coefficient matrix $\bX$ would be sparse. In the ideal case, $\bX$ is block diagonal (see Figure \ref{fig:srcidea}).
\begin{figure}[t]
\centering
\includegraphics[width = 0.79\textwidth]{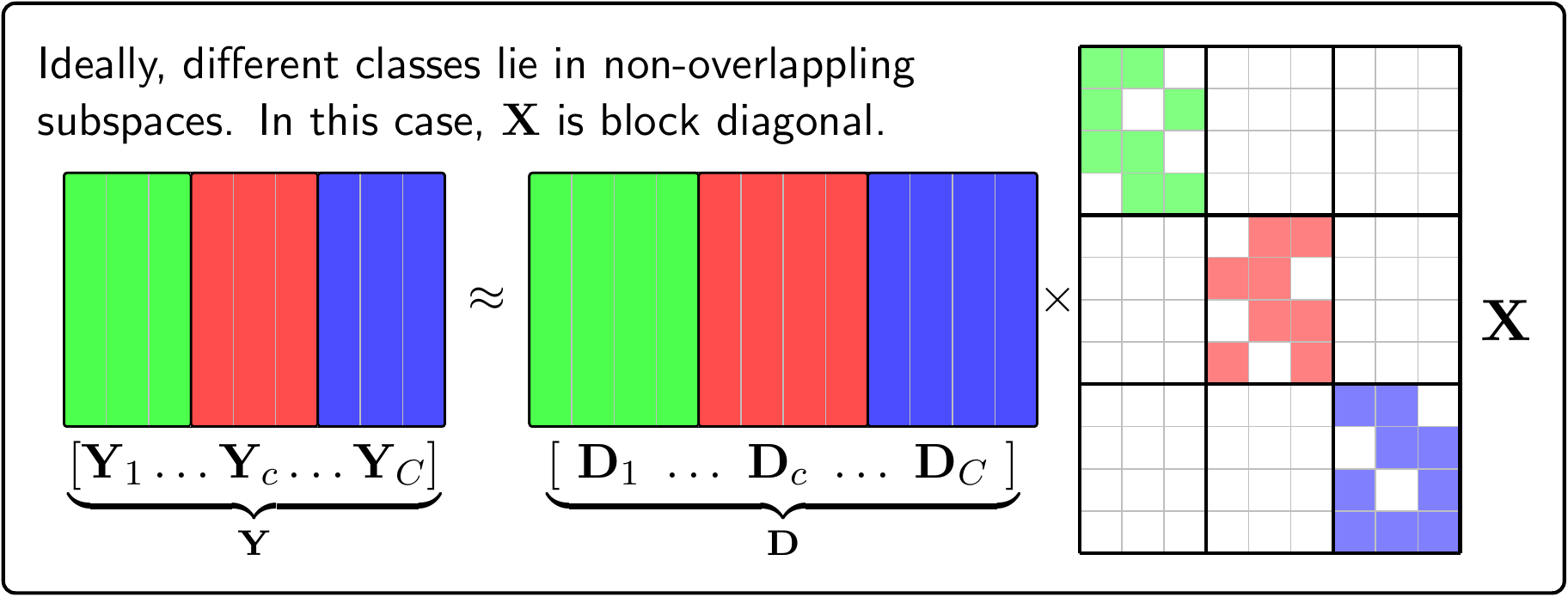}
\caption{\small Ideal structure of the coefficient matrix in SRC.}
\label{fig:srcidea}
\vspace{-.075in}
\end{figure}
\par
\vspace{-.3in}
\subsection{Closely Related work and Motivation} 
\vspace{-.1in}
\label{sub:subsection_name}
The assumption made by most discriminative dictionary learning methods, i.e.
non-overlapping subspaces, is unrealistic in practice. Often objects from
different classes share some common features, e.g. background in scene
classification. This problem has been partially addressed by recent efforts,
namely DLSI~\cite{ramirez2010classification}, COPAR~\cite{kong2012dictionary},
{JDL~\cite{zhou2014jointly}} and CSDL~\cite{gao2014learning}. However, DLSI does
not explicitly learn shared features since they are still hidden in the
sub-dictionaries. COPAR, JDL and CSDL explicitly learn a shared dictionary
$\bD_0$ but suffer from the following drawbacks. First, we contend that the
subspace spanned by columns of the shared dictionary must have low rank.
Otherwise, class-specific features may also get represented by the shared
dictionary. In the worst case, the shared dictionary span may include all
classes, greatly diminishing the classification ability. Second, the
coefficients (in each column of the sparse coefficient matrix) corresponding to
the shared dictionary should be similar. This implies that features are shared
between training samples from different classes via the ``shared dictionary''.
In this chapter, we develop a new low-rank shared dictionary learning framework
(LRSDL) which satisfies the aforementioned properties. Our framework is
basically a generalized version of the well-known FDDL
\cite{Meng2011FDDL,yang2014sparse} with the additional capability of capturing
shared features, resulting in better performance. We also show practical merits
of enforcing these constraints are significant.

The typical strategy in optimizing general dictionary learning problems is to
alternatively solve their subproblems where sparse coefficients $\bX$ are found
while fixing dictionary $\bD$ or vice versa. In discriminative dictionary
learning models, both $\bX$ and $\bD$ matrices furthermore comprise of several
small class-specific blocks constrained by complicated structures, usually
resulting in high computational complexity. Traditionally, $\bX$, and $\bD$ are
solved block-by-block until convergence. Particularly, each block $\bX_c$ (or
$\bD_c$ in dictionary update ) is solved by again fixing all other blocks
$\bX_i, i \neq c$ (or $\bD_i, i \neq c$). Although this greedy process leads to
a simple algorithm, it not only produces inaccurate solutions but also requires
huge computation. In this chapter, we aim to mitigate these drawbacks by proposing
efficient and accurate algorithms which allows to directly solve $\bX$ and $\bD$
in two fundamental discriminative dictionary learning methods: FDDL
\cite{yang2014sparse} and DLSI \cite{ramirez2010classification}. These
algorithms can also be applied to speed-up our proposed LRSDL, COPAR
\cite{kong2012dictionary}, $D^2L^2R^2$ \cite{li2014learning} and other related
works.
\begin{figure}[t]
\centering
\includegraphics[width = 0.79\textwidth]{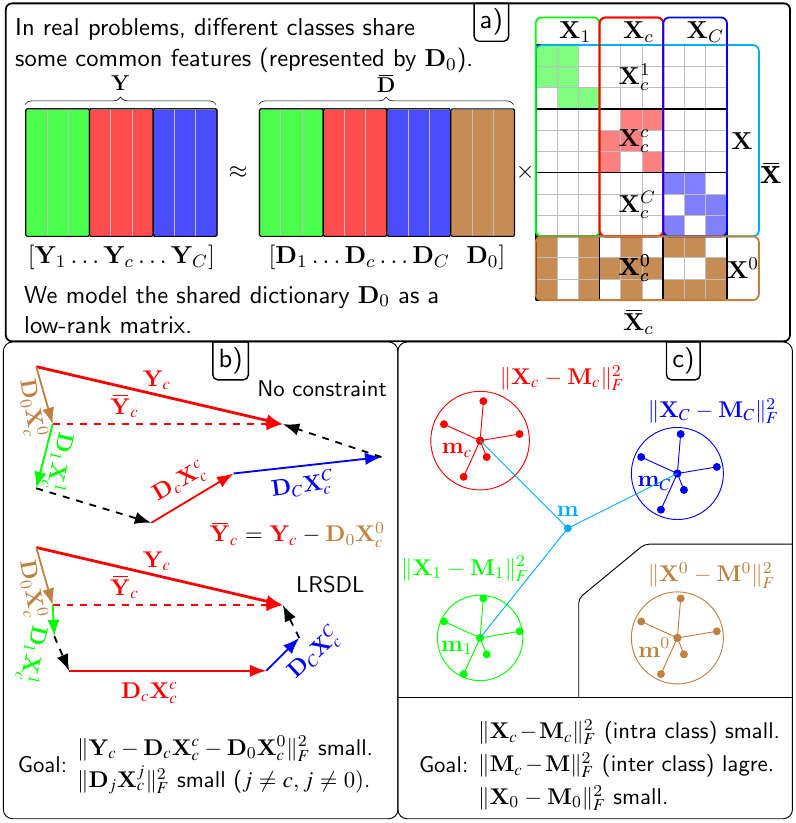}
\caption{\small {LRSDL idea with}: brown items -- shared; red, green, blue items
-- class-specific. a) Notation. b) The discriminative fidelity constraint: class-$c$ sample is mostly represented by $\bD_0$ and $\bD_c$. c) The Fisher-based discriminative coefficient constraint.}
\label{fig:lrsdl_idea}
\end{figure}

\section{Discriminative dictionary learning framework}
\label{sec:contribution}
\subsection{Notation} 
\label{sub:notaions}

In addition to notation stated in the Introduction, let $\bD_0$ be the shared dictionary, $\bI$ be the identity matrix with dimension inferred from context. For $c = 1, \dots, C$; $i = 0, 1, \dots, C$, suppose that $\bY_c \in \R^{d\times n_c}$ and $\bY \in \R^{d\times N}$ with $N = \sum_{c = 1}^C n_c$; $\bD_i \in \R^{d\times k_i}$, $\bD \in \R^{d\times K}$ with $K = \sum_{c=1}^C k_c$; and $\bX \in \R^{K\times N}$. Denote by $\bX^i$ the sparse coefficient of $\bY$ on $\bD_i$, by $\bX_c \in \R^{{K}\times N_c}$ the sparse coefficient of $\bY_c$ on ${\bD}$, by $\bX_c^i$ the sparse coefficient of $\bY_c$ on $\bD_i$. Let $\lbar{\bD} = \bmt\bD & \bD_0\emt$ be the total dictionary, $\lbar{\bX} = [\bX^T, (\bX^0)^T]^T$ and $\lbar{\bX}_c = [(\bX_c)^T, (\bX^0_c)^T ]^T$. For every dictionary learning problem, we implicitly constrain each basis to have its Euclidean norm no greater than 1. These variables are visualized in Figure \ref{fig:lrsdl_idea}a).
\par
Let $\bm, \bm^0,$ and $\bm_c$ be the mean of $\bX, \bX^0,$ and $\bX_c$ columns, respectively.
{Given a matrix $\mathbf{A}$ and a natural number $n$, define $\mu(\bA, n)$ as a matrix with $n$ same columns, each column being the mean vector of all columns of $\mathbf{A}$. If $n$ is ignored, we implicitly set $n$ as the number of columns of $\bA$.
 Let $\bM_c = \mu(\bX_c), \bM^0 = \mu(\bX^0)$, and $\bM = \mu(\bX, n)$ be the mean matrices. {The number of columns $n$ depends on context, e.g. by writing $\bM_c - \bM$, we mean that $n=n_c$} . The `mean vectors' are illustrated in Figure \ref{fig:lrsdl_idea}c).}

\par Given a function $f(A, B)$ with $A$ and $B$ being two sets of variables, define $f_A(B) = f(A, B)$ as a function of $B$ when the set of variables $A$ is fixed. Greek letters ($\lambda, \lambda_1, \lambda_2, \eta$) represent positive regularization parameters. Given a block matrix $\bA$, define a function $\M(\bA)$ as follows:
{ \begin{equation}
    \underbrace{\bmt
        \bA_{11} &  \dots & \bA_{1C}\\
        \bA_{21} &  \dots & \bA_{2C}\\
        \dots    &  \dots & \dots\\
        \bA_{C1} &  \dots & \bA_{CC}
        \emt}_{\bA}
    \mapsto \underbrace{\bA + \bmt
        \bA_{11} & \dots & \bzeros\\
        \bzeros  & \dots & \bzeros\\
        \dots    & \dots & \dots\\
        \bzeros  & \dots & \bA_{CC}
        \emt}_{\M(\bA)}.    
\end{equation}}%
\noindent That is, $\M(\bA)$ doubles diagonal blocks of $\bA$. The row and column partitions of $\bA$ are inferred from context. $\M(\bA)$ is a computationally inexpensive function of $\bA$ and will be widely used in our LRSDL algorithm and the toolbox.
\par
We also recall here  the FISTA algorithm \cite{beck2009fast} for solving the family of problems:
\begin{equation}
\label{eqn:fista}
    \bX = \arg\min_{\bX} h(\bX) + \lambda\|\bX\|_1,
\end{equation}
where $h(\bX)$ is convex, continuously differentiable with Lipschitz continuous gradient. FISTA is an iterative method which requires to calculate gradient of $h(\bX)$ at each iteration. In this chapter, we will focus on calculating the gradient of $h$.

\subsection{Closely related work: Fisher discrimination dictionary learning (FDDL)} 
\label{sub:fisher_discrimination_dictionary_learning}
FDDL \cite{Meng2011FDDL} has been used broadly as a technique for exploiting both structured dictionary and learning discriminative coefficient. Specifically, the discriminative dictionary $\bD$ and the sparse coefficient matrix $\bX$ are learned based on minimizing the following cost function:
\begin{eqnarray}
\label{eqn:fddl_cost_fn}
    J_{\bY}(\bD, \bX) = \frac{1}{2}f_{\bY}(\bD, \bX) + \lambda_1\|\bX\|_1 +
    \frac{\lambda_2}{2} g(\bX),
\end{eqnarray}
where  $\displaystyle f_{\bY}(\bD, \bX)  = \sum_{c=1}^C r_{\bY_c}(\bD, \bX_c)$ is the discriminative fidelity with:\\
\scalebox{.97}{\parbox{\linewidth}{%
\begin{equation*}
r_{\bY_c}(\bD, \bX_c) = \|\bY_c - \bD\bX_c\|_F^2 +  \|\bY_c - \bD_c\bX_c^c\|_F^2 + \sum_{j\neq c}\|\bD_j\bX^j_c\|_F^2,
\end{equation*}%
}}
$ g(\bX) = \sum_{c=1}^C (\|\bX_c - \bM_c\|_F^2 - \|\bM_c - \bM\|_F^2) + \|\bX\|_F^2$ is the Fisher-based discriminative coefficient term, and the $l_1$-norm encouraging the sparsity of coefficients.
{The last term in $r_{\bY_c}(\bD, \bX_c)$ means that $\bD_j$ has a  small contribution to the representation of $\bY_c$ for all $j \neq c$. With the last term $\|\bX\|_F^2$ in $g(\bX)$, the cost function becomes convex with respect to $\bX$.}

\begin{figure}[t]
\centering
\includegraphics[width = 0.79\textwidth]{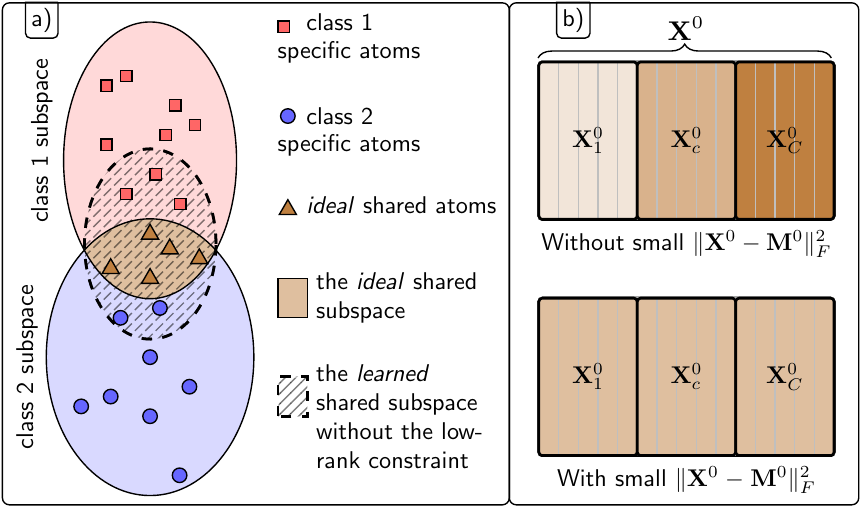}
\caption{\small Two constraints on the shared dictionary. a) Low-rank constraints. b) Similar shared codes.}
\label{fig:whylowrank}
\end{figure}

\vspace{-.2in}
{
\subsection{Proposed Low-rank shared dictionary learning (LRSDL) }
\label{sub:low_rand_shared_dictionary_learning_}
The shared dictionary needs to satisfy the following properties: \\
\textbf{\textit{1) Generativity:}}
As the common part, the most important property of the shared dictionary is to represent samples from all classes \cite{kong2012dictionary,zhou2014jointly,gao2014learning}. In other words, it is expected that $\bY_c$ can be well represented by the collaboration of the particular dictionary $\bD_c$ and the shared dictionary $\bD_0$. Concretely, the discriminative fidelity term $f_{\bY}(\bD, \bX)$ in (\ref{eqn:fddl_cost_fn}) can be extended to $\lbar{f}_{\bY}(\lbar{\bD}, \lbar{\bX}) = \sum_{c=1}^C \lbar{r}_{\bY_c} (\lbD, \lbX_c)$ with $\lbar{r}_{\bY_c} (\lbD, \lbX_c)$ being defined as:
\begin{equation*}
\|\bY_c - \lbar{\bD}\lbar{\bX}_c\|_F^2 + \|\bY_c - \bD_c\bX_c^c -
\bD_0\bX^0_c\|_F^2 + \sum_{j = 1, j\neq c}^C \|\bD_j\bX^j_c\|_F^2.
\end{equation*}
Note that since $\bar{r}_{\bY_c}(\lbD, \lbX_c) = r_{\bar{\bY}_c}(\bD, \bX_c)$ with $\lbar{\bY}_c = \bY_c - \bD_0\bX^0_c$ (see Figure \ref{fig:lrsdl_idea}b)), we have:
\begin{equation*}
\label{eqn:f_fbar}
\lbar{f}_{\bY}(\lbD, \lbX) = f_{\bar{\bY}}(\bD, \bX),
\end{equation*}
with $\lbar{\bY} = \bY - \bD_0\bX^0.$\\
This generativity property can also be seen in Figure \ref{fig:whylowrank}a). In this figure, the intersection of different subspaces, each representing one class, is one subspace visualized by the light brown region. One class subspace, for instance class 1, can be well represented by the \textit{ideal} shared atoms (dark brown triangles) and the corresponding class-specific atoms (red squares).}
\par 
{
\textbf{\textit{2) Low-rankness:}}
The stated generativity property  is only the necessary condition for a set of atoms to qualify a shared dictionary. Note that the set of atoms inside the shaded ellipse in Figure \ref{fig:whylowrank}a) also satisfies the generativity property: along with the \textit{remaining} red squares, these atoms well represent class 1 subspace; same can be observed for class 2. In the worst case, the set including all the atoms can also satisfy the generativity property, and in that undesirable case, there would be no discriminative features remaining in the class-specific dictionaries.  Low-rankness is hence {\em necessary} to prevent the shared dictionary from \textit{absorbing} discriminative atoms. The constraint is natural based on the observation that the subspace spanned by the shared dictionary has low dimension. Concretely, we use the nuclear norm regularization $\|\bD_0\|_*$, which is the convex relaxation of $\text{rank}(\bD_0)$ \cite{recht2010guaranteed}, to force the shared dictionary to be low-rank. In contrast with our work, existing approaches that employ shared dictionaries, i.e.\ COPAR~\cite{kong2012dictionary} and JDL~\cite{zhou2014jointly}, do not incorporate this crucial constraint.
}

\par 
{
\textbf{\textit{3) Code similarity:}}
In the classification step, a test sample $\by$ is decomposed into two parts: the part represented by the shared dictionary $\bD_0\bx^0$ and the part expressed by the remaining dictionary $\bD\bx$. Because $\bD_0\bx^0$ is not expected to contain class-specific features, it can be excluded before doing classification. The shared code $\bx^0$ can be considered as the contribution of the shared dictionary to the representation of $\by$. Even if the shared dictionary already has low-rank, its contributions to each class might be different as illustrated in Figure \ref{fig:whylowrank}b), the top row. In this case, the different contributions measured by $\bX^0$ convey class-specific features, which we aim to avoid. Naturally, the regularization term $\|\bX^0 - \bM^0\|$ is added to our proposed objective function to force each $\bx^0$ to be close to the mean vector $\bm^0$ of all $\bX^0$. \\
With this constraint, the Fisher-based discriminative coefficient term $g(\bX)$ is extended to $\lbar{g}(\lbar{\bX})$ defined as:
\begin{equation}
\label{eqn:deffbar}
\lbar{g}(\lbar{\bX}) = g(\bX) + \|\bX^0 -\bM^0\|_F^2,
\end{equation}
Altogether, the cost function $\lbar{J}_{\bY}(\lbar{\bD}, \lbar{\bX})$ of our proposed LRSDL is:
\begin{equation}
\label{eqn:lrsdl_cost_fn}
\lbar{J}_{\bY}(\lbD, \lbX) = \frac{1}{2}\lbar{f}_{\bY}(\lbar{\bD}, \lbar{\bX})
+ \lambda_1\|\lbar{\bX}\|_1 + \frac{\lambda_2}{2} \lbar{g}(\lbar{\bX})
+ \eta \|\bD^0\|_*.
\end{equation}
By minimizing this objective function, we can jointly find the class specific and shared dictionaries. Notice that if there is no shared dictionary $\bD_0$ (by setting $k_0 = 0$), then $\lbar{\bD}, \lbar{\bX}$ become $\bD, \bX$, respectively, $\lbar{J}_{\bY}(\lbar{\bD}, \lbar{\bX}) $ becomes $J_{\bY}(\bD, \bX)$ and LRSDL reduces to FDDL.
}

\par
\noindent \textbf{Classification scheme:}
\par
After the learning process, we obtain the total dictionary $\lbar{\bD}$ and mean vectors $\bm_c, \bm^0$. For a new test sample $\by$, first we find its coefficient vector $\lbar{\bx} = [\bx^T, (\bx^0)^T]^T$ with the sparsity constraint on $\lbar{\bx}$ and further encourage $\bx^0$ to be close to $\bm^0$:
\begin{equation}
\label{eqn:findcodebarx}
    \lbar{\bx} = \arg\min_{\lbar{\bx}} \frac{1}{2}\|\by - \lbar{\bD}\lbar{\bx}\|_2^2
    + \frac{\lambda_2}{2}\|\bx^0 - \bm^0\|_2^2 + \lambda_1\|\lbar{\bx}\|_1.
\end{equation}

Using $\lbar{\bx}$ as calculated above, we extract the contribution of the shared dictionary to obtain $\lbar{\by} = \by - \bD_0\bx^0$. The identity of $\by$ is determined by:
\begin{equation}
    \arg\min_{1 \leq c \leq C} (w\|\lbar{\by} - \bD_c\bx^c\|_2^2
    + (1-w)\|\bx - \bm_c\|_2^2),
\end{equation}
where $w \in [0,1]$ is a preset weight for balancing the contribution of the two terms.
\subsection{Efficient solutions for optimization problems} 
\label{sub:solving_the_opt}
Before diving into minimizing the LRSDL objective function in (\ref{eqn:lrsdl_cost_fn}),
 we first present efficient algorithms for minimizing the FDDL objective
 function in (\ref{eqn:fddl_cost_fn}).

\subsubsection{Efficient FDDL dictionary update} 
\label{ssub:fddl_dictionary_update}
\par
Recall that in \cite{Meng2011FDDL}, the dictionary update step is divided into $C$ subproblems, each updates one class-specific dictionary $\bD_c$ while others fixed. This process is repeated until convergence. This approach is not only highly time consuming but also inaccurate. We will see this in a small example presented in Section \ref{sub:valid_eff_algs}. We refer this original FDDL dictionary update as O-FDDL-D.

\par We propose here an efficient algorithm for updating dictionary called E-FDDL-D where the \textit{total dictionary} $\bD$ will be optimized when $\bX$ is fixed, significantly reducing the computational cost.
\par {Concretely, when we fix $\bX$ in equation \eqref{eqn:fddl_cost_fn}, the problem of solving $\bD$ becomes}:
\begin{equation}
\label{eqn:fddl_updateD}
    \bD = \arg\min_{\bD}f_{\bY, \bX}(\bD)
\end{equation}
\vspace{-.1in}
Therefore, $\bD$ can be solved by using the following lemma.
\begin{theorem}
\label{lem:fddl_updateD}
\textit{The optimization problem (\ref{eqn:fddl_updateD}) is equivalent to: }
\begin{equation}
    \label{eqn:efddl_d_odl}
    \bD = \arg\min_{\bD}\{ -2\trace(\bE\bD^T) + \trace(\Fb\bD^T\bD) \},
\end{equation}
\textit{where $\bE = \bY\M(\bX^T)$ and $\Fb = \M(\bX\bX^T)$. }
\end{theorem}
\textit{Proof:} See Appendix \ref{apd:proof_fddl_updateD}.
\par
The problem \eqref{eqn:efddl_d_odl} can be solved effectively by Online Dictionary Learning (ODL) method
\cite{mairal2010online}.

\subsubsection{Efficient FDDL sparse coefficient update (E-FDDL-X)} 
\label{ssub:eff_fddl_sparse_coefficient_update}
When $\bD$ is fixed, $\bX$ will be found by solving:
\begin{equation}
\label{eqn:fddl_updateX}
    \bX = \arg\min_{\bX} h(\bX)    + \lambda_1 \|\bX\|_1,
\end{equation}
where $h(\bX) = \frac{1}{2} f_{\bY, \bD}(\bX)  + \frac{\lambda_2}{2}g(\bX)$.
The problem (\ref{eqn:fddl_updateX}) has the form of equation (\ref{eqn:fista}), and can hence be solved by FISTA \cite{beck2009fast}. We need to calculate gradient of
$f(\bullet)$ and $g(\bullet)$ with respect to $\bX$.
\begin{theorem}
\label{lem:fddl_updateX}
{\textit{ Calculating gradient of $h(\bX)$ in equation \eqref{lem:fddl_updateX} }}
    \begin{eqnarray}
    \frac{\partial \frac{1}{2}f_{\bY, \bD}(\bX)}{\partial \bX} =& \M(\bD^T\bD) \bX - \M(\bD^T\bY), \\
     \frac{\partial\frac{1}{2}g(\bX)}{\partial(\bX)} =& 2\bX + \bM -
    2\underbrace{\bmt \bM_1 & \bM_2 & \dots \bM_c\emt}_{\widehat{\bM}}.
\end{eqnarray}
    Then we obtain:
{\small \begin{equation}
    \label{eqn:fddl_dif_x}
    \frac{\partial h(\bX)}{\partial \bX} =
        (\M(\bD^T\bD) + 2\lambda_2 \bI) \bX - \M(\bD^T\bY) + \lambda_2(\bM - 2\widehat{\bM}).
\end{equation}}
\end{theorem}
\textit{Proof:} See Appendix \ref{apd:proof_fddl_updateX}.
\par {Since the proposed LRSDL is an extension of FDDL, we can also extend these two above algorithms to optimize LRSDL cost function as follows. }

\subsubsection{LRSDL dictionary update (LRSDL-D)} 
\label{ssub:lrsdl_dictionary_update}
Returning to our proposed LRSDL problem, we need to find $\lbD = [\bD,\bD_0]$ when $\lbX$ is fixed. We propose a method to solve $\bD$ and $\bD_0$ separately.
\par
\textbf{For updating $\bD$}, recall the observation that $\lbar{f}_{\bY}(\lbD, \lbX) = f_{\lbar{\bY}}(\bD, \bX)$, with $\lbY \triangleq \bY - \bD_0\bX^0$ (see equation \eqref{eqn:f_fbar}), and the E-FDDL-D presented in section \ref{ssub:fddl_dictionary_update}, we have:
\begin{equation}
    \bD = \arg\min_{\bD}\{-2\trace(\bE\bD^T) + \trace(\Fb\bD^T\bD)\},
\end{equation}
with $\bE = \lbY\M(\bX^T)$ and $\Fb = \M(\bX\bX^T).$
\par
\textbf{For updating $\bD_0$}, we use the following lemma:
\begin{theorem}
\label{lem:lrsdld0x0}
{\textit{When $\bD, \bX$ in \eqref{eqn:lrsdl_cost_fn} are fixed,}}
\begin{eqnarray}
    \nonumber
    \lbar{J}_{\bY, \bD, \bX}(\bD_0, \bX_0) &=& \|\bV - \bD_0\bX_0\|_F^2 + \frac{\lambda_2}{2}\|\bX^0 - \bM^0\|_F^2 + \\
    \label{eqn:lrsdl_updateD0}
    && + \eta\|\bD_0\|_* + \lambda_1\|\bX^0\|_1 + \text{constant},
\end{eqnarray}
where $\bV = \bY - \frac{1}{2}\bD\M(\bX)$.
\end{theorem}
\textit{Proof:} See Appendix \ref{apd:proof_lrsdl_updateD0X0}.
\par
Based on the Lemma \ref{lem:lrsdld0x0}, $\bD_0$ can be updated by solving:
\begin{align}
\nonumber
    &\bD_0 = \arg\min_{\bD_0} \trace{(\Fb\bD_0^T\bD_0)} -2\trace(\bE\bD_0^T) +  \eta\|\bD_0\|_* \\
    \label{eqn:lrsdl_d0_form_EF}
    &\text{where:}~~\bE = \bV(\bX^0)^T; \quad\quad\Fb = \bX^0(\bX^0)^T
\end{align}
using the ADMM \cite{boyd2011distributed} method and the singular value thresholding algorithm \cite{cai2010singular}. The ADMM procedure is as follows. First, we choose a positive $\rho$, initialize $\bZ= \bU = \bD_0$,
then alternatively solve each of the following subproblems until convergence:
\begin{align}
\label{eqn:lrsdl_d0_1}
    \bD_0 =& \arg\min_{\bD_0} -2\trace(\lbar{\bE}\bD_0^T) + \trace\left(\lbar{\Fb}\bD_0^T\bD_0\right),\\
\label{eqn:lrsdl_d0_2}
     & \text{with~~} \lbar{\bE} = \bE + \frac{\rho}{2} (\bZ - \bU); ~ \lbar{\Fb} = \Fb + \frac{\rho}{2} \bI,\\
\label{eqn:lrsdl_d0_3}
    \bZ =& \mathcal{D}_{\eta/\rho}(\bD_c + \bU),\\
\label{eqn:lrsdl_d0_4}
    \bU =& \bU + \bD_0 - \bZ,
\end{align}

where $\mathcal{D}$ is the shrinkage thresholding operator \cite{cai2010singular}. {The optimization problem \eqref{eqn:lrsdl_d0_1} can be solved by ODL \cite{mairal2010online}. Note that \eqref{eqn:lrsdl_d0_2} and \eqref{eqn:lrsdl_d0_4} are computationally inexpensive. }


\vspace{-.1in}
\subsubsection{LRSDL sparse coefficients update (LRSDL-X)} 
\label{sub:sparse_coding_update}
In our  preliminary work \cite{vu2016icip}, we proposed a method for effectively solving $\bX$ and
$\bX^0$ alternatively, now we combine both problems into one and find
$\lbX$ by solving the following optimization problem:
\begin{equation}
    \lbX = \arg\min_{\lbX} \lbar{h}(\lbX) + \lambda_1\|\lbX\|_1.
\end{equation}
where $\displaystyle \lbar{h}(\lbX) = \frac{1}{2}\lbar{f}_{\bY, \overline{\bD}}(\lbX) + \frac{\lambda_2}{2} \lbar{g}(\lbX)$.
We again solve this problem using FISTA \cite{beck2009fast} with the gradient of $\displaystyle \lbar{h}(\lbX)$:
\begin{equation}
\displaystyle
    \label{eqn:hbar_diff_both}
    \nabla \lbar{h}(\lbX) = 
    \bmt \displaystyle \frac{\partial \lbar{h}_{\bX^0}(\bX)}{\partial \bX}\\
         \displaystyle \frac{\partial \lbar{h}_{\bX}(\bX^0)}{\partial \bX^0} \emt.
\end{equation}
\textit{For the upper term}, by combining the observation
\begin{eqnarray}
\nonumber
 \lbar{h}_{\bX^0}(\bX) &=& \frac{1}{2}\lbar{f}_{\bY, \overline{\bD}, \bX^0}(\bX) +
                            \frac{\lambda_2}{2} \lbar{g}_{\bX^0}(\bX) , \\
 &=& \frac{1}{2}f_{\lbar{\bY}, \bD}(\bX) + \frac{\lambda_2}{2}g(\bX) + \text{constant},
\end{eqnarray}
  and using equation, we obtain:
    \begin{equation}
    \label{eqn:hbar_diff_up}
        \displaystyle \frac{\partial \lbar{h}_{\bX^0}(\bX)}{\partial \bX} =
                (\M(\bD^T\bD) + 2\lambda_2 \bI) \bX - \M(\bD^T\lbY) + \lambda_2(\bM - 2\widehat{\bM}).
    \end{equation}

\textit{For the lower term}, by using Lemma \ref{lem:lrsdld0x0}, we have:
    \begin{equation}
        \lbar{h}_{\bX}(\bX^0) = \|\bV - \bD_0\bX^0\|_F^2 + \frac{\lambda_2}{2}\|\bX^0 - \bM^0\|_F^2 + \text{constant}.
    \end{equation}
    \begin{eqnarray}
        \nonumber
         \imply \frac{\partial \lbar{h}_{\bX}(\bX^0)}{\partial \bX^0} =
         2\bD_0^T\bD_0\bX^0 - 2\bD_0^T\bV + \lambda_2(\bX^0 - \bM^0),\\
         \label{eqn:hbar_diff_lo}
         = (2\bD_0^T\bD_0 + \lambda_2\bI)\bX^0 - 2\bD_0^T\bV -\lambda_2\bM^0.
    \end{eqnarray}

{By combining these two terms, we can calculate \eqref{eqn:hbar_diff_both}.}
\par 
{
Having $\nabla \lbar{h}(\lbX)$ calculated, we can update $\lbX$ by the FISTA algorithm \cite{beck2009fast} as given in Algorithm 1. Note that we need to compute a Lipschitz coefficient $L$ of $\nabla \lbar{h}(\lbX)$. 
The overall algorithm of LRSDL is given in Algorithm 2.
}
\begin{algorithm}[t]
\label{alg:LRSDLX}
    \caption{{LRSDL sparse coefficients update by FISTA\cite{beck2009fast}}}
    \begin{spacing}{1.3}
    \begin{algorithmic}
    \Function {$(\hat{\bX}, \hat{\bX}^0)$ = LRSDL\_X}{$\bY, \bD, \bD_0, \bX, \bX^0, \lambda_1, \lambda_2$}.
    \State 1. Calculate:
    \begin{align*}
        \mathbf{A} &= \mathcal{M}(\bD^T\bD) + 2\lambda_2 \mathbf{I}; \\
        \mathbf{B} &= 2\bD_0^T\bD_0 + \lambda_2\mathbf{I} \\
        L &= \lambda_{\max}(\mathbf{A}) + \lambda_{\max}({\mathbf{B}}) + 4\lambda_2 + 1 \text{\footnotemark}
    \end{align*}
    \State 2. Initialize $\bW_1 = \bZ_0 = \bmt \bX \\ \bX^0\emt, t_1 = 1, k = 1$
    \While {\text{not convergence and $ k < k_{\max}$}}
        \State 3. Extract $\bX, \bX^0$ from $\bW_k$.
        \State 4. Calculate gradient of two parts:
        \begin{eqnarray*}
            \bM &=& \mu(\bX), \bM_c = \mu(\bX_c), \widehat{\bM} = [\bM_1, \dots, \bM_C].\\
            \bV &=& \bY - \frac{1}{2}\bD\mathcal{M}(\mathbf{X})\\
            \bG &=& \bmt
                \bA\bX - \mathcal{M}(\bD^T(\bY - \bD_0\bX^0)) + \lambda_2(\bM - \widehat{\bM})\\
                \mathbf{B}\bX^0 - \bD_0^T\bV -\lambda_2\mu(\bX^0)
            \emt
        \end{eqnarray*}
        \State 5. $\bZ_k = \mathcal{S}_{\lambda_1/L}\left(\bW_k - \bG/L\right)$ ($\mathcal{S}_{\alpha}()$ is the element-wise soft thresholding function. $\mathcal{S}_{\alpha}(x) = \text{sgn}(x)(|x| - \alpha)_+$).

        \State 6. $t_{k+1} = (1 + \sqrt{1 + 4t_k^2})/2$

        \State 7. $\bW_{k+1} = \bZ_k + \frac{t_k - 1}{t_{k+1}} (\bZ_k - \bZ_{k-1})$
        \State 8. $k = k + 1$
    \EndWhile
    \State 9. OUTPUT: Extract $\bX, \bX^0$ from $\bZ_k$.
    \EndFunction
    \end{algorithmic}
    \end{spacing}
\end{algorithm}
\footnotetext{In our experiments, we practically choose this value as an upper bound of the Lipschitz constant of the gradient.}

\begin{algorithm}[t!]
    \caption{{LRSDL algorithm}}
    \begin{spacing}{1.2}
    \begin{algorithmic}

    \Function {$(\hat{\bX}, \hat{\bX}^0)$ = LRSDL}{$\bY, \lambda_1, \lambda_2, \eta$}.
    \State 1. Initialization $\bX = \mathbf{0}$, and:
    \begin{eqnarray*}
        (\bD_c, \bX_c^c) &=&\arg\min_{\bD, \bX} \frac{1}{2} \|\bY_c - \bD\bX\|_F^2 + \lambda_1 \|\bX\|_1\\
        (\bD_0, \bX^0) &=&\arg\min_{\bD, \bX} \frac{1}{2} \|\bY - \bD\bX\|_F^2 + \lambda_1 \|\bX\|_1
    \end{eqnarray*}
    \While {not converge}
        \State 2. Update $\bX$ and $\bX^0$ by Algorithm 1.
        \State 3. Update $\bD$ by ODL \cite{mairal2010online}:
        \begin{eqnarray*}
            \mathbf{E} &=& (\bY - \bD_0\bX^0)\mathcal{M}(\bX^T)\\
            \Fb &=& \mathcal{M}(\bX\bX^T) \\
            \bD &=& \arg\min_{\bD}\{-2\trace(\mathbf{E}\bD^T) + \trace(\Fb\bD^T\bD)\}
        \end{eqnarray*}
        \State 4. Update $\bD_0$ by ODL \cite{mairal2010online} and ADMM \cite{boyd2011distributed} (see equations \eqref{eqn:lrsdl_d0_1} - \eqref{eqn:lrsdl_d0_4}).
    \EndWhile

    \EndFunction

    \end{algorithmic}
    \end{spacing}
\end{algorithm}

\subsection{Efficient solutions for other dictionary learning methods} 
\vspace{-.1in}
\label{sub:edlsi}
We also propose here another efficient algorithm for updating dictionary in
two other well-known dictionary learning methods: DLSI \cite{ramirez2010classification}
and COPAR \cite{kong2012dictionary}. 
\par
The cost function $J_1(\bD, \bX)$
in DLSI is defined as:
\begin{equation}
\label{eqn:dlsi_cost_function}
    \sum_{c = 1}^C\big(||\bY_c - \bD_c \bX^c\|_F^2 +
    \lambda\|\bX^c\|_1 +       \frac{\eta}{2}\sum_{j=1,j \neq c}^C \|\bD_j^T\bD_c\|_F^2   \big)
\end{equation}
Each class-specific dictionary $\bD_c$ is updated by fixing others and solve:
\begin{equation}
\label{eqn:dlsi_updateDc}
    \bD_c = \arg\min_{\bD_c} \|\bY_c - \bD_c\bX^c\|_F^2 + \eta\|\bA\bD_c\|_F^2,
\end{equation}
with $\bA = \bmt \bD_1, \dots , \bD_{c-1}, \bD_{c+1}, \dots, \bD_{C}\emt^T$.
\par
The original solution for this problem, which will be referred as O-FDDL-D,
updates each column $\bd_{c,j}$ of $\bD_c$ one by one based on the procedure:
\begin{align}
    \label{eqn:odlsid_step1}
    \bu &= (\|\bx_c^j\|_2^2 \bI + \eta\bA^T\bA)^{-1}(\bY_c - \displaystyle\sum_{i \neq j}\bd_{c, i}\bx_c^i)\bx_c^j,\\
    \label{eqn:odlsid_step2}
    \bd_{c, j} &= \displaystyle \bu/\|\bu\|_2^2,
\end{align}

where $\bd_{c, i}$ is the $i$-th column of $\bD_c$ and $\bx_c^j$ is the $j$-th row of $\bX_c$. This algorithm is highly computational since it requires one matrix inversion for \textit{each} of $k_c$ columns of $\bD_c$. We propose one ADMM \cite{boyd2011distributed} procedure to update $\bD_c$ which requires \textit{only one} matrix inversion, which will be referred as E-DLSI-D. First, by letting $\bE = \bY_c(\bX^c)^T$ and $\Fb = \bX^c(\bX^c)^T$, we rewrite (\ref{eqn:dlsi_updateDc}) in a more general form:
\begin{equation}
    \label{eqn:dlsi_updateDc_admm_form}
    \bD_c = \displaystyle\arg\min_{\bD_c}\trace(\Fb\bD_c^T\bD_c) -2\trace(\bE\bD_c^T) + \eta \|\bA\bD_c\|_F^2.
\end{equation}

In order to solve this problem, first, we choose a $\rho$, let $\bZ= \bU = \bD_c$,
then alternatively solve each of the following sub problems until convergence:
\begin{align}
\label{eqn:edlsid_firststep}
    \bD_c =& \arg\min_{\bD_c} -2\trace(\lbar{\bE}\bD_c^T) + \trace\left(\lbar{\Fb}\bD_c^T\bD_c\right),\\
\label{eqn:edlsid_2}
     & \text{with~~} \lbar{\bE} = \bE + \frac{\rho}{2} (\bZ - \bU); ~ \lbar{\Fb} = \Fb + \frac{\rho}{2} \bI.\\
\label{eqn:edlsid_3}
    \bZ =& {(2\eta \bA^T\bA + \rho \bI)^{-1}}(\bD_c + \bU). \\
\label{eqn:edlsid_laststep}
    \bU =& \bU + \bD_c - \bZ.
\end{align}
This efficient algorithm requires only one matrix inversion. Later in this chapter,
we will both theoretically and experimentally show that E-DLSI-D is much more
efficient than O-DLSI-D \cite{ramirez2010classification}. Note that this algorithm can be beneficial for two
subproblems of updating the common dictionary and the particular dictionary
in COPAR \cite{kong2012dictionary} as well.
\section{Complexity analysis} 
\label{sec:complexity_analysis}
We compare the computational complexity for the efficient algorithms and their corresponding original algorithms. We also evaluate the total complexity of the proposed LRSDL and competing dictionary learning methods: DLSI \cite{ramirez2010classification}, COPAR \cite{kong2012dictionary} and FDDL \cite{Meng2011FDDL}. The complexity for each algorithm is estimated as the (approximate) number of multiplications required for one iteration (sparse code update and dictionary update). For simplicity, we assume: i) number of training samples, number of dictionary bases in each class (and the shared class) are the same, which means: $n_c = n, k_i = k$. ii) The number of bases in each dictionary is comparable to number of training samples per class and much less than the signal dimension, i.e. $k \approx n \ll d$. iii) Each iterative algorithm requires $q$ iterations to convergence. For consistency, we have changed notations in those methods by denoting $\bY$ as training sample and $\bX$ as the sparse code.

\par
In the following analysis, we use the fact that: i) if $\bA \in \R^{m\times n}, \bB \in \R^{n \times p}$, then the matrix multiplication $\bA\bB$ has complexity $mnp$. ii) If $\bA \in \R^{n \times n}$ is nonsingular, then the matrix inversion $\bA^{-1}$ has complexity $n^3$. iii) The singular value decomposition of a matrix $\bA\in \R^{p \times q}$, $p > q$, is assumed to have complexity $O(pq^2)$.
\subsection{Online Dictionary Learning (ODL)} 
\vspace{-.1in}
\label{sub:online_dictionary_learning}
We start with the well-known Online Dictionary Learning \cite{mairal2010online} whose cost function is:
\begin{equation}
    J(\bD, \bX) = \frac{1}{2} \|\bY - \bD\bX\|_F^2 + \lambda\|\bX\|_1.
\end{equation}
where $\bY \in \R^{d \times n}, \bD \in \R^{d \times k}, \bX \in \R^{k \times n}$.
Most of dictionary learning methods find their solutions by alternatively solving one variable while fixing others. There are two subproblems:
\subsubsection{Update $\bX$ (ODL-X)}
\label{ssub:odl_x}%
When the dictionary $\bD$ is fixed, the sparse coefficient $\bX$ is updated by solving the problem:
\begin{equation}
\label{eqn:l1norm}
    \bX = \arg\min_{\bX} \frac{1}{2} \|\bY - \bD\bX\|_F^2 + \lambda \|\bX\|_1
\end{equation}
using FISTA \cite{beck2009fast}. In each of $q$ iterations, the most computational task is to compute $\bD^T\bD \bX - \bD^T\bY$ where $\bD^T\bD$ and $\bD^T\bY$ are precomputed with complexities $k^2d$ and $kdn$, respectively. The matrix multiplication $(\bD^T\bD)\bX$ has complexity $k^2n$. Then, the total complexity of ODL-X is:
\begin{equation}
\label{eqn:cplxt_l1norm}
    k^2d + kdn + qk^2n = k(kd + dn + qkn).
\end{equation}
\subsubsection{Update $\bD$ (ODL-D)}
After finding $\bX$, the dictionary $\bD$ will be updated by:
\begin{align}
    \bD &= \arg\min_{\bD} -2\trace(\bE\bD^T) + \trace(\Fb\bD^T\bD),
\end{align}
   subject to: $\|\bd_i\|_2 \leq 1$, with $\bE = \bY\bX^T$, and $\Fb = \bX\bX^T$.\\
Each column of $\bD$ will be updated by fixing all others:
\begin{eqnarray*}
    \label{eqn:odl_u}
     \bu \leftarrow \frac{1}{\Fb_{ii}} (\be_i - \bD \fb_i) - \bd_i;~~\bd_i & \leftarrow & \frac{\bu}{\max(1, \|\bu\|_2)},
\end{eqnarray*}
where $\bd_i, \be_i, \fb_i$ are the $i-$th columns of $\bD, \bE, \Fb$ and $\Fb_{ii}$ is the $i-$th element in the diagonal of $\Fb$. The dominant computational task is to compute $\bD\fb_i$ which requires $dk$ operators. Since $\bD$ has $k$ columns and the algorithm requires $q$ iterations,
the complexity of ODL-D is $qdk^2$.
\subsection{Dictionary learning with structured incoherence (DLSI)} 
\label{sub:dictionary_update_in_dlsi}
DLSI \cite{ramirez2010classification} proposed a method to encourage the independence
between bases of different classes by minimizing coherence between cross-class bases. The cost function $J_1(\bD, \bX)$ of DLSI is defined as \eqref{eqn:dlsi_cost_function}.
\subsubsection{Update $\bX$ (DLSI-X)}
In each iteration, the algorithm solves $C$ subproblems:
\begin{equation}
    \bX^c = \arg\min_{\bX^c} \|\bY_c - \bD_c\bX^c\|_F^2 + \lambda\|\bX^c\|_1
\end{equation}
with $\bY_c \in \R^{d\times n}, \bD_c \in \R^{d \times k}$, and $\bX^c \in \R^{k \times n}$. Based on~\eqref{eqn:cplxt_l1norm}, the complexity of updating $\bX$ ($C$ subproblems) is:
\begin{equation}
    Ck(kd + dn + qkn).
\end{equation}
\subsubsection{Original update $\bD$ (O-DLSI-D)}
For updating $\bD$, each sub-dictionary $\bD_c$ is solved via (\ref{eqn:dlsi_updateDc}).
The main step in the algorithm is stated in (\ref{eqn:odlsid_step1}) and (\ref{eqn:odlsid_step2}).
    The dominant computational part is the matrix inversion which has complexity $d^3$. Matrix-vector multiplication and vector normalization can be ignored here.
    Since $\bD_c$ has $k$ columns, and the algorithm requires $q$ iterations,
    the complexity of the O-DLSI-D algorithm is $Cqkd^3$.
\subsubsection{Efficient update $\bD$ (E-DLSI-D)} 
Main steps of the proposed algorithm are presented in equations (\ref{eqn:edlsid_firststep})--(\ref{eqn:edlsid_laststep}) where (\ref{eqn:edlsid_2}) and (\ref{eqn:edlsid_laststep}) require much less computation compared to (\ref{eqn:edlsid_firststep}) and (\ref{eqn:edlsid_3}). The total (estimated) complexity of efficient $\bD_c$ update is a summation of two terms: i) $q$ times ($q$ iterations) of ODL-D in (\ref{eqn:edlsid_firststep}). ii) One matrix inversion ($d^3$) and $q$ matrix multiplications in (\ref{eqn:edlsid_3}). Finally, the complexity of E-DLSI-D is:
\begin{equation}
 C(q^2dk^2 + d^3 + qd^2k) = Cd^3 + Cqdk(qk+d).
\end{equation}
Total complexities of O-DLSI (the combination of DLSI-X and O-DLSI-D) and E-DLSI (the combination of DLSI-X and E-DLSI-D) are summarized in Table \ref{tab:complexity_analysis}.

\begin{table}[t]
\centering
\caption{Complexity analysis for proposed efficient algorithms and their original versions}
\label{tab:complexity_analysis_subproblems}
{\small \begin{tabular}{|l|l|l|}
\hline
Method & Complexity &\begin{tabular}[c]{@{}c@{}} Plugging \\ numbers\end{tabular}\\
\hline
O-DLSI-D & $Cqkd^3$                  & $6.25\times 10^{12}$\\
E-DLSI-D & $Cd^3 + Cqdk(qk + k)$     & $2.52\times 10^{10}$\\
\hline
O-FDDL-X & $C^2k(dn + qCkn + Cdk)$   & $1.51 \times 10^{11}$\\
E-FDDL-X & $C^2k(dn + qCnk + dk)$    & $1.01 \times 10^{11}$\\
\hline
O-FDDL-D & $Cdk(qk + C^2n)$          & $10^{11}$\\
E-FDDL-D & $Cdk(Cn + Cqk) + C^3k^2n$ & $2.8\times 10^{10}$\\
\hline
\end{tabular}}
\end{table}

\vspace{-.26in}
\subsection{Separating the particularity and the commonality dictionary learning (COPAR)} 
\label{sub:COPAR}
\subsubsection{Cost function}
{COPAR \cite{kong2012dictionary} is another dictionary learning method which also considers the shared dictionary (but without the low-rank constraint)}.
By using the same notation as in LRSDL, we can rewrite the cost function of COPAR in the following form:
\begin{align*}
\frac{1}{2}f_1(\bY, \lbD, \lbX) + \lambda\norm{\lbX}_1 +
                \eta\sum_{c=0}^C\sum_{i=0, i \neq c}^C\|\bD_i^T\bD_c\|_F^2,
\end{align*}
where $\displaystyle f_1(\bY, \lbD, \lbX) = \sum_{c=1}^Cr_1(\bY_c, \lbD, \lbX_c)$ and $r_1(\bY_c, \lbD, \lbX_c)$ is defined as
\begin{align*}
     \|\bY_c - \lbD \lbX_c\|_F^2 + \|\bY_c - \bD_0\bX^0_c - \bD_c \bX^c_c\|_F^2 + \sum_{j=1,j\neq c}^C \|\bX_c^j\|_F^2.
\end{align*}

\subsubsection{Update $\bX$ (COPAR-X)}
In sparse coefficient update step, COPAR \cite{kong2012dictionary} solve $\lbX_c$ one by one via one $l_1$-norm regularization problem:
\begin{equation*}
    \tilde{\bX} = \arg\min_{\tilde{\bX}} \|\tilde{\bY} - \tilde{\bD}\tilde{\bX}\|_F^2 + \tilde{\lambda}\|\tilde{\bX}\|_1,
\end{equation*}
where $\tilde{\bY} \in \R^{\tilde{d} \times n}, \tilde{\bD} \in \R^{\tilde{d} \times \tilde{k}}$, $ \tilde{\bX} \in \R^{(\tilde{k} \times n}$, $\tilde{d} =2d + (C-1)k$ and $\tilde{k} = (C+1)k$ (details can be found in Section 3.1 of \cite{kong2012dictionary}). Following results in Section \ref{ssub:odl_x} and supposing that $C \gg 1, q \gg 1, n \approx k \ll d$, the complexity of COPAR-X is:
\begin{eqnarray*}
    C\tilde{k}(\tilde{k}\tilde{d} + \tilde{d}n + q\tilde{k}n) &\approx& C^3k^2(2d + Ck + qn).
\end{eqnarray*}
\vspace{-.3in}
\subsubsection{Update $\bD$ (COPAR-D)}
The COPAR dictionary update algorithm requires to solve $(C+1)$ problems of form (\ref{eqn:dlsi_updateDc_admm_form}). While O-COPAR-D uses the same method as O-DLSI-D (see equations (\ref{eqn:odlsid_step1}-\ref{eqn:odlsid_step2})), the proposed E-COPAR-D takes advantages of E-DLSI-D presented in Section \ref{sub:edlsi}. Therefore, the total complexity of O-COPAR-D is roughly $Cqkd^3$, while the total complexity of E-COPAR-D is roughly $C(q^2dk^2 + d^3 + qd^2k).$ {Here we have supposed $C+1 \approx C$ for large $C$}.

Total complexities of O-COPAR (the combination of COPAR-X and O-COPAR-D) and E-COPAR (the combination of COPAR-X and E-COPAR-D) are summarized in Table \ref{tab:complexity_analysis}.

\subsection{Fisher discrimination dictionary learning (FDDL)} 
\vspace{-.1in}
\label{sub:sparse_coding_update_in_fddl}
\subsubsection{Original update $\bX$ (O-FDDL-X)}
Based on results reported in DFDL~\cite{vu2016tmi}, the complexity of O-FDDL-X is roughly $C^2kn(d + qCk) + C^3dk^2 = C^2k(dn + qCkn + Cdk)$.

\subsubsection{Efficient update $\bX$ (E-FDDL-X)}
Based on section \ref{ssub:eff_fddl_sparse_coefficient_update}, the complexity of E-FDDL-X mainly comes from equation (\ref{eqn:fddl_dif_x}). Recall that function $\M(\bullet)$ does not require much computation. The computation of $\bM$ and ${\bM_c}$ can also be neglected since each required calculation of one column, all other columns are the same. Then the total complexity of the algorithm E-FDDL-X is roughly:
\begin{align}
\nonumber
    &\underbrace{(Ck)d(Ck)}_{\M(\bD^T\bD + \lambda_2\bI)}  +
    \underbrace{(Ck)d(Cn)}_{\M(\bD^T\bY)} +
    q\underbrace{(Ck)(Ck)(Cn)}_{\M(\bD^T\bD + \lambda_2\bI)\bX}, \\
    &=   C^2k(dk + dn + qCnk).
\end{align}

\subsubsection{Original update $\bD$ (O-FDDL-D)}
The original dictionary update in FDDL is divided in to $C$ subproblems. In each subproblem, one dictionary $\bD_c$ will be solved while all others are fixed via:
\par{\small\begin{align}
    \nonumber
    \bD_c &= \arg\min_{\bD_c} \|\widehat{\bY} - \bD_c \bX^c\|_F^2 + \|\bY_c - \bD_c\bX^c_c\|_F^2  + \sum_{i \neq c} \|\bD_c\bX^c_i\|_F^2, \\
        &= \underbrace{\arg\min_{\bD_c} -2\trace(\bE \bD_c^T) + \trace(\Fb\bD_c^T\bD_c)}_{\text{complexity: ~} qdk^2},
\end{align}}
\\where:
\begin{align*}
    \widehat{\bY} &= \bY - \sum_{i \neq c} \bD_i\bX^i & \text{complexity:}~ (C-1)dkCn,\\
    \bE &= {\widehat{\bY}(\bX^c)^T} + \bY_c (\bX_c^c)^T & \text{complexity:}~ d(Cn)k + dnk, \\
    \Fb &=  2(\bX^c)(\bX^c)^T& \text{complexity}~ k(Cn)k.
\end{align*}
When $d \gg k, C \gg 1$, complexity of updating $\bD_c$ is:
\begin{equation}
    qdk^2 + (C^2 + 1)dkn + Ck^2n \approx qdk^2 + C^2dkn
\end{equation}
Then, complexity of O-FDDL-D is $Cdk(qk + C^2n)$.


\subsubsection{Efficient update $\bD$ (E-FDDL-D)}
Based on Lemma \ref{lem:fddl_updateD}, the complexity of E-FDDL-D is:
\begin{align}
\nonumber
    &\underbrace{d(Cn)(Ck)}_{\bY\M(\bX)^T} +
    \underbrace{(Ck)(Cn)(Ck)}_{\M(\bX\bX^T)} +
    \underbrace{qd(Ck)^2}_{\text{ODL in (\ref{eqn:efddl_d_odl})}},  \\
    & = Cdk(Cn + Cqk) + C^3k^2n.
\end{align}
Total complexities of O-FDDL and E-FDDL are summarized in Table \ref{tab:complexity_analysis}.


\subsection{LRSDL} 
\label{sub:lrsdl}
\subsubsection{Update $\bX, \bX^0$}
From (\ref{eqn:hbar_diff_both}), (\ref{eqn:hbar_diff_up}) and (\ref{eqn:hbar_diff_lo}), in each iteration of updating $\lbX$, we need to compute:
\begin{eqnarray*}
    &&(\M(\bD^T\bD) + 2\lambda_2\bI)\bX - {\M(\bD^T\bY)} +\\
    &&\quad\quad\quad +\lambda_2(\bM - 2\widehat{\bM}) - \M(\bD^T\bD_0\bX^0), ~\text{and} \\
    &&{(2\bD_0^T\bD_0 + \lambda_2\bI)}\bX^0 - {2\bD_0^T\bY} + \bD_0^T\bD\M(\bX) - \lambda_2 \bM^0.
\end{eqnarray*}
Therefore, the complexity of LRSDL-X is:
\begin{eqnarray}
\nonumber
    &&\underbrace{(Ck)d(Ck)}_{\bD^T\bD} +
    \underbrace{(Ck)d(Cn)}_{\bD^T\bY} +
    \underbrace{(Ck)dk}_{\bD^T\bD_0} +
    \underbrace{kdk}_{\bD_0^T\bD_0} +
    \underbrace{kd(Cn)}_{\bD_0^T\bY} + \\
\nonumber
    &&+ q\left(
    \begin{matrix}
    \underbrace{(Ck)^2(Cn)}_{(\M(\bD^T\bD) + 2\lambda_2\bI)\bX } +
    \underbrace{(Ck)k(Cn)}_{\M(\bD^T\bD_0\bX^0)} + \\
    +\underbrace{k^2Cn}_{{(2\bD_0^T\bD_0 + \lambda_2\bI)}\bX^0} +
    \underbrace{k(Ck)(Cn)}_{ \bD_0^T\bD\M(\bX)}
    \end{matrix}
    \right), \\
\nonumber
\label{eqn:complexity_lrsdl_X}
    &&\approx C^2k(dk + dn) + Cdk^2 + qCk^2n(C^2 + 2C + 1),\\
    && \approx C^2k(dk + dn + qCkn).
\end{eqnarray}
which is similar to the complexity of E-FDDL-X.
Recall that we have supposed number of classes $C \gg 1$.

\subsubsection{Update $\bD$}
Compare to E-FDDL-D, LRSDL-D requires one more computation of $\lbY = \bY - \bD_0\bX^0$ (see section \ref{ssub:lrsdl_dictionary_update}). Then, the complexity of LRSDL-D is:
\begin{align}
\nonumber
    &\underbrace{Cdk(Cn + Cqk) + C^3k^2n}_{\text{E-FDDL-D}} + \underbrace{dk(Cn)}_{\bD_0\bX^0}, \\
    \label{eqn:complexity_lrsdl_D}
    &\approx Cdk(Cn + Cqk) + C^3k^2n,
\end{align}
which is similar to the complexity of E-FDDL-D.
\subsubsection{Update $\bD_0$}
The algorithm of LRSDL-D0 is presented in section \ref{ssub:lrsdl_dictionary_update} with the main computation comes from (\ref{eqn:lrsdl_d0_form_EF}), (\ref{eqn:lrsdl_d0_1}) and (\ref{eqn:lrsdl_d0_3}). The shrinkage thresholding operator in (\ref{eqn:lrsdl_d0_3}) requires one SVD and two matrix multiplications. The total complexity of LRSDL-D0 is:
\begin{align}
\nonumber
    &\underbrace{d(Ck)(Cn)}_{\bV = \bY - \frac{1}{2} \bD\M(\bX)} +
    \underbrace{d(Cn)k}_{\bE \text{~in (\ref{eqn:lrsdl_d0_form_EF})}} +
    \underbrace{k(Cn)k}_{\Fb \text{~in (\ref{eqn:lrsdl_d0_form_EF})}} +
    \underbrace{qdk^2}_{(\ref{eqn:lrsdl_d0_1})} +
    \underbrace{O(dk^2) + 2dk^2}_{(\ref{eqn:lrsdl_d0_3})}, \\
    \nonumber
    &\approx C^2dkn + qdk^2 + O(dk^2),\\
    \label{eqn:complexity_LRSDL_D0}
    & = C^2dkn + (q + q_2)dk^2, ~~\text{for some}~ q_2.
\end{align}
{By combing \eqref{eqn:complexity_lrsdl_X}, \eqref{eqn:complexity_lrsdl_D} and \eqref{eqn:complexity_LRSDL_D0}, we obtain the total complexity of LRSDL, which is specified in the last row of Table \ref{tab:complexity_analysis}.}
\def\ct{$\pm$}
\begin{table}[t]
\caption{Complexity analysis for different dictionary learning methods}
    \centering
    \label{tab:complexity_analysis}
    {\small
    \begin{tabular}{|l|l|l|}
    \hline
    Method & Complexity & \begin{tabular}[c]{@{}c@{}} Plugging \\ numbers\end{tabular}\\ \hline
    \hline
    O-DLSI & $Ck(kd + dn + qkn)+ Cqkd^3 $& $6.25\times 10^{12}$\\
    \hline
    E-DLSI & \begin{tabular}[c]{@{}c@{}} $Ck(kd + dn + qkn) +$ \\ $Cd^3 + Cqdk(qk+d) $ \end{tabular}& $3.75\times 10^{10}$\\
    \hline
    O-FDDL  & \begin{tabular}[c]{@{}c@{}} $C^2dk(n+Ck+Cn) +$ \\ $+Ck^2q(d + C^2n)$ \end{tabular}& $2.51\times 10^{11}$\\
    \hline
    E-FDDL & $ C^2k((q+1)k(d + Cn) + 2dn)    $  & $1.29\times 10^{11}$\\ 
    \hline
    O-COPAR & $C^3k^2(2d + Ck + qn) + Cqkd^3 $  & $6.55\times 10^{12}$\\
    \hline
    E-COPAR & \begin{tabular}[c]{@{}c@{}}$C^3k^2(2d + Ck + qn)+$ \\ $+Cd^3 + Cqdk(qk+d)$\\ \end{tabular}  & $3.38\times 10^{11}$\\
    \hline
    LRSDL & \begin{tabular}[c]{@{}c@{}}$C^2k((q+1)k(d + Cn) + 2dn)$ \\ $C^2dkn + (q+q_2)dk^2 $ \end{tabular} & $1.3\times 10^{11}$\\
    \hline
    \end{tabular}
    }
\end{table}

\subsection{Summary}
\label{sub:complexity_analysis_summary}
Table \ref{tab:complexity_analysis_subproblems} and Table \ref{tab:complexity_analysis} show final complexity analysis of each  proposed efficient algorithm and their original counterparts. Table \ref{tab:complexity_analysis} compares LRSDL to other state-of-the-art methods. We pick a typical set of parameters with 100 classes, 20 training samples per class, 10 bases per sub-dictionary and shared dictionary, data dimension 500 and 50 iterations for each iterative method. Concretely, $C = 100,~ n = 20,~ k = 10,~ q = 50,~ d = 500$. We also assume that in (\ref{eqn:complexity_LRSDL_D0}), $q_2 = 50$. Table \ref{tab:complexity_analysis_subproblems} shows that all three proposed efficient algorithms require less computation than original versions with most significant improvements for speeding up DLSI-D. Table \ref{tab:complexity_analysis} demonstrates an interesting fact. LRSDL is the least expensive computationally when compared with other \textit{original} dictionary learning algorithms, and only E-FDDL has lower complexity, which is to be expected since the FDDL cost function is a special case of the LRSDL cost function. COPAR is found to be the most expensive computationally.

\section{Experimental results} 
\label{sec:experiment_results}
\vspace{-0.1in}
\subsection{Comparing methods and datasets}
\label{sub:methods_datasets}
\vspace{-0.15in}
\begin{figure}[t]
\centering
\includegraphics[width = 0.79\textwidth]{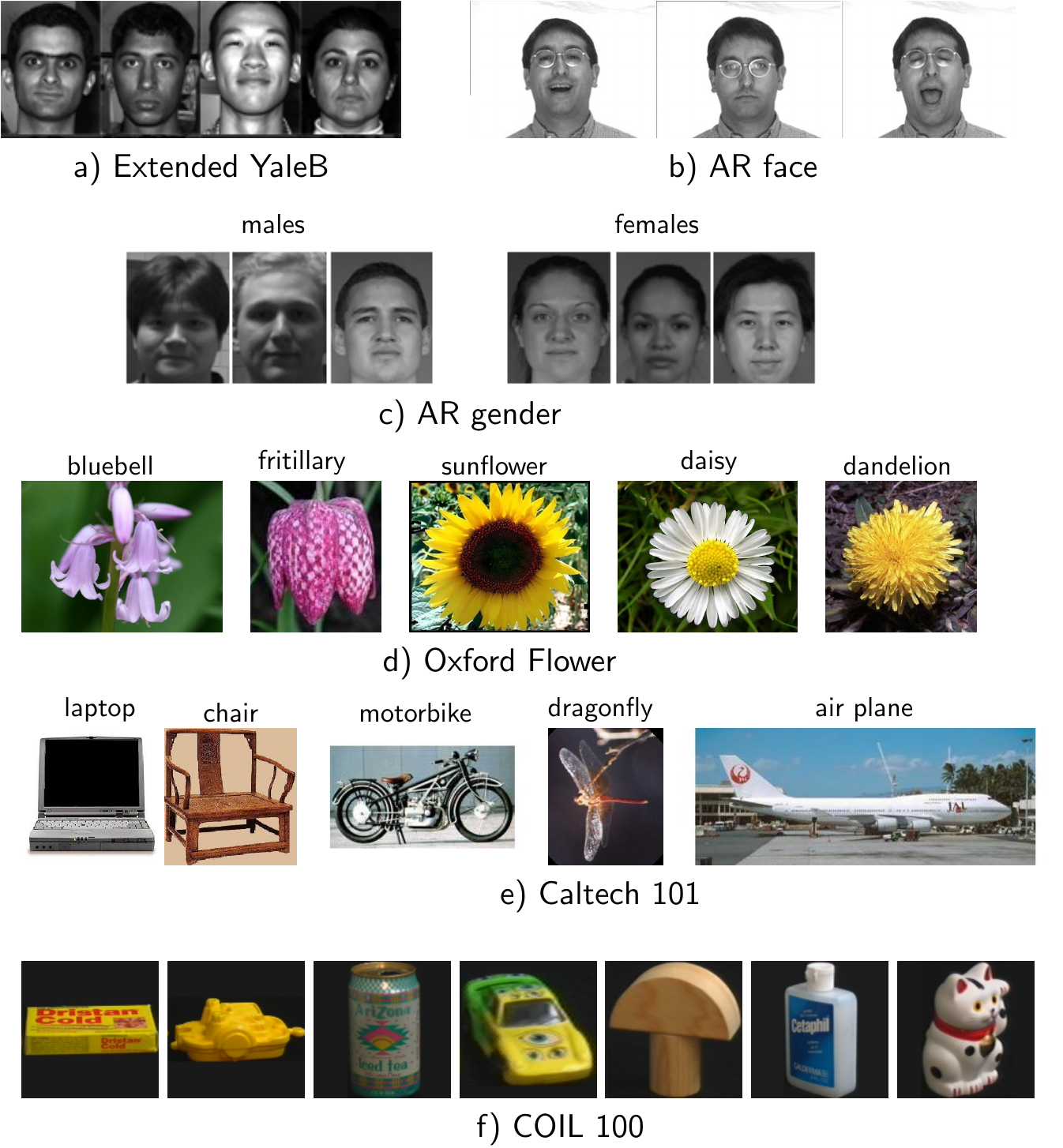}
\caption{\small {Examples from six datasets}. }
\label{fig:examples}
\end{figure}

We present the experimental results of applying these methods to five diverse
datasets: the Extended YaleB face dataset \cite{georghiades2001few}, the AR
face dataset~\cite{ardataset}, the AR gender dataset, the Oxford Flower dataset~\cite{Nilsback06}, and two multi-class object category dataset --
the Caltech 101~\cite{fei2007learning} {and COIL-100~\cite{nene1996columbia}}. Example images from these datasets are
shown in Figure \ref{fig:examples}. We compare our results with those using
SRC~\cite{Wright2009SRC} and other state-of-the-art dictionary learning methods:
LC-KSVD~\cite{Zhuolin2013LCKSVD}, DLSI~\cite{ramirez2010classification},
FDDL~\cite{yang2014sparse}, COPAR~\cite{kong2012dictionary}, $D^2L^2R^2$~\cite{li2014learning}, {DLRD~\cite{ma2012sparse}, JDL~\cite{zhou2014jointly}, and SRRS~\cite{li2016learning}}.
{Regularization parameters in all methods are chosen using five-fold cross-validation \cite{Kohavi95astudy}.} {For each experiment, we use 10 different randomly split training and test sets and report averaged results.}

\begin{figure}[t]
\centering
\includegraphics[width = 0.79\textwidth]{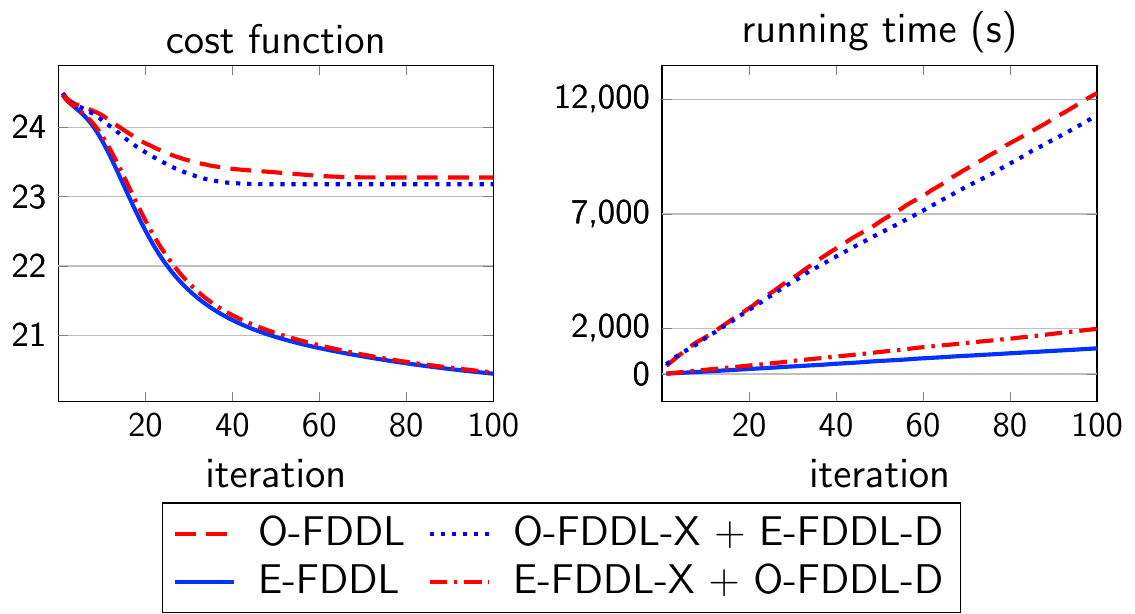}
\caption{\small Original and efficient FDDL convergence rate comparison. }
\label{fig:compare_fddl_algs}
\end{figure}

\begin{figure}[t]
\centering
\includegraphics[width = 0.79\textwidth]{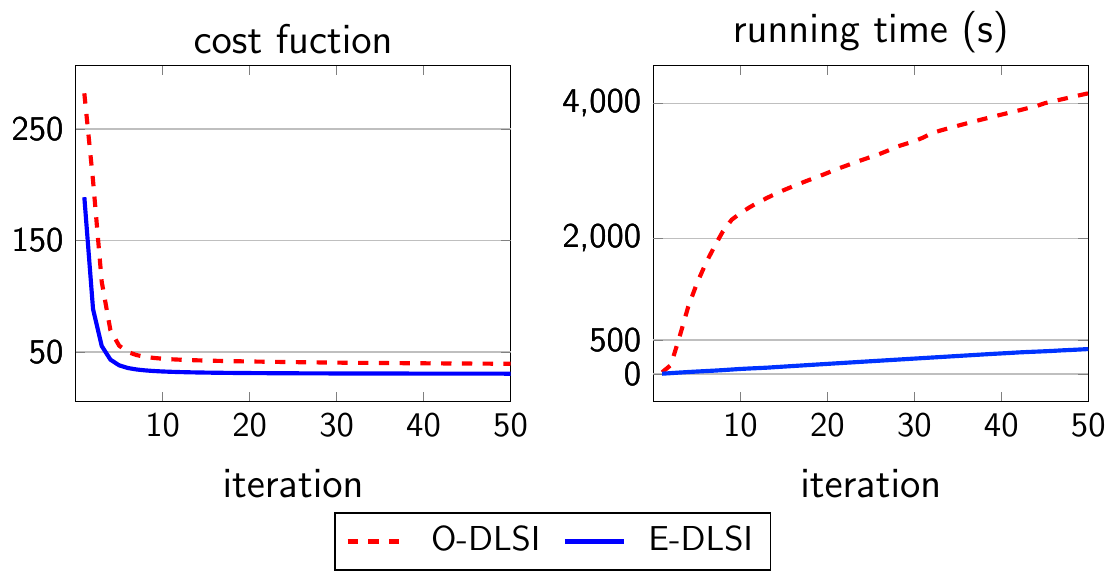}
\caption{\small DLSI convergence rate comparison. }
\label{fig:compare_dlsi_alg}
\end{figure}

\begin{figure}[t]
\centering
\includegraphics[width = 0.79\textwidth]{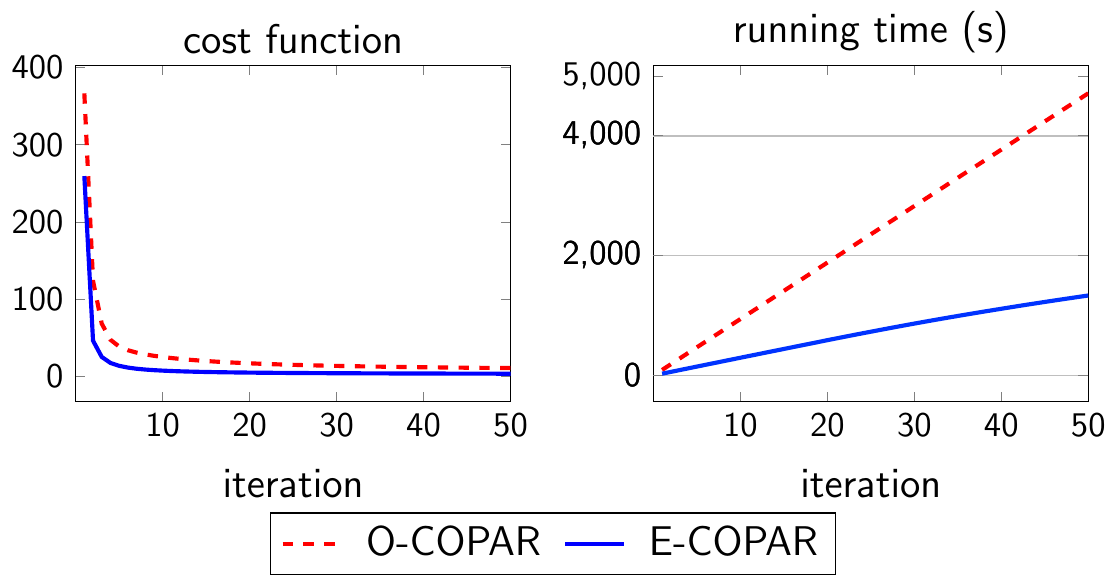}
\caption{\small COPAR convergence rate comparison.}
\label{fig:compare_COPAR_alg}
\end{figure}

\par For {\bf two face datasets}, feature descriptors are random faces, which are made by  projecting face images onto a random vector using a random projection matrix.  As in
\cite{zhang2010discriminative}, the dimension of a random-face feature in the
Extended YaleB is $d= 504$, while the dimension in AR face is $d = 540$. Samples of these two datasets are shown in  Figure \ref{fig:examples}a) and b).

\par For {\bf the AR gender dataset}, we first choose a non-occluded subset (14 images per person)
from the AR face dataset, which consists of 50 males and 50 females, to conduct experiment of gender
classification. Training images are taken from the first 25 males and 25 females, while test images
comprises all samples from the remaining 25 males and 25 females. PCA was used to reduce the dimension
of each image to 300. Samples of this dataset are shown in Figure \ref{fig:examples}c).
\par {\bf The Oxford Flower dataset} is a collection of images of flowers drawn from 17 species with
80 images per class, totaling 1360 images. For feature extraction, based on the impressive results
presented in \cite{yang2014sparse}, we choose the Frequent Local Histogram feature
extractor \cite{fernando2012effective} to obtain feature vectors of dimension 10,000.
The test set consists of 20 images per class, the remaining 60 images per class are
used for training. Samples of this dataset are shown in Figure \ref{fig:examples}d).

\par For the {\bf Caltech 101 dataset}, we use a dense SIFT (DSIFT) descriptor. The DSIFT
descriptor is extracted from $25\times 25$ patch which is densely sampled on a dense grid
with 8 pixels. We then extract the sparse coding spatial pyramid matching (ScSPM) feature \cite{yang2009linear},
which is the concatenation of vectors pooled from words of the extracted DSIFT descriptor.
Dimension of words is 1024 and max pooling technique is used with pooling grid of
$1\times 1, 2 \times 2$, and $4 \times 4$. With this setup, the dimension of ScSPM
feature is 21,504; this is followed by dimension reduction to $d = 3000$ using PCA.

\begin{figure*}[t]
\centering
  \includegraphics[width=\textwidth]{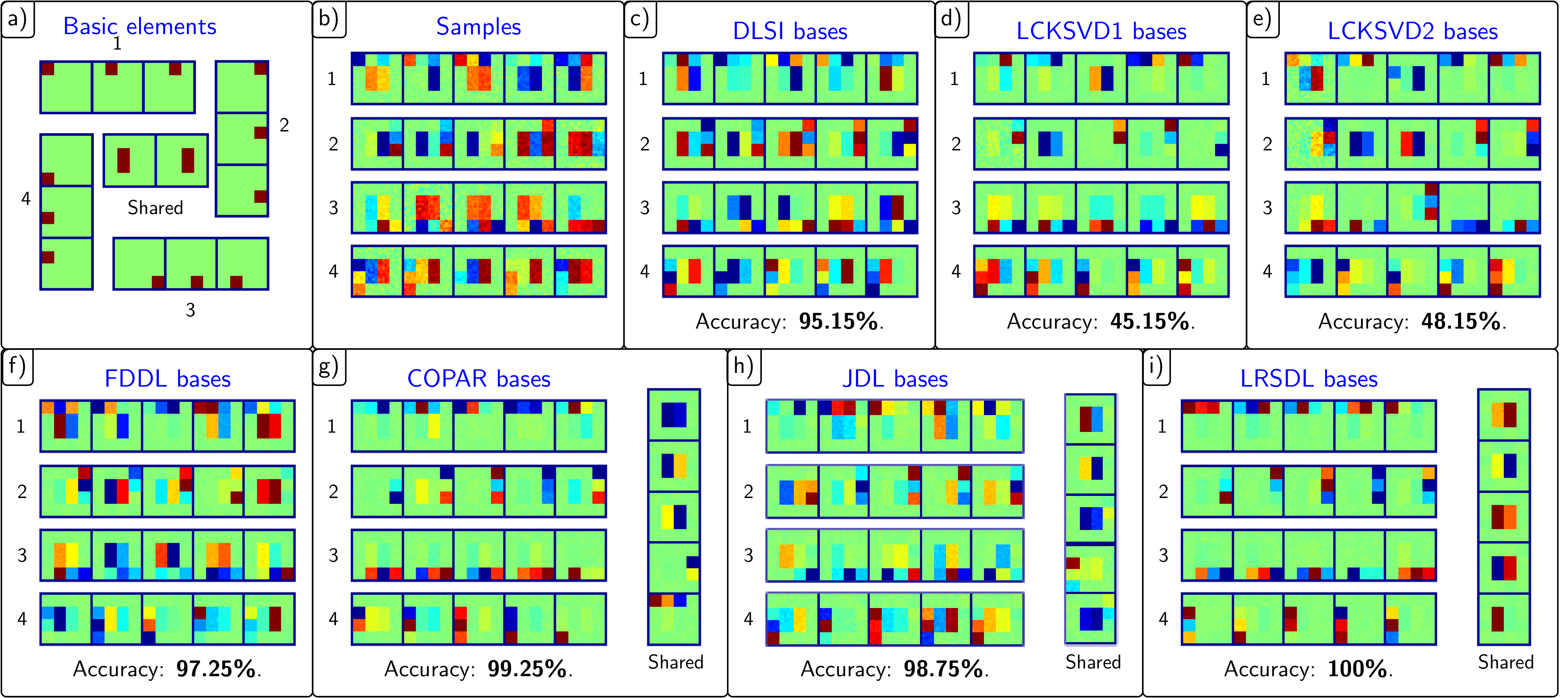}
  \vspace{-0.1in}
 \caption{\small Visualization of learned bases of different dictionary learning methods on the simulated data.}
  \label{fig: bases_simulated}
\end{figure*}

\par {The {\bf COIL-100 dataset} contains various views of 100 objects with different lighting conditions. Each object has 72 images captured from equally spaced views. Similar to the work in~\cite{li2016learning}, we randomly choose 10 views of each object for training, the rest is used for test. To obtain the feature vector of each image, we first convert it to grayscale, resize to 32 $\times$ 32 pixel, vectorize this matrix to a 1024-dimensional vector, and finally normalize it to have unit norm. }

\par
Samples of this dataset are shown in Figure \ref{fig:examples}e).
\vspace{-0.2in}
\subsection{Validation of efficient algorithms} 
\label{sub:efficient_algs}
\vspace{-0.1in}
\label{sub:valid_eff_algs}


To evaluate the improvement of three efficient algorithms proposed in section \ref{sec:contribution}, we apply these efficient algorithms and their original versions on training samples from the AR face dataset to verify the convergence speed of those algorithms. In this example, number of classes $C = 100$, the random-face
feature dimension $d = 300$, number of training samples per class $n_c = n = 7$,
number of atoms in each particular dictionary $k_c = 7$.
\subsubsection{E-FDDL-D and E-FDDL-X}
Figure \ref{fig:compare_fddl_algs} shows the cost functions and running time after each of 100 iterations of 4 different versions of FDDL: the original FDDL (O-FDDL), combination of O-FDDL-X and E-FDDL-D, combination of E-FDDL-X and O-FDDL-D, and the efficient FDDL (E-FDDL). The first observation is that O-FDDL converges quickly to a suboptimal solution, which is far from the best cost obtained by E-FDDL. In addition, while O-FDDL requires more than 12,000 seconds (around 3 hours and 20 minutes) to run 100 iterations, it takes E-FDDL only half an hour to do the same task.

\subsubsection{E-DLSI-D and E-COPAR-D}

Figure \ref{fig:compare_dlsi_alg} and \ref{fig:compare_COPAR_alg} compare convergence rates of DLSI and COPAR algorithms. As we can see, while the cost function value improves slightly, the run time of efficient algorithms reduces significantly. Based on benefits in both cost function value and computation, in the rest of this chapter, we use efficient optimization algorithms instead of original versions for obtaining classification results. 

\def\ct{$\pm~$}
\begin{table*}[t]
\centering
\caption{Overall accuracy (mean \ct standard deviation) (\%) of different dictionary learning methods on different datasets. Numbers in parentheses are number of training samples per class.}
\label{tab:overall_results}
{\tiny \begin{tabular}{|l||c|c|c|c|c|c|}
\hline
        & \begin{tabular}[c]{@{}c@{}}Ext. \\ YaleB (30) \end{tabular} & AR (20)    & \begin{tabular}[c]{@{}c@{}}AR \\ gender (250) \end{tabular} & \begin{tabular}[c]{@{}c@{}}Oxford \\ Flower (60) \end{tabular} & \begin{tabular}[c]{@{}c@{}}Caltech 101 \\ (30)\end{tabular} & COIL100 (10) \\  \hline \hline
SRC \cite{Wright2009SRC}              & 97.96 \ct 0.22           & 97.33 \ct 0.39           & 92.57 \ct 0.00          & 75.79 \ct 0.23          & 72.15 \ct 0.36          & 81.45 $\pm$ 0.80\\ \hline
LC-KSVD1 \cite{Zhuolin2013LCKSVD}     & 97.09 \ct 0.52           & 97.78 \ct 0.36           & 88.42 \ct 1.02          & 91.47 \ct 1.04                  & 73.40  \ct 0.64                 & 81.37 \ct 0.31\\ \hline
LC-KSVD2 \cite{Zhuolin2013LCKSVD}     & 97.80 \ct 0.37           & 97.70 \ct 0.23           & 90.14 \ct 0.45          & \textbf{\textit{92.00 \ct 0.73}} & 73.60 \ct 0.53                  & 81.42 \ct 0.33\\ \hline
DLSI \cite{ramirez2010classification} & 96.50 \ct 0.85           & 96.67 \ct 1.02           & 93.86 \ct 0.27          & 85.29 \ct 1.12          & 70.67 \ct 0.73           & 80.67 \ct 0.46  \\ \hline
DLRD~\cite{ma2012sparse}              & 93.56 \ct 1.25     & 97.83 \ct 0.80           & 92.71 \ct 0.43           &  -                      &  -                      &- \\ \hline
FDDL \cite{yang2014sparse}            & 97.52 \ct 0.63           & 96.16 \ct 1.16           & 93.70 \ct 0.24          & 91.17 \ct 0.89          & 72.94 \ct 0.26          & 77.45 \ct 1.04 \\ \hline
$D^2L^2R^2$\cite{li2014learning}      & 96.70 \ct 0.57           & 95.33 \ct 1.03           & 93.71 \ct 0.87          & 83.23 \ct 1.34          & 75.26 \ct 0.72          & 76.27 \ct 0.98\\ \hline
COPAR \cite{kong2012dictionary}       & \textbf{\textit{98.19 \ct 0.21}}  & 98.50 \ct 0.53  & \textbf{\textit{95.14 \ct 0.52}} & 85.29 \ct 0.74                  & \textbf{\textit{76.05 \ct 0.72}} & 80.46 \ct 0.61\\ \hline
JDL~\cite{zhou2014jointly}            & 94.99 \ct 0.53         & 96.00 \ct 0.96         & {93.86 \ct 0.43}        & 80.29 \ct 0.26        & {75.90 \ct 0.70}        & 80.77 \ct 0.85 \\ \hline
JDL$^*$~\cite{zhou2014jointly}        & 97.73 \ct 0.66         & \textbf{\textit{98.80 \ct 0.34}}         & 92.83 \ct 0.12       & 80.29 \ct 0.26        & 73.47 \ct 0.67        & 80.30 \ct 1.10\\ \hline
SRRS~\cite{li2016learning}            & 97.75 \ct 0.58         &   96.70 \ct 1.26                       & 91.28 \ct 0.15          & 88.52 \ct 0.64                         &   65.22 + 0.34                      & \bf{85.04 \ct 0.45}\\ \hline
LRSDL                                 & {\bf 98.76 \ct 0.23} & {\bf 98.87 \ct 0.43} & \bf{95.42 \ct 0.48}     & \bf{92.58 \ct 0.62}     & \bf{76.70 \ct 0.42}     & \textbf{\textit{84.35 \ct 0.37}}\\ \hline
\end{tabular}}
\end{table*}
\begin{figure}[t]
\centering
  \includegraphics[width=.6\textwidth]{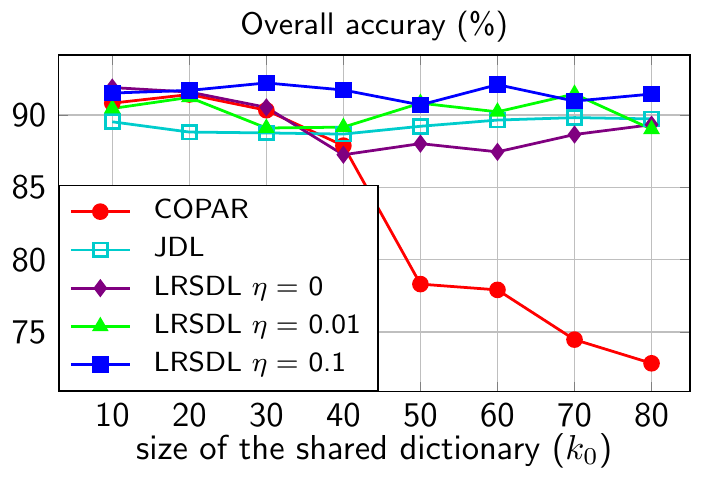}
 \caption{\small Dependence of overall accuracy on the shared dictionary.}
  \label{fig: compare_shared}
\end{figure}

\begin{figure}[t]
\centering
  \includegraphics[width=.6\textwidth]{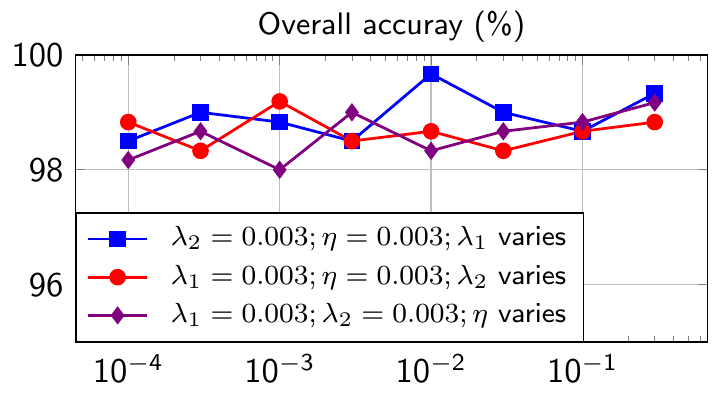}
 \caption{\small {Dependence of overall accuracy on parameters (the AR face dataset, $C = 100, n_c = 20, k_c = 15, k_0 = 10$).}}
  \label{fig:par_change}
\end{figure}
\subsection{Visualization of learned shared bases}
To demonstrate the behavior of dictionary learning methods on a dataset in the presence of shared features, we create a toy example in Figure \ref{fig: bases_simulated}. This is a classification problem with 4 classes whose basic class-specific elements and shared elements are visualized in Figure \ref{fig: bases_simulated}a). Each basis element has dimension $20$ pixel$\times 20$ pixel. From these elements, we generate 1000 samples per class by linearly combining class-specific elements and shared elements followed by noise added; 200 samples per class are used for training, 800 remaining images are used for testing. Samples of each class are shown in Figure \ref{fig: bases_simulated}b).

Figure \ref{fig: bases_simulated}c) show sample learned bases using DLSI \cite{ramirez2010classification} where shared features are still hidden in class-specific bases. In LC-KSVD bases (Figure \ref{fig: bases_simulated}d) and e)), shared features (the squared in the middle of a patch) are found
but they are classified as bases of class 1 or class 2, diminishing classification accuracy since most of test samples are classified as class 1 or 2. The same phenomenon happens in FDDL bases (Figure \ref{fig: bases_simulated}f)).
\par
The best classification results happen in three shared dictionary learnings (COPAR~\cite{kong2012dictionary} in Figure \ref{fig: bases_simulated}g), JDL~\cite{zhou2014jointly} in Figure \ref{fig: bases_simulated}h) and the proposed LRSDL in Figure \ref{fig: bases_simulated}i)) where the shared bases are extracted and gathered in the shared dictionary. However, in COPAR and JDL, shared features still appear in class-specific dictionaries and the shared dictionary also includes class-specific features. In LRSDL, class-specific elements and shared elements are nearly perfectly decomposed into appropriate sub dictionaries. The reason behind this phenomenon is the low-rank constraint on the shared dictionary of LRSDL. Thanks to this constraint, LRSDL produces perfect results on this simulated data.





\begin{figure*}[t]
\centering
  \includegraphics[width=\textwidth]{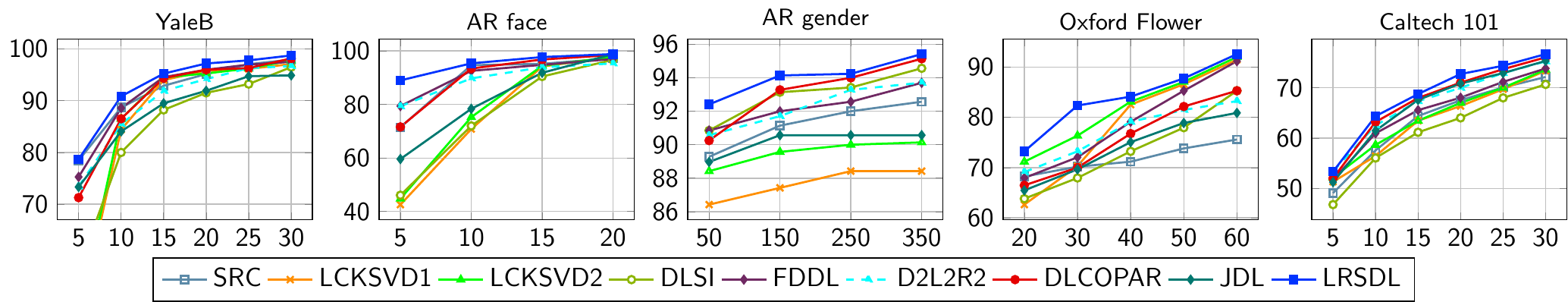}
  \vspace{-0.1in}
 \caption{\small Overall classification accuracy (\%) as a function of training set size per class.}
  \label{fig: compare_ntrains}
\end{figure*}
\vspace{-.2in}
\subsection{Effect of the shared dictionary sizes on overall accuracy} 
\vspace{-.1in}

We perform an experiment to study the effect of the shared dictionary size on the overall classification results of three shared dictionary methods: COPAR \cite{kong2012dictionary}, JDL~\cite{zhou2014jointly} and LRSDL in the AR gender dataset. In this experiment, 40 images of each class are used for training. The number of shared dictionary bases varies from 10 to 80. In LRSDL, because there is a regularization parameter $\eta$ which is attached to the low-rank term (see equation (\ref{eqn:lrsdl_cost_fn})), we further consider three values of $\eta$: $\eta = 0$, i.e.\ no low-rank constraint, $\eta = 0.01$ and $\eta=0.1$ for two different degrees of emphasis. Results are shown in Figure \ref{fig: compare_shared}.
\par

We observe that the performance of COPAR heavily depends on the choice of $k_0$ and its results worsen as the size of the shared dictionary increases. The reason is that when $k_0$ is large, COPAR tends to absorb class-specific features into the shared dictionary. This trend is {\em not} associated with LRSDL even when the low-rank constraint is ignored ($\eta = 0$), because LRSDL has another constraint ($\|\bX^0 - \bM^0\|_F^2$ small) which forces the coefficients corresponding to the shared dictionary to be similar. Additionally, when we increase $\eta$, the overall classification of LRSDL  also gets better. These observations confirm that our two proposed constraints on the shared dictionary are important, and the LRSDL exhibits robustness to parameter choices. {For JDL, we also observe that its performance is robust to the shared dictionary size, but the results are not as good as those of LRSDL.}


\vspace{-0.1in}
\subsection{Overall Classification Accuracy} 
\label{sub:overall_classification_acc}
Table \ref{tab:overall_results} shows overall classification results of various methods on all presented datasets {in terms of \textit{mean \ct standard deviation}}. It is evident that in most cases, three dictionary learning methods with shared features (COPAR \cite{kong2012dictionary}, JDL~\cite{zhou2014jointly} and our proposed LRSDL) outperform others with all five highest values presenting in our proposed LRSDL. {Note that JDL method represents the query sample class by class. We also extend this method by representing the query sample on the whole dictionary and use the residual for classification as in SRC. This extended version of JDL is called JDL*.}

\begin{table*}[]
\centering
{
\caption{Training and test time per sample (seconds) of different dictionary learning method on the Oxford Flower dataset ($n_c = 60, d = 10000, C = 17, K \approx 40\times 17$).} 
\label{tab:running_time}
{\tiny \begin{tabular}{c|ccccccccccc}
\hline
      & SRC    & LCKSVD1 & LCKSVD2 & DLSI  & FDDL   & $D^2L^2R^2$  & COPAR & JDL    & SRRS & LRSDL \\ \hline
Train & 0      & 1.53e3    &  1.46e3   & 1.4e3 & 9.2e2  & $>$1 day      & 1.8e4 & 7.5e1  & 3.2e3     & 1.8e3\\ \hline
Test  & 3.2e-2 &  6.8e-3 &  6.1e-3 & 4e-3  & 6.4e-3 & 3.3e-2           & 5.5e-3& 3.6e-3 & 3.7e-3     & 2.4e-2\\ \hline
\end{tabular}}}
\end{table*}

\vspace{-.2in}
\subsection{Performance vs. size of training set} 
\label{sub:performance_vs_size_of_training_set}
\vspace{-.1in}
Real-world classification tasks often have to contend with lack of availability of large training sets. To understand training dependence of the various techniques, we present a comparison of overall classification accuracy as a function of the training set size of the different methods. In Figure \ref{fig: compare_ntrains}, overall classification accuracies are reported for first five datasets\footnote{For the COIL-100 dataset, number of training images per class is already small (10), we do not include its results here.} corresponding to various scenarios. It is readily apparent that LRSDL exhibits the most graceful decline as training is reduced. In addition, LRSDL also shows high performance even with low training on AR datasets.
\vspace{-.2in}
{
\subsection{Performance of LRSDL with varied parameters}
\label{sub:par_change}
Figure \ref{fig:par_change} shows the performance of LRSDL on the AR face dataset with different values of $\lambda_1$, $\lambda_2$, and $\eta$ with other parameters fixed. We first set these three parameters to $0.003$ then vary each parameter from $10^{-4}$ to $0.3$ while two others are fixed. We observe that the performance is robust to different values with the accuracies being greater than 98\% in most cases. It also shows that LRSDL achieves the best performance when $\lambda_1 = 0.01, \lambda_2 = 0.003, \eta = 0.003$.
\subsection{Run time of different dictionary learning methods} 
\label{sub:running_time_of_different_dictionary_learning_methods}
Finally, we compare training and test time per sample of different dictionary learning methods on the Oxford Flower dataset. Note that, we use the efficient FDDL, DLSI, COPAR in this experiment. Results are shown in Table \ref{tab:running_time}. This result is consistent with the complexity analysis reported in Table \ref{tab:complexity_analysis} with training time of LRSDL being around half an hour, 10 times faster than COPAR~\cite{kong2012dictionary} and also better than other low-rank models, i.e. $D^2L^2R^2$~\cite{li2014learning}, and SRRS~\cite{li2016learning}. }
\section{Discussion and Conclusion} 
\label{sec:discussion_and_conclusion}
In this chapter, our primary contribution is the development of a discriminative dictionary learning framework via the introduction of a shared dictionary {with two crucial constraints. First, the shared dictionary is constrained to be low-rank. Second, the sparse coefficients corresponding to the shared dictionary obey a similarity constraint.} In conjunction with discriminative model as proposed in \cite{Meng2011FDDL,yang2014sparse}, this leads to a more flexible model where shared features are excluded before doing classification. An important benefit of this model is the robustness of the framework to size ($k_0$) and the regularization parameter ($\eta$) of the shared dictionary. In comparison with state-of-the-art algorithms developed specifically for these tasks, our LRSDL approach offers better classification performance on average.
\par

In Section~\ref{sub:solving_the_opt} and \ref{sub:edlsi}, we discuss the efficient algorithms for FDDL \cite{yang2014sparse}, DLSI \cite{ramirez2010classification}, then flexibly apply them into more sophisticated models. Thereafter in Section~\ref{sec:complexity_analysis} and ~\ref{sub:valid_eff_algs}, we both theoretically and practically show that the proposed algorithms indeed significantly improve cost functions and run time speeds of different dictionary learning algorithms. The complexity analysis also shows that the proposed LRSDL requires less computation than competing models.
\par
As proposed, the LRSDL model learns a dictionary shared by every class. In some practical problems, a feature may belong to more than one but not all classes. Very recently, researchers have begun to address this issue \cite{yang2014latent, yoon2014hierarchical}.
In future work, we will investigate the design of hierarchical models for extracting common features among classes. 

\chapter{Classifying Multi-channel UWB SAR Imagery via Tensor
Sparsity Learning Techniques}

\def\bA{\mathbfcal{A}}
\def\bD{\mathbfcal{D}}
\def\bM{\mathbfcal{M}}
\def\bN{\mathbfcal{N}}
\def\bP{\mathbfcal{P}}
\def\bX{\mathbfcal{X}}
\def\bY{\mathbfcal{Y}}
\def\bxt{\mathbcal{x}^{(t)}}
\def\byt{\mathbcal{y}^{(t)}}
\def\bDt{\mathbfcal{D}^{(t)}}
\def\bw{\mathbcal{w}}
\def\bx{\mathbcal{x}}
\def\by{\mathbcal{y}}
\def\bg{\mathbcal{g}}
\def\vec{\text{vec}}

\label{chapter:contrib3}
\section{Introduction}
\label{sec:intro}

Over the past two decades, the U.S. Army has been investigating the capability
of low-frequency, ultra-wideband (UWB) synthetic aperture radar (SAR) systems
for the detection of buried and obscured targets in various applications, such
as foliage penetration~\cite{lam1997}, ground penetration~\cite{lam1998}, and
sensing-through-the-wall~\cite{lam2008}. These systems must operate in the
low-frequency spectrum spanning from UHF frequency band to L band to achieve
both resolution and penetration capability. Although a lot of progress has been
made over the years, one critical challenge that
low-frequency UWB SAR technology still faces is discrimination of targets of
interest from other natural and manmade clutter objects in the scene. The key
issue is that that the targets of interest are typically small compared to the
wavelength of the radar signals in this frequency band and have very low radar
cross sections (RCSs). Thus, it is very difficult to discriminate targets and
clutter objects using low-resolution SAR imagery.

\subsection{SAR geometry and image formation overview}

\begin{figure}[t]
\centering
\includegraphics[width = .75\textwidth]{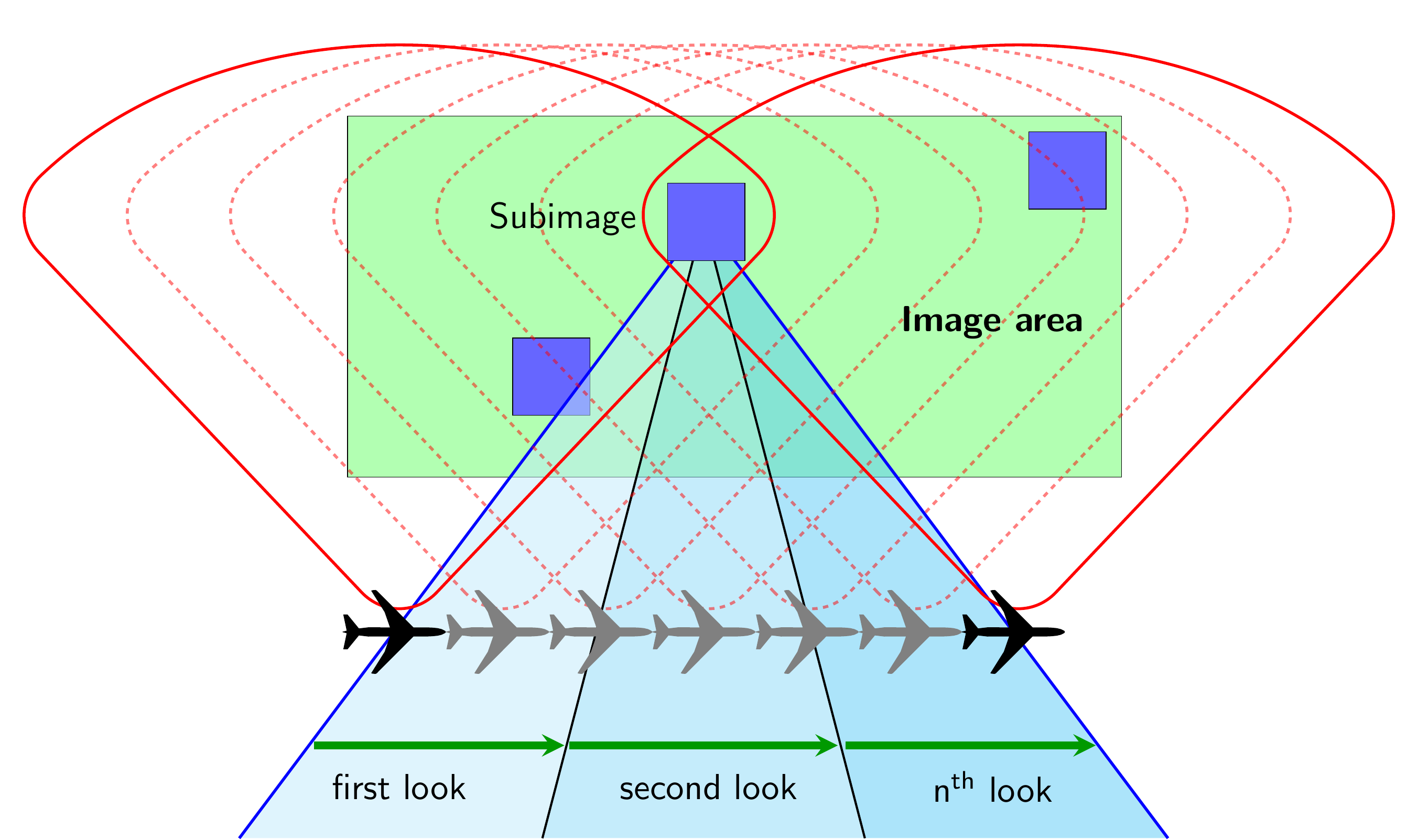}
\caption{\small Constant integration angle and multi-look SAR image formation 
via backprojection.}
\label{fig1}
\end{figure}

Figure~\ref{fig1} shows a typical side-looking SAR geometry
where a vehicle or an airborne-based radar transmits wideband signals to
the imaging area and measures the backscatter signals as it moves along an
aperture. Let $s_k$ be the received range-compressed data received at
$k^{\text{th}}$ aperture; the backprojection algorithm~\cite{nguyen1998ultra}
computes the value for each SAR image pixel from the imaging area as follows:
\begin{equation}
    P_i = \sum_{k = N_{i_1}}^{N_{i_2}}w_k * s_k(f(i, k)),
\end{equation}
where $N_{i_2} - N_{i_1}+1$ is the total number of aperture records used for the
coherent integration, $w_k$ is the weighting value at the $k^{\text{th}}$
aperture position, and $f(i, k)$ is the shift index to the signal $s_k$ for
$i^{\text{th}}$ pixel at the $k^{\text{th}}$ aperture position. For a typical
SAR image with $0^{\circ}$ aspect angle, $N_{i_1}$ and $N_{i_2}$ correspond to
the angle values of $-\alpha/2$ and $\alpha/2$ to form a SAR image with an
integration angle of $\alpha$. Note that for a true constant integration angle
SAR image formation, $N_{i_1}$ and $N_{i_2}$ are computed for every pixel of the
SAR image. However, for computational efficiency, a large image area is divided
into smaller subimages. For each subimage, SAR image formation is computed using
the $N_{i_1}$ and $N_{i_2}$ values derived from the geometry of the center of
the subimage and the radar aperture. To exploit the aspect dependence
information of target, instead of forming a single SAR image with $0^{\circ}$
aspect angle as described above with a single sector $[-\alpha/2, \alpha/2]$, we
form multiple sectors to generate the corresponding SAR images at different
aspect angles. For example, in our \textit{consecutive multi-look experiment},
three SAR images are formed: left side image that covers sector $[-\alpha/2,
\alpha/6]$, broadside image that covers sector $[-\alpha/6,
\alpha/6]$, and right side image at sector {$[\alpha/6, \alpha/2]$}.

To achieve a constant cross-range resolution SAR image, a large image area is
also divided into smaller subimages. Each subimage is formed using a different subset
of aperture. For a constant integration angle, subimages at farther range would
be integrated using a longer aperture than near-range subimages. The subimages
are then mosaicked together to form a single large image that covers the area of
interest. In consecutive multi-look SAR image processing mode, instead of
generating a single SAR image, multiple SAR images are formed at different
viewing angles to exploit the aspect angle dependent information from
targets. Thus, each aperture for each subimage is further divided into smaller
segments (either disjoint or overlapped) that are used to form consecutive
multi-look SAR images as illustrated in Figure~\ref{fig1}. The
aspect angle dependence features from targets have been exploited in {past}
research before using different
techniques~\cite{damarla2001detection,runkle2001multi,nguyen1998ultra}.

\subsection{Closely related works and motivation}
\label{ssec:related}

{UWB radar techniques have recently attracted increasing attention in the
area of penetration and object detection, thanks to their usage in
security applications and surveillance systems~\cite{anabuki2017ultrawideband}.
T. Sakamoto \etal~\cite{sakamoto2016fast} proposed fast methods {for}
   ultra-wideband (UWB) radar imaging that can be applied to a moving target.
   The technology has been also applied to 3-D imaging applications~\cite{yamaryo2018range}, human
   posture~\cite{kiasari2014classification}, human
   activity~\cite{bryan2012application}, vital sign~\cite{liang2018through},
   and liquid material~\cite{wang2015classification} classification problems. In
   these papers, due to high dimensionality and small signal-to-ratio (SNR), the
   signals need to be preprocessed, e.g., dimensionality reduction and
   background subtraction, before being used to train a classifier. It has been
   shown that support vector machines usually provide the best
   performance~\cite{bryan2012application,wang2015classification}. It is worth
   noting that in the aforementioned applications, objects are usually big
   (human) and captured from a relatively small distance. On the contrary,
   objects in our problem are relatively small and quite far from the radar. }

In this chapter, we consider the problem of discriminating and classifying buried
targets of interest (metal and plastic mines, 155-mm unexploded ordinance [UXO],
etc.) from other natural and manmade clutter objects (a soda can, rocks, etc.)
in the presence of noisy responses from the rough ground surfaces for
low-frequency UWB 2-D SAR images. For classification problems, sparse
representation-based classification~\cite{Wright2009SRC} (SRC) has been
successfully demonstrated in other imagery domains such as medical image
classification~\cite{vu2015dfdl, Srinivas2013, vu2016tmi, Srinivas2014SHIRC},
hyperspectral image classification~\cite{sun2015task, sun2014structured,
chen2013hyperspectral}, high-resolution X-band SAR image
classification~\cite{zhang2012multi}, video anomaly detection
\cite{mo2014adaptive}, and several others
\cite{vu2016icip,Mousavi2014ICIP, vu2016fast, srinivas2015structured,
zhang2012joint, dao2014structured, dao2016collaborative, van2013design}.
However, in the low-frequency RF UWB SAR domain, although we have studied the
feasibility of using SRC for higher-resolution 3-D down-looking SAR
imagery~\cite{lamnguyen2013}, the application of SRC to low-frequency UWB 2-D
SAR imagery has not been studied to date due to the aforementioned
low-resolution issue. In this chapter, we generalize the traditional SRC to
{address} target classification using either a single channel (radar
polarization) or multiple channels of SAR imagery. Additionally, we further
propose a novel discriminative tensor sparsity framework for multi-look
multi-channel classification problem, which is naturally suitable for our
problem. In sparse representations, many signals can be expressed by a linear
combination of a few basic elements taken from a ``dictionary''. Based on this theory,
SRC~\cite{Wright2009SRC} was originally developed for robust face recognition.
The main idea in SRC is to represent a test sample as a linear combination of
samples from the available training set. Sparsity manifests because most of the
nonzero components correspond to basic elements with the same class as the test sample.

Multi-channel SRC has
been {investigated} before in medical images~\cite{Srinivas2013, Srinivas2014SHIRC}.
In these papers, one dictionary for each channel is formed from training data
with locations of all channels of one training point being the same in all
dictionaries. Then intuitively, when sparsely encoding each channel of a new
test point using these dictionaries, we obtain sparse codes whose active
(nonzero) elements tend to happen at the same locations in all channels. In
other words, active elements are simultaneously located at the same location
across all channels. This intuition is formulated based on $l_0$ pseudo-norm,
which is solved using a modified version of simultaneous orthogonal matching
pursuit (SOMP)~\cite{tropp2006algorithms}. The cost function is nonconvex, and
hence, it is difficult to find the global solution. Furthermore, when more
constraints involved, there is no straightforward way to extend the algorithm.
In this chapter, we proposed another way of formulating the simultaneity
constraint based on the $l_{12}$ norm, which enforces the row sparsity of the
code matrix (in tensor form, we call it \textit{tube sparsity}). The newly
convex optimization problem can be solved effectively using the fast iterative
shrinkage thresholding algorithm (FISTA)~\cite{beck2009fast}. We also propose
other different tensor sparsity models for our multi-channel classification
problems.

It has been shown that learning a dictionary from the training samples instead
of concatenating all of them as a dictionary can further enhance performance of
sparsity-based methods. On one hand, the training set can be compacted into a smaller dictionary, reducing computational burden at the test time. On the other
hand, by using dictionary learning, discriminative information of different
classes can be trained via structured discriminative constraints on the
dictionary as well as the sparse tensor code. A comprehensive study of
discriminative dictionary learning methods with implementations is presented
at~\cite{vu2016fast,vu2017dictol}. These dictionary learning methods, however, are all
applied to single-channel problem where samples are often represented in form of
vectors. While multi-channel signals can be converted to a long vector by
concatenating all channels, this trivial modification not only leads to the
\textit{curse of dimensionality} of high-dimensional space, but also possibly
neglects cross-channel information, which might be crucial for classification. In
this chapter, we also propose a method named TensorDL, which is a natural
extension of single-channel dictionary learning frameworks to multi-channel
dictionary learning ones. Particularly, the cross-channel information will be
captured using the aforementioned simultaneity constraint.

Naturally, when a radar carried by a vehicle or aircraft moves around an
object of interest, it can capture multiple consecutive views of that object
(see Fig.~\ref{fig1}). Consequently, if the multi-look
information is exploited, the classification accuracy will be improved. While
the multi-look classification problem has been
approached before by SRC-related methods~\cite{zhang2012multi, Mousavi2014ICIP}, none of these works
uses the relative continuity of different views. We propose a framework to
intuitively exploit this important information. More importantly, the
optimization problem corresponding to this structure can be converted to the
simultaneous sparsity model by using an elegant trick {that we call} ShiftSRC.
Essentially, a tensor dictionary is built by circularly shifting an appropriate
amount of a single-channel dictionary. When we sparsely code the multi-look
signals using this tensor dictionary, the tensor sparse code becomes tube
sparsity.

\subsection{Contributions}
{The main contributions of this chapter are as follows:}

\begin{enumerate}

    \item \textbf{A framework for simultaneously denoising and classifying 2-D
    UWB SAR imagery}\footnote{The preliminary version of this work was presented
    in IEEE Radar Conference, 2017~\cite{vu2017tensor}}. Subtle features from
    targets of interest are directly learned from their SAR imagery. The
    classification also exploits polarization diversity and consecutive aspect
    angle dependence information of targets.


    \item \textbf{A generalized tensor discriminative dictionary learning} 
    (TensorDL) is also proposed when more training data involved. These
    dictionary learning frameworks are shown to be robust even with high levels
    of noise. 

    \item \textbf{A relative SRC framework} (ShiftSRC) is proposed to deal with
    multi-look data. Low-frequency UWB SAR signals are often captured at
    different views of objects, depending on the movement of the radar carriers.
    These signals contain uniquely important information of consecutive views.
    With ShiftSRC, this information will be comprehensively exploited.
    Importantly, a solution to the ShiftSRC framework can be obtained by an
    elegant modification on the training dictionary, resulting in a tensor
    sparse coding problem, which is similar to a problem proposed in 
    contribution 1).

\end{enumerate}

The remainder of this chapter is organized as follows. Section II presents
different tensor sparsity frameworks and  the discriminative tensor dictionary
learning scheme for multi-channel classification problems. The ShiftSRC for
multiple-relative-look and solutions to all proposed frameworks are also presented
in this section. Section III shows extensive experimental results on a
simulated dataset for several scenarios. An experiment with a realistic dataset
is also included. Section IV concludes the chapter.

\section{Sparse representation-based classification} 
\label{sec:sparse_representation_based_classifications}

\subsection{Notation} 
\label{sub:notation}
Scalars are denoted by italic letters and may be either lower or uppercase,
e.g., $d, N, k$. Vectors and matrices are denoted by bold lowercase ($\mathbf{x,
y}$) and bold upper case ($\mathbf{X, Y}$), respectively. In this chapter, we also
consider 3-D tensors (tensors for short) whose dimensions are named \textit{row,
column,} and \textit{channel}. A tensor with only one column will be denoted by
a bold, lowercase, calligraphic letter ($\bx, \by$). Tensors with more than one
column will be denoted by an bold, uppercase, calligraphic letters ($\bX, \bY,
\bD$).

For any tensor $\bM$, let $\bM^{(t)}$ be its $t$-th channel. For convenience,
given two tensors $\bM, \bN$, the tensor multiplication $\bP = \bM\bN$ is
considered channel-wise multiplication, i.e.,  $\bP^{(t)} = \bM^{(t)}\bN^{(t)}$.
For a tensor $\bM$, we also denote the sum of  square of all elements by
$\|\bM\|_F^2$ and the sum of absolute values of all elements by $\|\bM\|_1$.
Tensor addition/subtraction simply represents element-wise addition/subtraction.
Each target sample is represented by a UWB SAR image formed using either a
single (using co-pol) or multiple polarization (using both co-pol and cross-pol)
channels.  Thus, one target sample is denoted by  $\by \in \R^{d\times 1\times
T}$, where $d$ is the total number of image pixels and $T$ is the number of
polarization channels. A collection of $N$ samples is denoted by $\bY \in \R^{d\times N\times
T}$.

{Consider} a general classification problem with $C$ different
classes. Let $\bD_c (1 \leq c \leq C)$ be the collection of all training samples
from class $c$, $\bD_0$ be the collection of samples in the shared class, and
$\bD = [\bD_1, \dots, \bD_C, \bD_0]$ be the total dictionary with
the concatenation being done at the second dimension (column). In our problem,
the shared class can be seen as the collection of ground images.

\subsection{Classification scheme} 
\label{sub:classification_scheme}


\par

\par
{Using} the definition of tensor multiplication, a sparse representation of
$\by$ using $\bD$ can be obtained by solving
\begin{equation}
\label{eqn:general_opt}
    \bx = \arg\min_{\bx} \frac{1}{2} \|\by - \bD \bx\|_F^2 + \lambda g(\bx)
\end{equation}
\noindent where $\lambda$ is a positive regularization parameter and
$g(\bx)$ is a function that encourages {$\bx$ to be sparse}.
Denote by $\bx^i$ the sparse coefficient of $\by$ on $\bD_i$. Then, the
tensor $\bx$ can be divided into $C+1$ tensor parts
$\bx^1,\bx^2,\dots,\bx^C,\bx^0$.

After solving the sparse coding problem~\eqref{eqn:general_opt}, shared features
(grounds in our problem) are eliminated by taking $\bar{\by} = \by -
\bD_0\bx^0$. Then the identity of one sample $\by$ can be determined by the 
dictionary that provides the minimum residual:
\begin{equation}
    \text{identity}(\by) = \min_{i \in \{1, 2, \dots, C\}} \|\bar{\by} - \bD_i
    \bx^i\|^2_2 \end{equation}

\textit{Confuser detection:} In practical problems, the set of confusers is not
limited to the training set. A confuser can be anything that is not a target; it
can be solely the ground or a capture of an unseen object. In the former case,
the test signal can be well represented by using only the ground $\bD_0$, while in
the latter case, the sparse representation assumption is no longer valid.
Therefore, one test signal $\by$ is classified as a confuser if one {of}
following three conditions {is satisfied}: i) it is not sparsely interpreted by the
total dictionary $\bD$; ii) it has the most active elements in the sparse code
locating at $\bx^0$; and iii) it is similar to known confusers.

\subsection{Generalized sparse representation-based classification} 
\label{sub:general_sparse_representation_based_classifications}
\par

\begin{figure}[t]
    \centering
    \includegraphics[width = 0.65\textwidth]{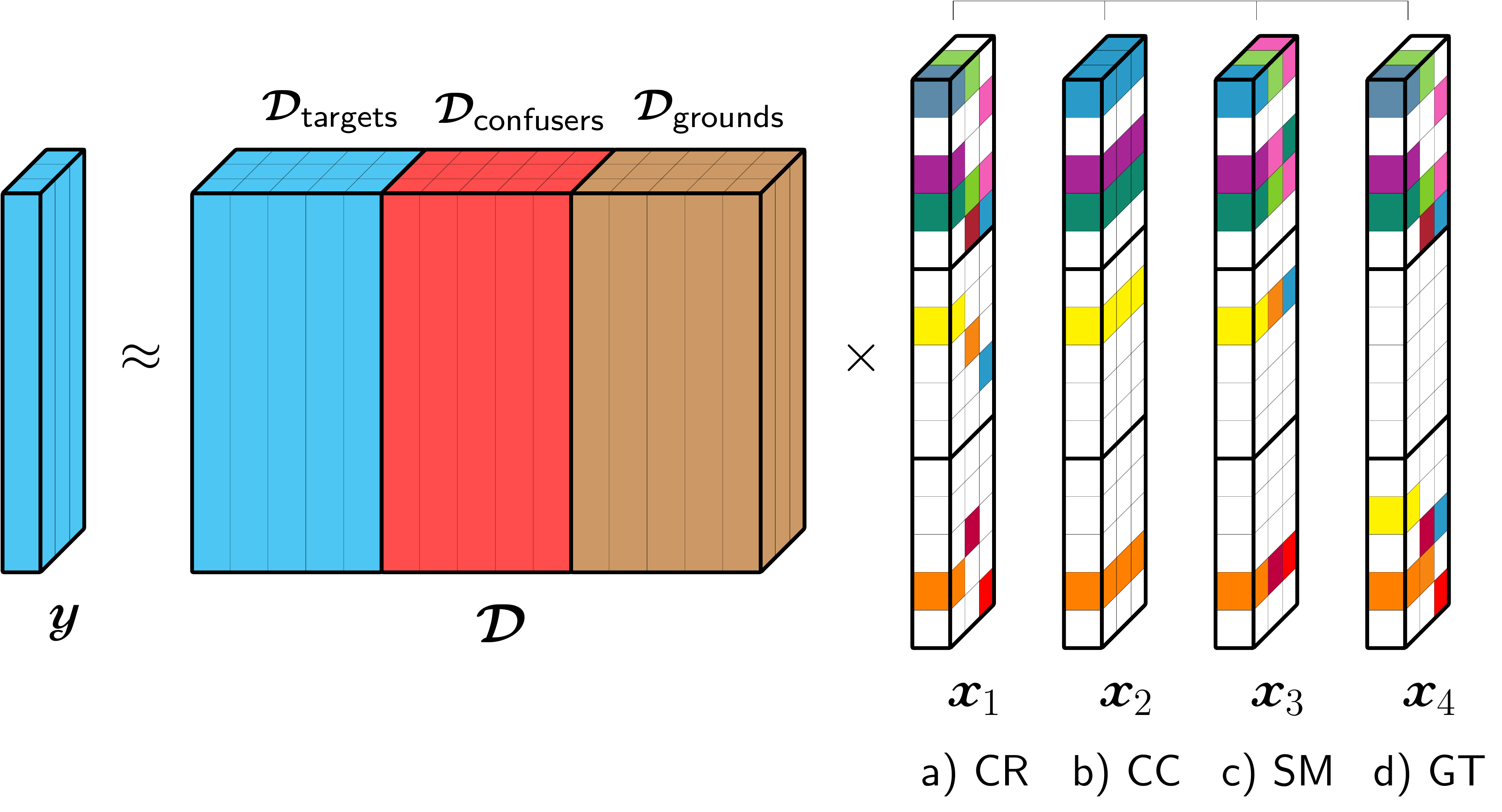}
    \caption{\small  Different sparsity constraints on the coefficient tensor
    $\bx$.}
    \label{fig2}
\end{figure}



In SRC~\cite{Wright2009SRC} where only one channel is considered, $g(\bx)$ is
simply a function that forces sparsity as $l_0$- or $l_1$-minimization.
$l_1$-minimization is often used since it leads to a convex optimization problem
with tractable solutions. In the following, we present two natural extensions of
SRC for multi-channel cases. We also proposed two tensor sparse representation
methods that can enhance classification accuracy by exploiting
cross-channel information. These four generalized SRC methods are as follows.

\textit{a) Apply a sparsity constraint on each column of the coefficient
tensor $\bx$, no cross-channel constraint}.

We can see that this is similar to solving $T$ separate sparse coding
problems, each for a polarization channel. The classification rule is
executed based on the sum of all the squares of the residuals. We refer
to this framework as SRC-cumulative residual or SRC-CR (CR in the short
version). (See Figure~\ref{fig2}a with sparse tensor $\bx_1$). The
sparse code corresponding to the $t^{\text{th}}$ channel,
$\bxt$, is computed via the traditional $l_1$-minimization:
\begin{equation}
    \bxt = \arg\min_{\bxt} \frac{1}{2} \|\byt - \bDt\bxt\|_2^2 + \lambda\|\bxt\|_1
\end{equation}
which can be solved effectively using FISTA~\cite{beck2009fast},
ADMM~\cite{boyd2011distributed}, etc., algorithms or the SPAMS
toolbox~\cite{SPAMS}.

\textit{b) Concatenate all channels}.

The most convenient way to convert a multi-channel problem to a single-channel problem is
to concatenate all $T$ channels of one signal to obtain a long vector. By doing
so, we have the original SRC framework with $l_1$-norm minimization. After
solving the sparse coding problem, if we \textit{break} the long vectors and
rearrange them back to tensor forms, then the tensor sparse coefficients $\bx$
can be formed by replicating the one-channel sparse coefficient at all channel
(see Figure~\ref{fig2}b with sparse tensor $\bx_2$). From this tensor
viewpoint, the code tensor $\bx$ will have few active ``tubes''; moreover, all
elements in a tube are the same. We refer to this framework as SRC-concatenation
or SRC-CC (CC in the short version).

The optimization problem in SRC-CC and its solution are very straightforward.
First, we stack all channel dictionaries into a long one: $\hat{\bD} =
[\bD^{(1)}; \bD^{(2)}; \dots; \bD^{(T)}]$ (symbol `;' represents the
concatenation in the first dimension). Then for every test signal, we also stack
all of its channels to form a long vector: $\hat{\by} = [\by^{(1)}; \by^{(2)};
\dots; \by^{(T)}]$. The optimization problem \eqref{eqn:general_opt} becomes the
traditional $l_1$ regularization problem:
\begin{equation}
    \bx = \arg\min_{\bx}\frac{1}{2} \|\hat{\by} - \hat{\bD}\bx\|_2^2 + \lambda\|\bx\|_1
\end{equation}
then also can be solved by aforementioned methods.

\textit{c) Use a simultaneous model}.

\par Similar to the SHIRC model proposed in~\cite{Srinivas2013,
Srinivas2014SHIRC}, we can impose one constraint on active elements of tensor
$\bx$ as follows: $\bx$ also has few nonzero tubes as in  SRC-CC; however,
elements in one active tube are not necessarily the same. In  other words, the
locations of nonzero coefficients of training samples in the  linear combination
exhibit a one-to-one correspondence across channels. If the  $j$-th training
sample in $\bD^{(1)}$ has a nonzero contribution to $\by^{(1)}$,  then for $t
\in \{2, \dots, T\}$, $\byt$ also has a nonzero contribution from the $j$-th
training sample in $\bDt$. We refer to this framework as SRC-Simultaneous  or
SRC-SM (SM in the short version). (See Figure~\ref{fig2}c with  sparse
tensor $\bx_3$). To achieve this requirement, we can impose on the tensor
$\bx_3$ (with one column and T channels) the $l_{1,2}$-minimization constraint,
which is similar to the row-sparsity constraint applied on matrices
in~\cite{zhang2012multi}.

\par \textbf{Remarks:} While SHIRC uses $l_0$-minimization on $\bx$ and
applies the modified SOMP~\cite{tropp2006algorithms}, our proposed SRC-SM
exploits the flexibility of $l_{1,2}$-regularizer since it is convex, and easily
modified when more constraints are present (e.g., non-negativity). In
addition, it is more efficient especially when dealing with problems of
multiple samples at input.

Concretely, the optimization problem of SRC-SM can be written in the form
\begin{equation}
    \label{eqn:src_sm}
    \bx = \arg\min_{\bx} \frac{1}{2} \|\by - \bD\bx\|_F^2 + 
            \lambda \sum_{k=1}^K \|\vec(\bx_{k::})\|_2
\end{equation}
where $\bx_{k::}$ denotes the $k^{\text{th}}$ \text{tube} of the tensor code
$\bx$ and $K$ is the total column of the dictionary $\bD$, and $\vec(\bx_{k::})$ 
is the vectorization of the tube $\bx_{k::}$. This problem is similar to the 
joint sparse representation (JSRC) problem proposed in~\cite{zhang2012multi}
except that SRC-SM enforces tube-sparsity instead of the row-sparsity.
Details of the algorithm {that solves~\label{eqn:src_sm} are} described in Section~\ref{sec:solution}.

\newpage
\textit{d) Use a group tensor model.}

Intuitively, since one object is ideally {represented} by a linear combination of
the corresponding dictionary and the shared dictionary, it is highly {likely}
that number of active (tensor) parts in $\bx$ is small, i.e., most of $\bx^1,
\dots, \bx^C, \bx^0$ are zero tensors. This suggests us a \textit{group tensor}
sparsity framework as an extension of~\cite{Yu2011} that can improve the
classification performance, which is referred to as SRC-GT (GT in the short
version). The visualization of this idea is shown in
Figure~\ref{fig2}d.

The optimization problem of SRC-GT is similar to \eqref{eqn:src_sm} with a slight
difference in the grouping coefficients:
\begin{equation}
    \label{eqn:src_gt}
    \bx = \arg\min_{\bx} \frac{1}{2} \|\by - \bD\bx\|_F^2 + 
            \lambda \sum_{c=1}^{C+1} \|\vec(\bx^c)\|_2
\end{equation}
where $C+1$ is the total number of classes (including the shared ground class), 
and $\vec(\bx^c)$ is the vectorization of the group tensor $\bx^c$. Solution to 
this problem will be discussed next.



The overall algorithm of generalized SRC applied to multi-channel signals in
the presence of a shared class is shown in Algorithm~\ref{alg:generalSRC}.

\begin{algorithm}[!t]
    \caption{Generalized SRC with a shared class}
    \label{alg:generalSRC}
    \begin{algorithmic}
    \Function {identity$(\by)$ = \\~~~~~~~~~~$~~~~~~~~~~\text{GENERALIZED\_SRC}(\by, \bD, \lambda,  g(\bullet), \varepsilon, \tau)$ }{}
    \State \textbf{INPUT}: \\
    $\by \in \R^{d \times 1 \times T}$ -- a test sample; \\
    $\bD = [\bD_1, \bD_2, \dots, \bD_C, \bD_0] \in \R^{d\times K \times T}$ -- the total dictionary with the shared dictionary $\bD_0$; \\
    $ g(\bullet)$ -- the sparsity constraint imposed on sparse codes.\\
    $\lambda \in \R^{+}$ -- a positive regularization parameter;\\
    $\varepsilon, \tau$ -- positive thresholds.
    \State \textbf{OUTPUT}: the identity of $\by$.
    \State 1. Sparsely code $\by$ on $\bD$ via solving:
   \begin{equation}
       {\bx} = \arg\min_{\bx} \{\|\by - \bD\bx\|_F^2 + \lambda g(\bx)\}
       \label{eqn: src}
   \end{equation}
   \State 2. Remove the contribution of the shared dictionary: $$\displaystyle
   \bar{\by} = \by - \bD_0\bx^0.$$
\State 3. Calculate the class-specific residuals :
{ $$\displaystyle r_c = \|\bar{\by} - \bD_c\bx^c\|_2, \forall c = 1, 2, \dots, C.$$}
    \State 4. Decision:
    \If{$\displaystyle\min_c(r_c) > \tau$ ({\it an unseen object}) \textbf{or} $\|\bar{\by}\|_2 < \varepsilon$ ({\it a ground})}
        \State $\by$ is a confuser.
    \Else
        \State $\displaystyle \text{identity}(\by) = \arg\min_{c}\{r_c\}$
   \EndIf
    \EndFunction
    \end{algorithmic}
\end{algorithm}


\subsection{Dictionary learning for tensor sparsity} 
\label{sub:dictionary_learning_for_tensor_sparsity}
As a natural extension, we can extend the tensor sparsity models to
dictionary learning ones. Most of dictionary learning methods focus on a
single-channel signal, which is not suitable for models with cross-channel
information. In this work, we extend single-channel dictionary learning
methods to multi-channel dictionary ones by applying aforementioned tensor
sparsity constraints.

\def\sbx{\mathbf{x}}
\def\sbX{\mathbf{X}}
\def\sbD{\mathbf{D}}
\def\sbY{\mathbf{Y}}

In single-channel, most of discrimination dictionary learning methods, such as 
FDDL~\cite{yang2014sparse}, DLSI~\cite{ramirez2010classification}, 
DFDL~\cite{vu2016tmi}, LRSDL~\cite{vu2016fast}, etc., have a cost function {that is of the form}
\begin{equation}
    \bar{J}_{\sbY}(\sbD, \sbX) = \bar{f}_{\sbY}(\sbD, \sbX) + \lambda \bar{g}
    (\sbX)
\end{equation}
where $\bar{g}(\sbX)$ is a function enforcing the sparsity of $\sbX$, and $
\bar{f}_{\sbY} (\sbD, \sbX)$, which includes fidelity and discriminant terms, is 
a function of $\sbD, \sbX$ and depends on the training samples $\sbY$.

One straightforward extension of these single-channel models to a multi-channel
case is to apply the first term $f_{\sbY}(\sbD, \sbX)$ to each channel and 
\textit{join} all channels by a sparsity constraint represented by $g(\sbX)$.
Naturally, $g(\sbX)$ can be one of four presented cross-channel sparsity
constraints. Concretely, the overall cost function would be in the form
\begin{equation}
    {J}_{\bY}(\bD, \bX) = {f}_{\bY}(\bD, \bX) + \lambda g(\bX)
\end{equation}
where $f_{\bY^{(t)}}(\bD^{(t)}, \bX^{(t)}) = \bar{f}_{\bY^{(t)}}(\bD^{(t)}, \bX^
{(t)})$ and $g(\bX)$ is one of \{CR, CC, SM, GT\} sparsity constraints. 

In this chapter, we particularly focus on extending FDDL~\cite{yang2014sparse} to
the multi-channel case. FDDL is a special case of LRSDL~\cite{vu2016fast} without
an explicit shared dictionary. FDDL is chosen rather than
LRSDL since in our problem, the shared dictionary, e.g., grounds, is already
separated out. We also adopt  fast and efficient algorithms proposed in the
dictionary learning toolbox DICTOL~\cite{vu2017dictol} to update each channel of
the dictionary $\bD$. Also, the sparse code tensor $\bX$ is updated using
FISTA~\cite{beck2009fast} algorithm, which is discussed in Section
\ref{sec:solution}.

The proposed cross-channel dictionary learning method is named TensorDL suffixed
by CR, CC, SM, or GT when different sparsity constraints are applied on $\bX$.

\begin{figure}[t]
    \centering
    \includegraphics[width = 0.48\textwidth]{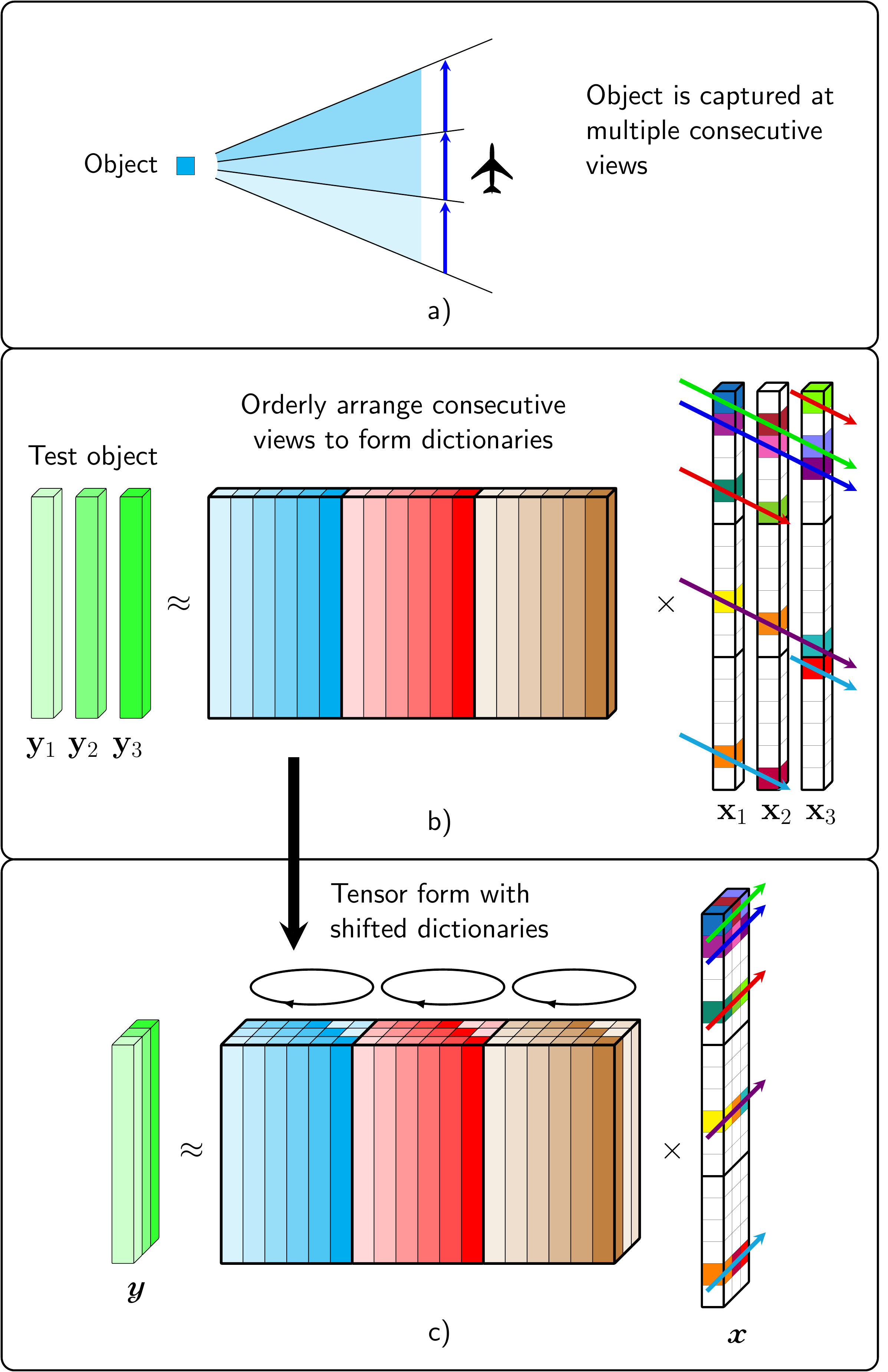}
    \caption{\small Tensor sparsity with relative views.}
    \label{fig3}
\end{figure}

\subsection{Tensor sparsity with multiple relative looks} 
\label{sub:tensor_sparsity_with_multi_relative_views}

\def\sby{\mathbf{y}}

In realistic situations, objects might be captured at different angles instead
of only one. Moreover, these are often consecutive angles that a plane or a
vehicle can capture (see Figure~\ref{fig3}a left) while moving
around objects. Based on this observation, in the training phase, we collect
data from different views of objects, as in Figure~\ref{fig3}a
right, and orderly arrange them into dictionaries for each object, as
illustrated in Figure~\ref{fig3}b. The test objects, which are
often captured at different relative views $\sby_1, \sby_2, \sby_3$, are then
sparsely coded by the whole dictionary.

Intuitively, if $\sbx_1$ is the sparse code of the first view $\sby_1$ with only
few active elements, then the sparse code $\sbx_2$ of the next view $\sby_2$
will be active at locations shifted by one. Similarly, active locations in
$\sbx_3$ of the third view will be shifted by two compared to $\sbx_1$, and so
on. In this case, active coefficients form a ``stair'', as illustrated in
Figure~\ref{fig3}b right.

The sparse coding problem with the ``stair'' sparsity is a novel problem and has
{not} been addressed before. In this chapter, we propose a method called ShiftSRC
to convert this problem to a {previously known} problem. The central idea is that if we
stack all views into a tensor and ``circularly shift'' the ordered dictionary by
one to the left at each view, then the obtained tensor code will be a tensor
with active elements forming a ``tube'' sparsity (see
Figure~\ref{fig3}c). The solution to this problem is similar to the
solution of SRC-SM as stated in this chapter.


\subsection{{Solution to optimization problems}} 
\label{sec:solution}

{ Both optimization problems \eqref{eqn:src_sm} and
\eqref{eqn:src_gt} have the form
\begin{equation}
    \label{eqn:common_fista}
    \bx = \arg\min_{\bx}\{F(\mathbcal{x}) \equiv f(\bx) + \lambda g(\bx)\},
\end{equation}
where
\begin{itemize}
    \item $g(\bx)$ is sum of norm 2, then it is a continuous convex function
    and {\em nonsmooth}.
    \item $f(\bx) = \frac{1}{2} \|\by - \bD\bx\|_F^2$ is a continuous convex
    function of the type $C^{1, 1}$, i.e., continuously differentiable with a
    Lipschitz continuous gradient $L$:
    \begin{equation*}
        \|\nabla f(\mathbcal{x}_1) -\nabla f(\mathbcal{x}_2)\|_F \leq L
        \|\bx_1 - \bx_2\|_F ~~ \text{for every}~ \bx_1, \bx_2.
    \end{equation*}
\end{itemize}
}
We observe that, with these properties, the optimization problem
\eqref{eqn:common_fista} can be solved by FISTA~\cite{beck2009fast}. There
are three main tasks in FISTA:\\

\noindent 1. Calculating $\nabla f(\bx)$, which can be easily computed
as $$\nabla f(\bx) = \bD^T(\bD\bx - \by).$$\\
\noindent where each channel of $\bD^T$ is the transpose of the corresponding
channel of
$\bD$.

\begin{algorithm}[t]
\label{alg:LRSDLX}
    \caption{{Tensor sparse coding by FISTA\cite{beck2009fast}}}
    \begin{spacing}{1.3}
    \begin{algorithmic}
    \Function {${\bx}$ = TENSOR\_SC}{$\by, \bD, \bx_{\text{init}}, \lambda$}.
    \State 1. Calculate
    \begin{align*}
        \bA &= \bD^T\bD, ~~~\bb = \bD^T\by \\
        L & = \max_{t = 1, 2, \dots, T} (\lambda_{\max}(\bA^{(t)}))
    \end{align*}
    \State 2. Initialize $\bx_0 = \bx_{\text{init}}$, $\bw_1 = \bx_0$, $j = 1, t_1 = 1$
    \While {non convergence and $j < j_{\max}$}
        \State 3. Calculate gradient: $\bg = \bA\bw_j - \by.$
        \State 4. Calculate $\bu = \bw_j - \bg/L.$
        \State 5. If SRC-SM, $\bx_j$ is the solution of \eqref{eqn:sm_norm2};
            if SRC-GT, $\bx_j$ is the solution of \eqref{eqn:gt_norm2}.
        \State 6. $t_{j+1} = (1 + \sqrt{1 + 4t_j^2})/2$

        \State 7. $\bw_{j+1} = \bx_j + \frac{t_j - 1}{t_{j+1}} (\bx_j - \bx_{j-1})$
        \State 8. $j = j + 1$

    \EndWhile
    \State 9. OUTPUT: $\bx = \bx_j$
    \EndFunction
    \end{algorithmic}
    \end{spacing}
\end{algorithm}

\noindent 2. Calculating a Lipschitz constant of $\nabla f(\bx)$. For our
function $f$, we can choose
\begin{equation}
    L = \max_{t = 1, 2, \dots, T} \left\{\lambda_{\max}\left((\bD^{(t)})^T\bDt\right)\right\}
\end{equation}
where $\lambda_{\max}$ is the maximum eigenvalue of a square matrix.

\noindent 3. Solving a suboptimization problem of the form
\begin{equation}
\label{eqn:fista_approx}
    \bx = \arg\min_{\bx} \left\{\frac{1}{2} \|\bx - \bu\|_F^2 + \eta g(\bx)\right\}
\end{equation}
with $\eta = \frac{\lambda}{L}$.

For SRC-SM, problem \eqref{eqn:fista_approx} has the form
\begin{eqnarray}
\nonumber
    \bx = \arg\min_{\bx} \left\{\frac{1}{2} \|\bx - \bu\|_F^2 + \eta \sum_{k=1}^K \|\vec(\bx_{k::})\|_2\right\} \\
    \label{eqn:sm_norm2} =  \arg\min_{\bx} \left\{ \sum_{k = 1}^K
    \left(\frac{1}{2} \|\bx_{k::} - \bu_{k::}\|_2^2 + \eta
    \|\vec(\bx_{k::})\|_2\right)\right\}
\end{eqnarray}
Each problem in \eqref{eqn:sm_norm2} is a minimum $l_2$-norm minimization with solution being
\begin{equation}
\label{eqn:sol_sm}
    \bx_{k::} = \max\left\{1 - \frac{\eta}{\|\vec(\bu_{k::})\|_2}, 0\right\}\bu_{k::}, \forall k = 1, 2, \dots, K.
\end{equation}
Similarly, for SRC-GT, problem \eqref{eqn:fista_approx} can be written as
\begin{eqnarray}
\nonumber
    \bx = \arg\min_{\bx} \left\{\frac{1}{2} \|\bx - \bu\|_F^2 + \eta \sum_{c=1}^{C+1}\|\vec(\bx^c\|_2\right\} \\
    \label{eqn:gt_norm2}
    =  \arg\min_{\bx} \left\{ \sum_{c = 1}^{C+1} \left(\frac{1}{2}
        \|\bx^c - \bu^c\|_2^2 + \eta \|\vec(\bx^c)\|_2\right)\right\}
\end{eqnarray}
with solution being:
\begin{equation}
\label{eqn:sol_gt}
    \bx^c = \max\left\{1 - \frac{\eta}{\|\vec(\bu^c)\|_2}, 0\right\}\bu^c,
    \forall c = 1, \dots, C+1.
\end{equation}

\begin{figure}
\centering
\includegraphics[width = .45\textwidth]{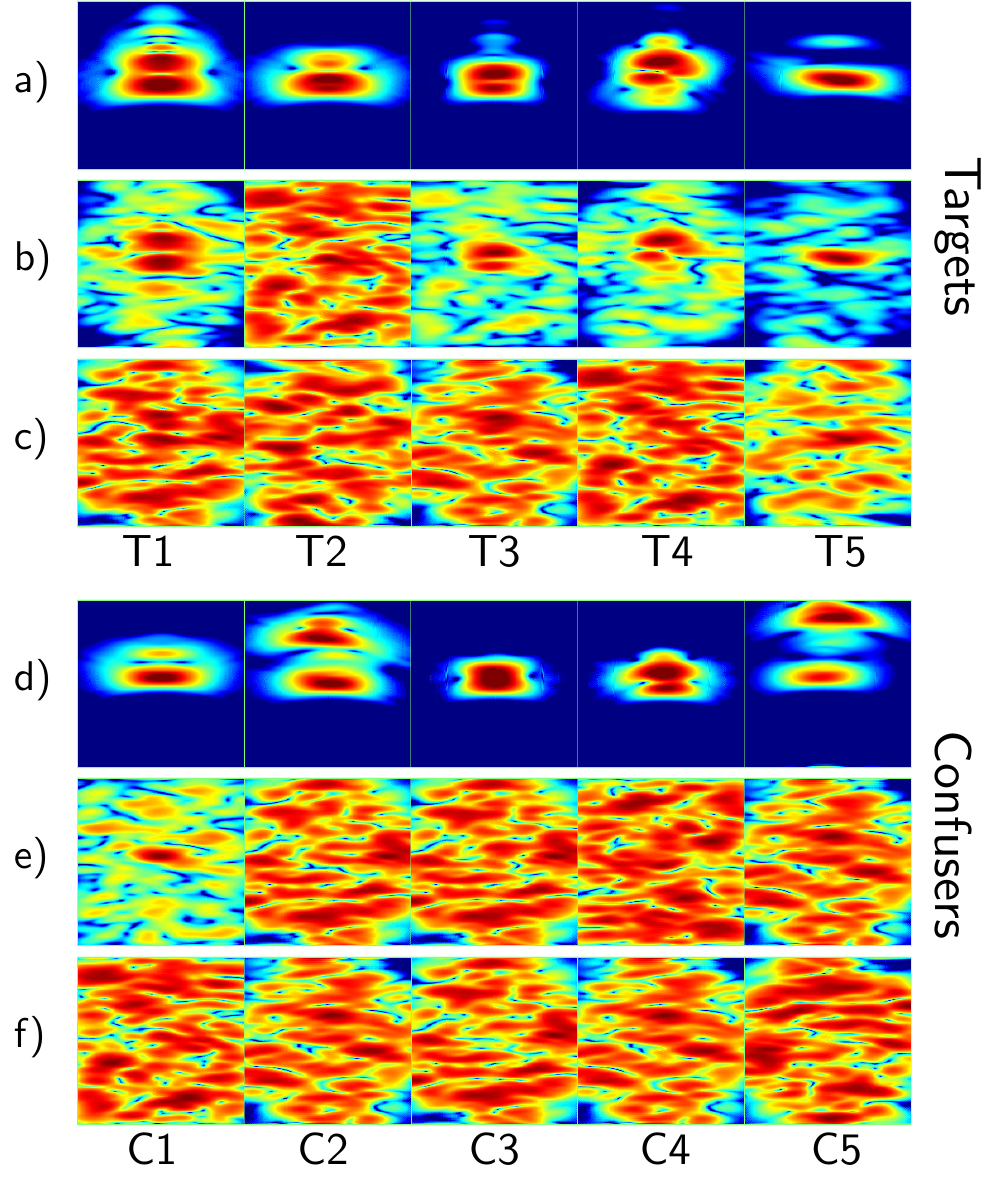}
\caption{Sample images of five targets and five clutter objects. T1 = M15
anti-tank mine, T2 = TM62P3 plastic mine, T5 = 155-mm artillery shell, C1 =
soda can, C2 = rocks, C3 = rocks, C4 = rocks, C5 = rocks. a) Targets under
smooth ground surface. b) Targets under rough ground surface (easy case,
scale = 1). c) Targets under rough ground surface (hard case, scale = 5). d)
Confusers under smooth ground surface. e) Confusers under rough ground
surface (easy case, scale=1). f) Confusers under rough ground surface (hard
case, scale=5).}
\label{fig4}
\end{figure}

{A step by step description of SRC-SM and SRC-GT algorithms are}
given in Algorithm 2.



\section{Experimental results} 
\label{sec:experimental_results}

\begin{figure}
\centering
\includegraphics[width = .69\textwidth]{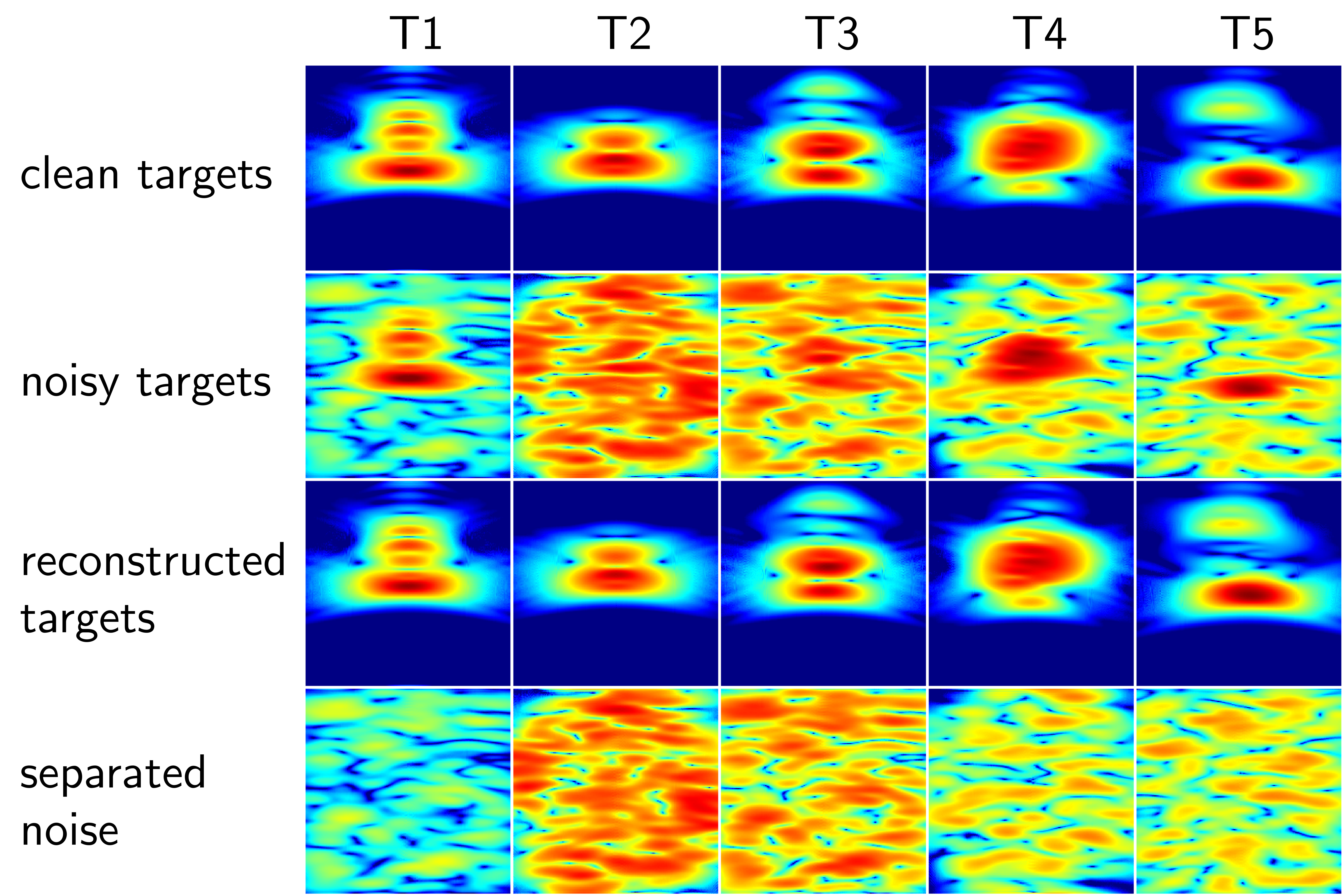}
\caption{Visualization of decomposed signals (HH polarization) after doing
sparse coding. Row 1: the original clean signals. Row 2: the corresponding
noisy signals with the interference of grounds. Row 3: reconstructed signal
after sparse coding. Row 4: separated noises.}
\label{fig5}
\end{figure}




In this section, we apply the proposed models to the problem of classifying
objects of interest. Extensive results are presented on simulated and real-life
datasets. A MATLAB toolbox for the tensor sparsity methods presented in this chapter
is available at~\cite{vu2017tensorsparsity}.
\begin{table*}[]
\centering
\footnotesize
\caption{Overall average classification (\%) of different methods on different
polarization combinations, with or without the non-negativity constraint (NN). }
\label{tab:tab1}
\begin{tabular}{c|c|c|c|c|c||c|c|c|c|c|c}
\cline{2-11}
                       & \multicolumn{5}{c||}{Separate targets} & 
                       \multicolumn{5}{c|}{All targets} &                                                                                                                    \\ \cline{1-11}
\multicolumn{1}{|l||}{} & VV   & HH   & VV+HH   & HH+HV  & ALL  & VV  & HH  & VV+HH  & HH+HV  & ALL  &                                                                                                                \\ \cline{1-11}
\multicolumn{1}{|l||}{SVM}    &62.17 &74.47 &65.00 &77.77 &70.33 &83.68 &89.36 &83.36  &66.36 &79.78 & \\ \hline \hline
\multicolumn{1}{|l||}{CR} &76.32 &83.43 &86.26 &85.93 &82.95 &86.22 &91.64
&87.14  &91.28 &90.64 &\multicolumn{1}{l|}
{\multirow{4}{*}{\begin{tabular}[c]{@
{}l@{}}wo/ \\ NN\end{tabular}}} \\ 
\cline{1-11}
\multicolumn{1}{|l||}{CC} &76.32 &83.43 &79.47 &86.21 &79.93 &86.22 &91.64 &87.96  &92.98 &81.80 & \multicolumn{1}{l|}{}                                                                                              \\ \cline{1-11}
\multicolumn{1}{|l||}{SM} &76.32 &83.43 &81.07 &86.52 &85.87 &86.22 &91.64 &89.66  &95.12 &94.14 & \multicolumn{1}{l|}{}                                                                                              \\ \cline{1-11}
\multicolumn{1}{|l||}{GT} &69.79 &78.79 &77.83 &85.12 &81.75 &81.90 &88.40 &87.28  &91.86 &90.46 & \multicolumn{1}{l|}{}                                                                                              \\ \hline \hline
\multicolumn{1}{|l||}{CR} &{\bf 79.05} &{\bf89.41} &79.85 &88.52 &84.92 &
{\bf86.32} &{\bf92.98} &87.48  &94.86 &94.26 & \multicolumn{1}{l|}{\multirow{4}
{*}{\begin{tabular}[c]{@{}l@{}}w/ \\ NN
\end{tabular}}}    \\ \cline{1-11}
\multicolumn{1}{|l||}{CC} &{\bf 79.05} &{\bf89.41} &83.99 &92.27 &85.11 &{\bf86.32} &{\bf92.98} &89.20  &93.82 &87.08 & \multicolumn{1}{l|}{}                                                                                              \\ \cline{1-11}
\multicolumn{1}{|l||}{SM} &{\bf 79.05} &{\bf89.41} &{\bf85.62} &{\bf90.85} &{\bf89.36} &{\bf86.32} &{\bf92.98} &{\bf90.00}  &{\bf96.82} &{\bf96.00} & \multicolumn{1}{l|}{}                                                                                              \\ \cline{1-11}
\multicolumn{1}{|l||}{GT} &75.55 &88.97 &81.85 &90.57 &87.28 &84.16 &90.32 &88.10  &94.58 &94.12 & \multicolumn{1}{l|}{}                                                                                              \\ \hline
\end{tabular}
\end{table*}
\subsection{Electromagnetic (EM) Simulation data}
SRC is applied to a SAR database consisting of targets (metal and plastic
mines, 155-mm unexploded ordinance [UXO], etc.) and clutter objects (a soda can,
rocks, etc.) buried under rough ground surfaces. The electromagnetic (EM) radar
data {is} simulated based on the full-wave computational EM method known as
the finite-difference, time-domain (FDTD) software~\cite{dogaru2010}, which was
developed by the U.S. Army Research Laboratory (ARL). The software was validated
for a wide variety of radar signature calculation
scenarios~\cite{dogaru2007,liao2012}. Our volumetric rough ground surface grid
-- with the embedded buried targets -- was generated by using the surface root-
mean-square (rms) height and the correlation length parameters. The  targets are
flush buried at 2-3 cm depth. In our experiments, the easiest case  of rough
ground surface in our experiments, the surface rms is 5.6 mm and the correlation
length is 15 cm. The SAR images of various targets and clutter objects are
generated from EM data by coherently integrated individual radar return signals
along over a range of aspect angles. The SAR images are formed using the
backprojection image formation~\cite{McCorkle1994} with an integration angle of
$30^{\circ}$. Figure~\ref{fig4}a shows the SAR images {(using vertical transmitter, vertical receiver -- VV --
polarization)} of some targets that are buried under a perfectly smooth ground
surface. Each target is imaged at a random viewing aspect angle and an
integration angle of $30^{\circ}$. Figures~\ref{fig4}b and
~\ref{fig4}c show the same targets as 
Figure~\ref{fig4}a, except that they are buried under a rough
ground surface (the easiest case corresponds to ground scale{/noise level}\footnote{{Note that a higher ground scale means more noisy images. Henceforth, in the text as well as figures we simply refer to ground scale as noise level for ease of exposition.}} = 1 and harder case
corresponds to noise level = 5). Similarly, Figures~\ref{fig4}d,
\ref{fig4}e, and~\ref{fig4}f show the SAR images of
some clutter objects buried under a smooth and rough surface, respectively. For
training, the target and clutter object are buried under a smooth surface to generate high signal-to-clutter ratio images. We include 12 SAR images
that correspond to 12 different aspect angles ($0^{\circ}, 30^{\circ},
\dots, 330^{\circ}$) for each target type. For testing, the SAR images of
targets and confusers are generated at random aspect angles and buried under 
rough ground surfaces. Various levels of ground surface roughness are simulated
by selecting different ground surface scaling factors when embedding the test
targets under the rough surfaces. Thus, the resulting test images are very noisy
with a very low signal-to-clutter ratio. 
{Each image is a polarization signal of object which is formed by one of transmitter-receiver setups: vertical-vertical (VV), horizontal-horizontal (HH), or horizontal-vertical (HV).}
Each data sample of one object is
represented by either i) one SAR image using data from one co-pol (VV, HH)
channel or ii) two or more images using data from co-pol (VV, HH) and cross-pol
(HV) channels. For each target type, we tested 100 image samples measured at
random aspect angles.

\subsection{Denoised signal visualization} 
\label{sub:denoised_signal_visualization}

We construct the dictionary $\bD = \bmt \bD_{t}, \bD_{c}, \bD_{c} \emt$ of 
all three polarizations, with $\bD_t, \bD_c, \bD_g$ being dictionaries of 
\textit{targets},
\textit{confusers}, and \textit{grounds}, respectively. The sub-dictionary $\bD_o
= \bmt \bD_t,
\bD_c \emt$ can be seen as the dictionary of the objects of interest. For
a noisy signal $\by$, we first solve the following problem:
\begin{equation}
    \bx = \arg\min_{\bx} \|\by - \bD\bx\|_F^2 + \lambda g(\bx)
\end{equation}
where $g(\bx)$ is the proposed SM constraint. The sparse tensor code $\bx$ is
then decomposed into two parts, $\bx^{o}$ and $\bx^{\text{g}}$. The latter can
be considered coefficients corresponding to the ground dictionary $\bD_g$. The
original signal can be approximately decomposed into two parts: $\bD_{g}\bx^{g}$
as separated noise, and $\bD_{o}\bx^o$ as the denoised signal. Visualization of
these signals are shown in Figure~\ref{fig5}. We can see that the
tensor framework successfully decompose{s} noisy signals into a clean part and a
ground signal. These results show the potential of the proposed tensor sparsity
frameworks for classification.

\subsection{Overall classification accuracy} 
\label{sub:overall_classification_accuracy_in_the_best_condition}
\begin{figure*}[t]
    \centering
    \includegraphics[width = 0.99\textwidth]{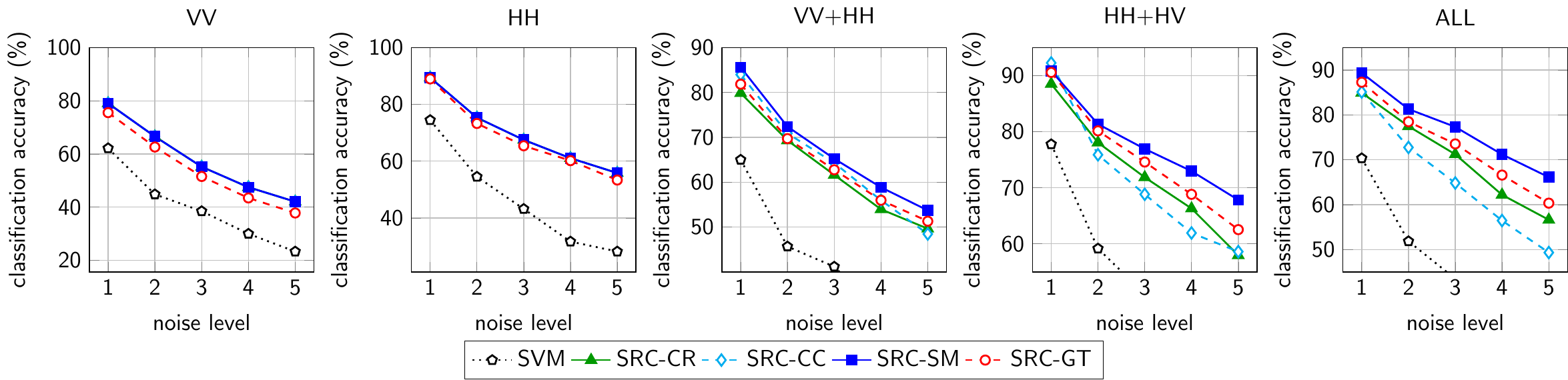}
    \caption{\small Classification accuracy (\%) with different {noise levels} and 
    different polarization combinations for the \textit{separate-target} scenario.}
    \label{fig6}
\end{figure*}

We apply four methods presented
in~\ref{sub:general_sparse_representation_based_classifications} to different
combinations of three polarizations: VV, HH, VV+HH, HH+HV, and VV+HH+HV (or
ALL), and also compare these results with those obtained by {a support vector machine} (SVM) using the libsvm
library \cite{CC01a}. {SVM was also applied to classify UWB
signals~\cite{wang2015classification,bryan2012application}}. The training set
comprises all five target sets, four out of five confuser sets (each time we
leave one confuser set out, which is meant to be unseen), and the ground set.
While all confuser sets can be seen as one class -- \textit{confuser class} --
there are two ways of organizing target sets. First, each of five targets is
considered one class in case the identity of each target is crucial (we name
this scenario \textit{separate-target} with five target classes and one confuser
class). Second, if we only need to know whether an object of interest is a
target or not, we can consider all five targets as one class and its
corresponding name is \textit{all-target} with one target class and one confuser
class. We also consider two families of the sparse coding problem, one with and
one without the non-negativity constraint on the sparse code $\bx$ in each of
the tensor sparse coding methods. Each experiment is conducted 10 times and
their results are reported as the average number. Parameters in each method are
chosen by a 5-fold cross-validation procedure. In this experiment, all test
samples are corrupted by small noise, i.e., the noise level is set to one. {In our experiments, we use overall classification accuracy as the performance metric, which computes percentage of correctly classified samples over total test samples across all classes.}

Overall classification accuracies of different methods on different
polarization combinations are reported in Table~\ref{tab:tab1}. From the table,
a few {inferences} can be made:

\begin{itemize}
    
    \item SRC-related methods with non-negative sparse coefficients perform
    better than those without this constraint\footnote{Based on this
    observation, from now on, all other results are implicitly reported with the
    non-negativity constraint.}. In addition, SVM is outperformed
    by all other methods in all tasks.

    \item SRC-SM provides the best results in all
    combinations for both the \textit{separate-target} and \textit{all-target}
    scenarios. The best accuracies in both cases are significantly high with
    slightly over 90\% in the six-class classification problem and nearly 97\%
    in the binary classification problem. 

    \item If only one polarization is used, SRC-CR, SRC-CC, and SRC-SM
    have identical results, since all of them are basically reduced to
    traditional SRC in this case. Additionally, these three methods slightly
    outperform SRC-GT in this scenario.

    \item If only one polarization can be obtained, HH always outperform{s} VV {and by a}
    significant {margin}. Additionally, the HH+VV combination often worsens
    the results {versus} using HH alone.

    \item If the same method is used on different combinations, the best results
    are mostly obtained by the combination of HH and HV polarizations in both
    the \textit{separate-target} and \textit{all-target} scenarios.

\end{itemize}

\subsection{{Effect of noise levels on overall accuracy}} 
\label{sub:effect_off_corruption_levels_on_overall_accuracy}

The results in the previous section are collected in the presence of small
corruption (noise level is only 1). In real problems, objects of interest are,
deliberately or not, buried under extremely rough surfaces in order to
\textit{fool} classifiers. In this section, we conduct an experiment to see how
each method performs when the level of corruption increases in the
\textit{separate-target} scenario.

Classification results of five different methods on different polarization
combinations and different noise levels are shown in
Figure~\ref{fig6}. First of all, similar trends to small corruption can
be observed in that SRC-SM shows best performance in all cases with bigger gaps
occurring at high levels of noise. In addition, of the four SRC-related methods,
SRC-CC is beaten by all three others when more than one polarization involved.
This can be explained by the fact that SRC-CC suffers from the \textit{curse of
dimensionality} when each sample is represented by concatenating long vectors.
It can also be seen that SRC-GT performs better than SRC-CR and SRC-CC in the
presence of multiple polarizations. Last but not least, the best classification
results can be seen at HH+HV and ALL among the five different combinations
considered.


\subsection{Effect of tensor dictionary learning on overall accuracy} 
\label{sub:effect_of_tensor_dictionary_learning_on_overall_accuracy}
We report the classification results for the
\textit{all-target} scenario with different noise levels. We also include the
results of experiments with discriminative tensor dictionary learning methods.
The results of HH+HV and ALL are reported, since they yield the best results, as
observed in previous sections.

\begin{figure}[t]
\centering
\includegraphics[width = .69\textwidth]{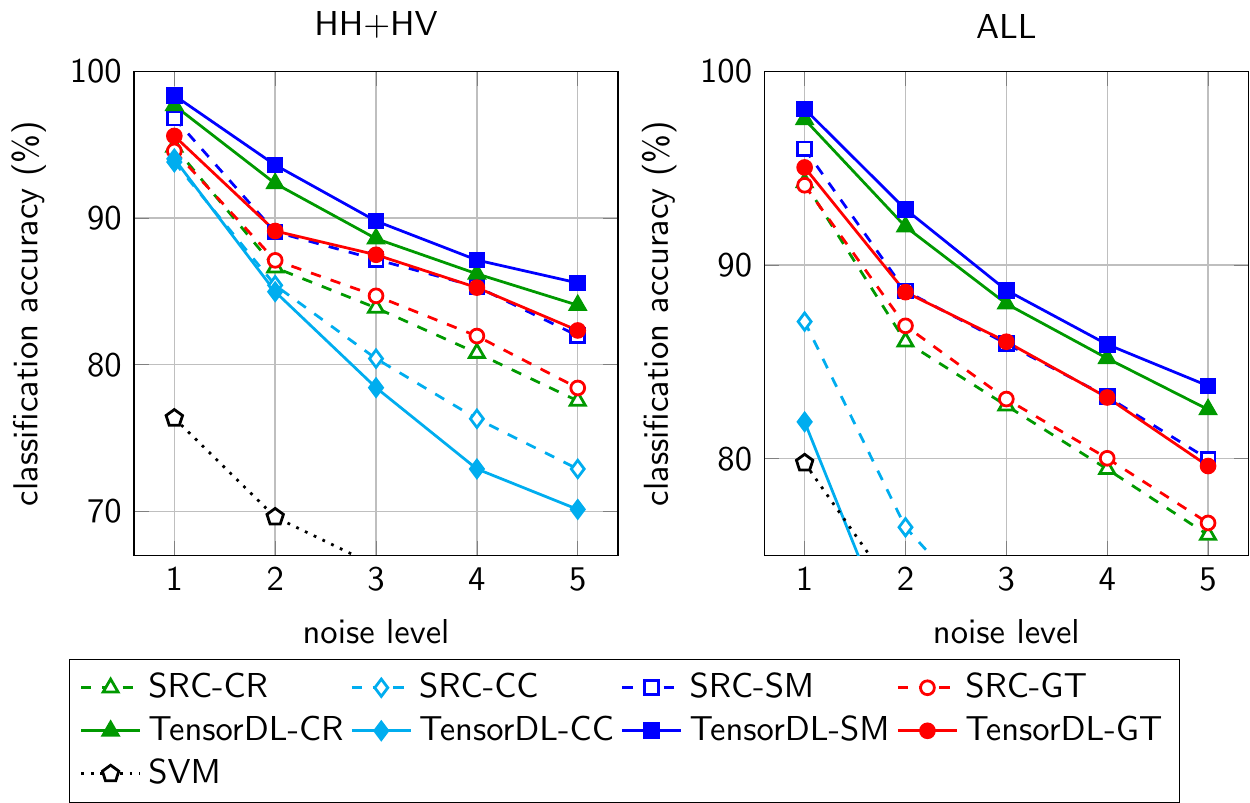}
\caption{\small Classification accuracy (\%) with different {noise levels},
    and a discriminative dictionary learning on the \textit{all-target}
    scenario.}
\label{fig7}
\end{figure}

The results of nine different methods are depicted in Figure~\ref{fig7}.
These methods include SVM (the dotted line), the four SRC-related methods (dashed
lines), and their corresponding tensor dictionary learning counterparts
(prefixed by {TensorDL}, solid lines). We can see that except for the
CC case, tensor dictionary learning methods outperform their
corresponding SRC with gaps widening as
the noise level increases. This observation confirms that tensor discriminative
dictionary learning methods indeed {provide} accuracy improvements. Methods with
the SM constraint {once again emerge} the winner in all noise levels, and the best accuracy numbers are observed using the HH+HV combination.


\subsection{Multiple-relative-look classification accuracy} 
\label{sub:multi_relative_look_classification_accuracy}


\begin{figure}[t]
\centering
\includegraphics[width = .45\textwidth]{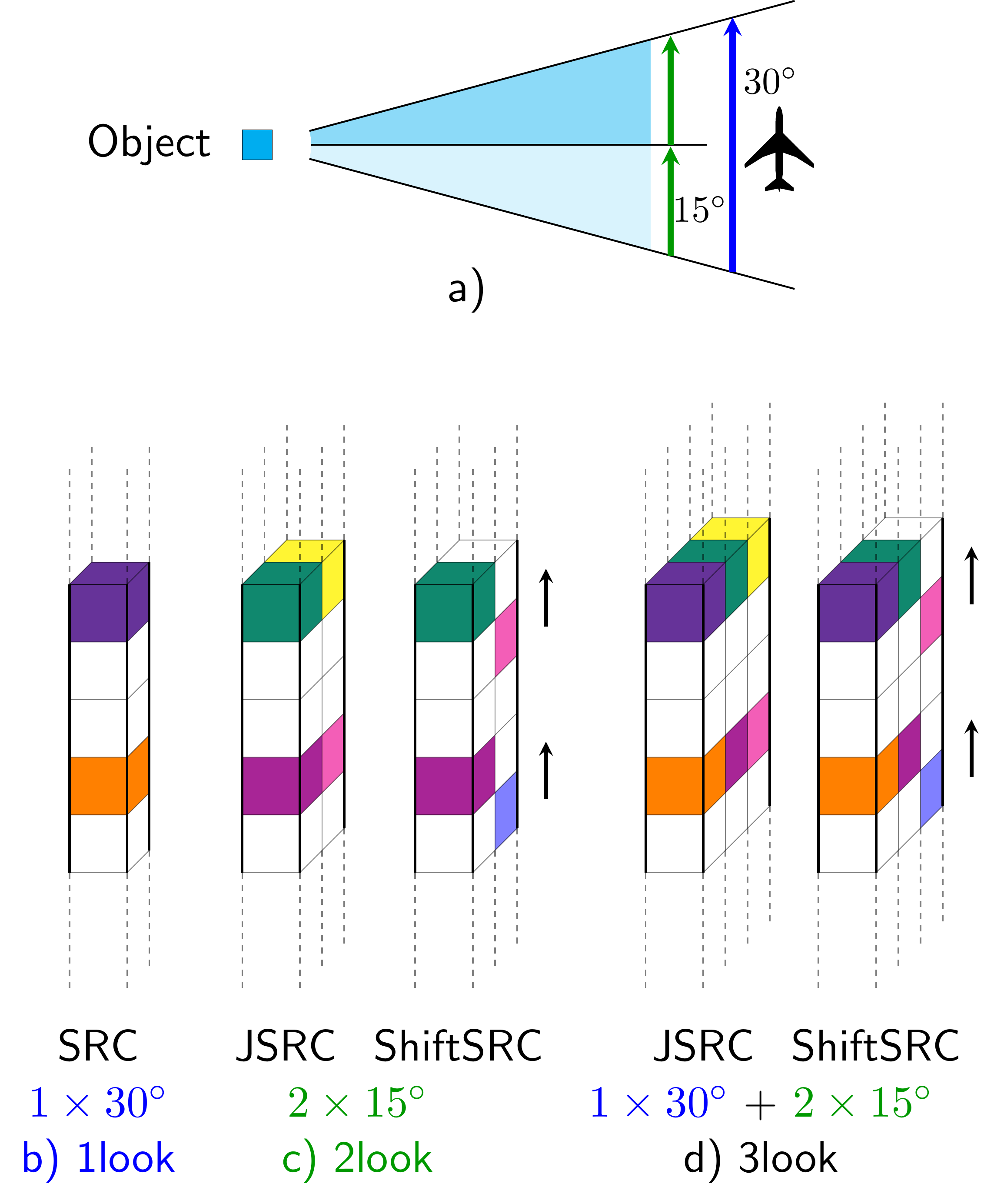}
\caption{\small Multi-relative-look experimental setup (a) and nonzeros locations
of sparse tensors in different scenarios (b, c, d).}
\label{fig8}
\end{figure}

We describe a key real-world scenario where multi-look signals of an object of
interest can be obtained. Figure~\ref{fig8}a depicts the relative
location of a radar carrier, a jet plane in this case, and an object of
interest. The plane moves around the object at an angle corresponding to the
blue arrow. One sample of object can be captured at this whole range, which can
be seen as one of its \textit{looks}. At the same time, the radar
can also capture multiple \textit{looks} of the object at smaller angles
represented by green arrows. By continuing considering even smaller angles, more
representatives of the object can be captured. These multiple views can provide
complementary information of the object, highly likely resulting in better
classification accuracy of the system.

For the training samples, for each object, we generate two sets of signals. Each
set contains samples captured after each $15^{\circ}$, and each set has total
of 24 views. Both sets start at the same relative angle but the first is
captured by an integration angle of $30^{\circ}$, the angle in the second set
is $15^{\circ}$. For the test samples, each object is represented by three
signals: one by an integration angle of $30^{\circ}$ and two others by
an integration angle of $15^{\circ}$, as depicted in Figure~\ref{fig8}a.
Similar to previous experiments, test samples are captured at random relative
angles. Ground samples are also simulated in the same way. Based on three
signals captured, we establish three different way of using this information in
the classification process:

\begin{enumerate}

    \item \textit{1look}: for each object, we use signals at integration angle of
    $30^{\circ}$ only. If only one polarization is used, an object of interest can
    be identified by SRC (and implicitly, SVM). If more than one polarization is
    involved, one object will be determined by SRC-SM, as this is the best method
    based on previous experiments (see Figure~\ref{fig8}b).

    \item \textit{2look}: each object is represented by two singles captured at
    $15^{\circ}$. This multi-look classification problem can be solved by joint
    SRC (JSRC) \cite{zhang2012multi}, or the proposed relative-look SRC
    (ShiftSRC) (see Figure~\ref{fig8}c).

    \item \textit{3look}: uses all three signals to represent an object. In this
    case, the relationship between the $30^{\circ}$ signal and the first
    $15^{\circ}$ signal can be modeled by the SRC-SM, while the relationship
    between two $15^ {\circ}$ signals can be formed by either JSRC or ShiftSRC
    (see Figure~\ref{fig8}d).

\end{enumerate}

\begin{figure}[t]
    \centering
    \includegraphics[width = .69\textwidth]{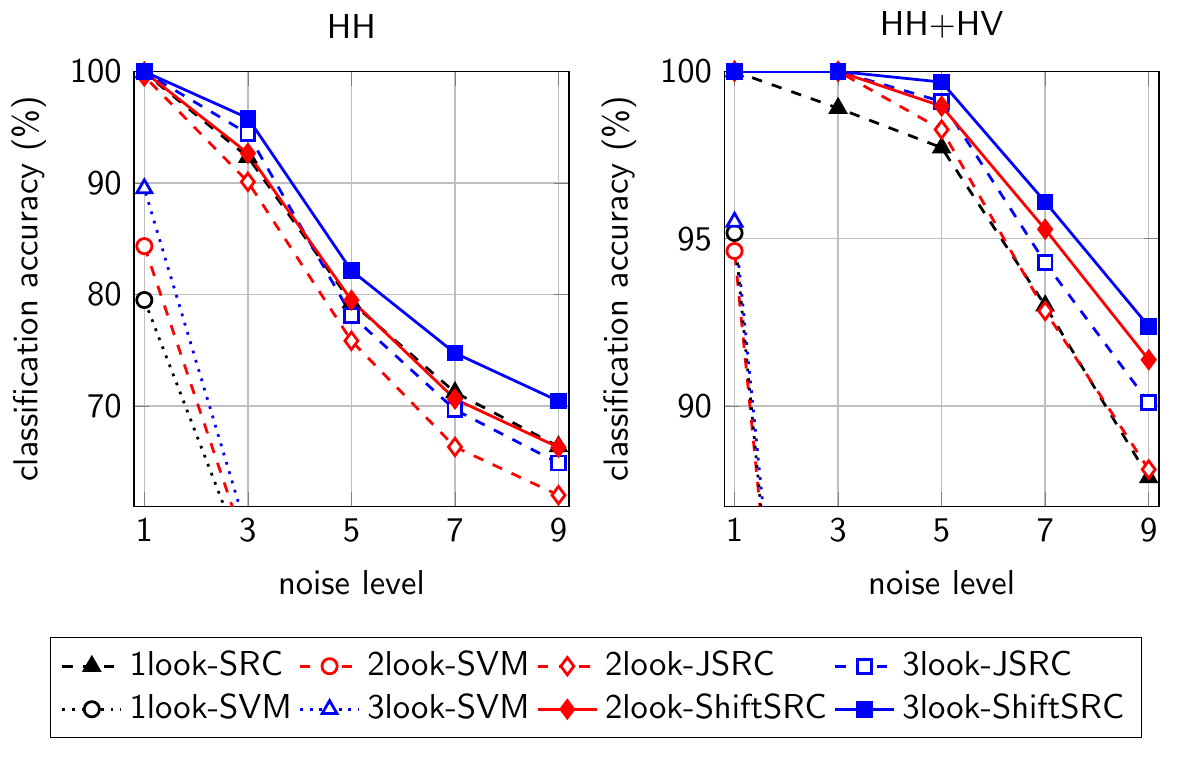}
    \caption{\small Classification accuracy (\%) of different methods on
    multiple-relative-look classification problem.}
    \label{fig9}
\end{figure}

For this experiment, we consider the \textit{all-target} scenario and two
polarization combinations, HH and HH+HV. It is worth noting that for the
HH+HV case, there will be four channels in \textit{2look} and six channels in
\textit{3look}. The results of different methods are shown in
Figure~\ref{fig9}. We can see that SVM still performs well at
the lowest noise level (1), but drastically degrades with a little more
corruption. On the other hand, SRC-based methods obtain good results
even if the noise level is large for the HH+HV combination. Of sparse representation
methods, ShiftSRC outperforms the others with the gap becoming larger for highly
corrupted signals. Interestingly, ShiftSRC at \textit{2look} provides even
better results than JSRC does at \textit{3look}. These results confirm the
advantages of the proposed relative-look sparse representation model.

\begin{figure}
    \centering
    \includegraphics[width = .69\textwidth]{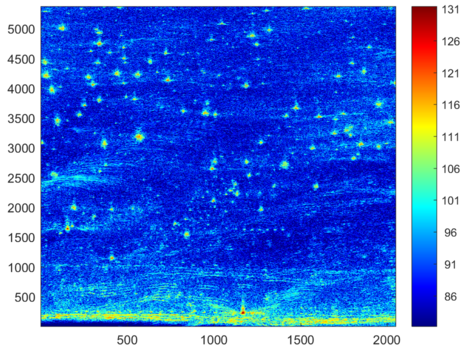}
    \caption{A VV-polarized SAR image of a minefield collected at Yuma Proving Grounds using the Army Research Laboratory UWB radar.}
    \label{fig10}
\end{figure}

\begin{figure}
    \centering
    \includegraphics[width = .47\textwidth]{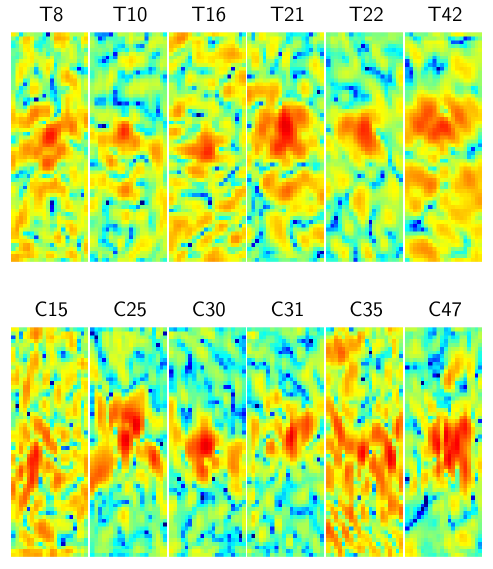}
    \caption{Visualization of real UWB SAR signals. Top: signals of targets,
    bottom: signals of confusers.}
    \label{fig11}
\end{figure}

\subsection{Overall accuracy on measured UWB SAR data} 
\label{sub:overall_accuracy_on_real_data}
In this section, the results of this technique are illustrated using the data
from the ARL UWB low-frequency SAR, which transmits
radar signals occupying the frequency spectrum that span approximately from 50
to 1150 MHz~\cite{ressler1995army}. Figure~\ref{fig10} shows a SAR image formed using data collected at
Yuma Proving Grounds (YPG)~\cite{lam1998}. The scene includes several rows of buried
mines surrounded by clutter objects such as bushes, rocks, tire tracks, etc.


The set contains signals of 50 targets and 50 confusers. Each signal has
resolution of $90\times 30$ and already includes noise from the ground.
Visualization of six samples in each class are shown in
Figure~\ref{fig11}. We conduct the \textit{all-target} experiment
and report results of different methods on different polarization combinations
in Figure~\ref{fig12}. For each combination, three competing methods
are considered: SVM, SRC, and TensorDL (both with the SM constraint). Since
grounds are fixed in this data set, we report the results based on size of the
training set. For each training size $N$ ($N = $10, 20, 30, or 40), we randomly
choose $N$ samples from each class for training; the rest $50 - N$ samples are
considered test samples.

The results are reported in Figure~\ref{fig12} as the average of 10
tests. In general, the tensor dictionary learning performs better than sparse
representation in all combinations except for the HH case. SVM also provides good
results but is outperformed by other {competing} methods.


\begin{figure}[t]
    \centering
    \includegraphics[width = .69\textwidth]{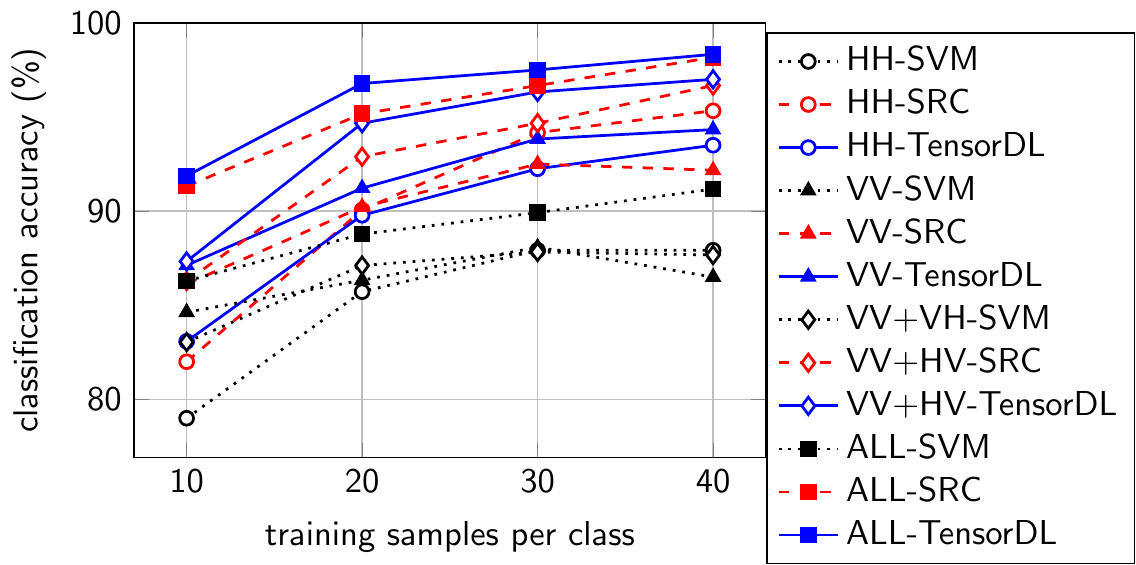}
    \caption{\small Classification accuracy on real data.}
    \label{fig12}
\end{figure}





\section{Conclusion} 

We have developed a novel discrimination and classification framework for 
low-frequency UWB SAR imagery using sparse representation-based methods. The
framework is applicable for either single channel or multiple channel
(polarizations, multilook) SAR imagery. The techniques are tested and the
discrimination/classification performance of targets of interest versus natural
and manmade clutter in challenging scenarios {is} measured using both rigorous
electromagnetic simulation data and real data from the ARL UWB radar. The
classification results show encouraging potential of tensor sparsity methods,
even when the test images are very noisy (buried under extremely rough ground
surfaces), targets have small RCSs, and the targets' responses have very little
detailed structures, i.e., targets are small compared to the wavelengths of the
radar signals. The SRC-SM technique and its dictionary learning version
consistently offers the best results with the combination of co- and cross-pol
data, e.g., HH+HV or ALL. In addition, the non-negativity constraints on sparse
tensor codes enhance the performance of the systems. Furthermore, we also show
the ShiftSRC model is particularly suitable for problems with
multiple-relative-look signals.
\chapter{Conclusions and Future Directions}
\section{Summary of Main Contributions}
The overarching theme in this dissertation is the design of \textit{signal and
image classification} algorithms by exploiting \textit{structurally meaningful
prior information of signal representations in associated models}. Different discriminative models have been explored based on signal sparsity structures. We have primarily considered low-training classification scenarios where the different signal representations exhibit discriminative structure that can be leveraged for robustness benefits. 

In Chapter 2, we focus on extracting discriminative features of
histopathological images that can be useful for classification and detection
tasks. In particular, we based our contributions on dictionary learning to
discriminative learning of complicated medical image features. By simply build
one dictionary for each image class that promote small intra-class
representation and prevent inter-class representation, we obtain class-specific
dictionaries that explicitly  capture crucial features. The framework is
theoretically and experimentally shown to have a low complexity compared to
other related methods. In order to verify that the framework is indeed
applicable in a variety of scenarios, we conduct several experiments on three
diverse histopathological datasets. It is illustrated our method is competitive
with or outperforms state of the art alternatives, particularly in the regime of
realistic or limited training set size. It is also shown that with minimal
parameter tuning and algorithmic changes, the proposed method can be easily
applied on different problems with different natures which makes it a good
candidate for automated medical diagnosis instead of using customized and
problem specific frameworks for every single diagnosis task. We also create a
software toolbox available to help deploy the method widely as a diagnostic tool
in existing histopathological image analysis systems. Particular problems such
as grading and detecting specific regions in histopathology may be investigated
using our proposed techniques.

Chapter 3, we propose a generalized dictionary learning method for different
image classification problem. Our primary contribution in this chapter is the
development of a discriminative dictionary learning framework via the
introduction of a shared dictionary with two crucial constraints. First, the
shared dictionary is constrained to be low-rank. Second, the sparse coefficients
corresponding to the shared dictionary obey a similarity constraint. In
conjunction with a widely used discriminative model, this leads to a more
flexible model where shared features are excluded before doing classification.
An important benefit of this model is the robustness of the framework to size
and the regularization parameter of the shared dictionary. In comparison with
state-of-the-art algorithms developed specifically for these tasks, the proposed
model offers better classification performance on average. Another important
contribution of this chapter is the dictionary learning toolbox -- DICTOL. The
toolbox implements many widely used generative and discriminative dictionary
learning methods. Many classical methods are sped up via new numerical
optimization innovations.

 \def\cm{\Checkmark}

\renewcommand{\arraystretch}{2}
\begin{table}[]
\caption{Comparison of different discriminative sparsity frameworks (more ticks represent better performance).}
\label{tab:comdl}
\begin{tabular}{|l|c|c|c|c|c|}
\hline
                          & SRC\cite{Wright2009SRC} & DLSI\cite{ramirez2010classification} & FDDL\cite{yang2014sparse} & COPAR\cite{kong2012dictionary} & LRSDL \\ \hline
training speed            & no  & \cm & - & \cm  & \cm\cm  \\ \hline
shared features & -  & -   & -   & \cm   & \cm   \\ \hline
low-rank constraint       & -  & -   & -   & -    & \cm   \\ \hline
structured coefficients   & -  & -   & \cm  & \cm   & \cm   \\ \hline
structured dictionaries   & -  & \cm  & \cm  & \cm   & \cm   \\ \hline
classification performance & - & - & \cm & \cm & \cm\cm \\ \hline
\end{tabular}
\end{table}

Table~\ref{tab:comdl} summarizes different characteristics of  widely used
dictionary learning methods and SRC. The proposed LRSDL requires a smallest
amount of training time compared with other frameworks using their
\textit{original} algorithms. In terms of classification performance, two
methods focusing on extracting shared features, COPAR~\cite{kong2012dictionary}
and LRSDL, outperform other methods with slightly better but much faster results
provided by LRSDL, thanks to the efficient algorithm and low-rank structure.

As an extension of discriminative sparsity models, we continue to propose
different tensor sparsity structures for the problem of discriminating UWB-SAR
imagery in Chapter 5. In this chapter, we present three novel sparsity-driven
techniques, which not only exploit the subtle features of raw captured data but
also take advantage of the polarization diversity and the aspect angle
dependence information from multi-channel SAR data. First, the traditional
sparse representation-based classification is generalized to exploit
shared information of classes and various sparsity structures of tensor
coefficients for multichannel data. Corresponding tensor dictionary learning
models are consequently proposed to enhance classification accuracy. Lastly, a
new tensor sparsity model is proposed to model responses from multiple
consecutive looks of objects, which is a unique characteristic of the dataset we
consider. Extensive experimental results on a high-fidelity electromagnetic
simulated dataset and radar data collected from the U.S. Army Research
Laboratory side-looking SAR demonstrate the advantages of proposed tensor
sparsity models.

\section{Potential Future Research Directions}
The contributions in the previous chapters naturally point towards various
directions for future research. We mention some of the possible extensions in
this section.

\subsection{Hierarchically shared feature learning}

    As proposed, the LRSDL model
    learns a dictionary shared by every class. In some practical problems, a
    feature may belong to more than one but not all classes. In future work, we
    will investigate the design of hierarchical models for extracting common
    features among classes.

\subsection{Exploiting shared features using deep learning models} 

    Deep learning has become a powerful framework for many image processing and computer
    vision applications such as image classification and object detection. To
    the best of our knowledge, there is no deep learning model that can reduce
    the effect of the shared features. As viable future research direction for
    this line of work, we propose to identify meaningful physical prior
    information for use in deep networks for extracting shared features and to
    demonstrate its benefits, especially in low training data scenarios.

\subsection{Evolution of structured sparsity}
    Almost all dictionary learning methods enforce sparsity using $\ell_1$ norm
    or $\ell_0$ pseudo norm. It has been shown that Bayesian inference can
    effectively capture sparsity of signals by introducing new
    priors~\cite{jenatton2011structured, zou2005regularization}. Amongst those
    priors,  a well-suited sparsity promoting prior is Spike and Slab prior
    which is widely used in Bayesian inference~\cite{ishwaran2005spike,
    andersen2014bayesian, titsias2011spike}. In fact, it is acknowledged that
    Spike and Slab prior is indeed the gold standard for inducing sparsity in
    Bayesian inference~\cite{titsias2011spike}. Naturally, we can incorporate
    this prior into a dictionary learning framework to further boost the
    classification performance. The main obstacle of using this prior is that
    coefficient update optimization problem is known to be a hard non-convex
    problem where most of existing solutions involve simplifying assumptions
    and/or relaxation. Fortunately, this hard problem has been addressed by two
    very recent works, ICR~\cite{mousavi2015iterative} and AMP~\cite{vu2016amp}.
    Both ICR and AMP propose algorithms to directly solve the problem while AMP
    can solve the problem very effectively. These promising results can be
    extended into a Bayesian dictionary learning framework to further enhance the performance.    

\renewcommand{\arraystretch}{1}

%
\appendix

\def\bA{\mathbf{A}}
\def\bD{\mathbf{D}}
\def\bX{\mathbf{X}}
\def\bM{\mathbf{M}}
\def\bY{\mathbf{Y}}
\def\bx{\mathbf{x}}
\def\by{\mathbf{y}}
\def\bd{\mathbf{d}}
\def\bm{\mathbf{m}}

\Appendix{Complexity Analysis of Dictionary Learning methods}
\label{Sec:appendixB}

{In this section, we compare the computational complexity for the proposed DFDL and competing dictionary learning methods: LC-KSVD\cite{Zhuolin2013LCKSVD}, FDDL\cite{Meng2011FDDL}, and Nayak's\cite{Nandita2013}. The complexity for each dictionary learning method is estimated as the (approximate) number of operations required by each method in learning the dictionary.
For simplicity, we assume that number of training samples, number of dictionary bases in each class are the same, which means: $N_i = N_j = N, k_i = k_j = k, \forall i,j = 1, 2, \dots, c$, and also $L_i = L_j = L, \forall i,j = 1, 2, \dots, c.$ For the consitence, we have changed notations in those methods by denoting $\bY$ as training samples and $\bX$ as the sparse code.
}
\par

{In most of dictionary learning methods, the complexity of sparse coding step, which is often a $l_0$ or $l_1$ minimization problem, dominates that of dictionary update step, which is typically solved by either block coordinate descent\cite{mairal2010online} or singular value decomposition\cite{Aharon2006KSVD}. Then, in order to compare the complexity of different dictionary learning methods, we focus on comparing the complexity of sparse coding steps in each iteration.} 
\section{{Complexity of the DFDL}} 
\label{sec:complexity_of_the_ompcite_tropp2007signal}
{The most expensive computation in DFDL is solving an Orthogonal Matching Pursuit (OMP \cite{tropp2007signal}) problem. Given a set of samples $\bY \in \R^{d \times N}$, a dictionary $\bD \in \R^{d\times k} $ and sparsity level $L$, the OMP problem is:
\begin{equation*}
    \bX^* = \arg\min_{\|\bX\|_0 \leq L} \|\bY - \bD\bX\|_F^2.
\end{equation*}
 R. Rubinstein \etal\cite{Rubinstein2008} reported the complexity of Batch-OMP when the dictionary is stored in memory in its entirety as:
     $T_{\text{b-omp}} = N(2dk + L^2k + 3Lk + L^3) + dk^2.$
Assuming an asymptotic behavior of $L \ll k \approx d \ll N $, the above expression can be simplified to:
\begin{equation}
    T_{\text{b-omp}} \approx N(2dk + L^2k) = kN(2d + L^2).
\end{equation}
 This result will also be utilized in analyzing complexity of LC-KSVD.}\par
\par
{The sparse coding step in our DFDL consists of solving $c$ sparse coding problems:
    $\hat{\bX} = \arg\min_{\|\bX\|_0 \leq L} \norm{\hat{\bY} - \bD_i\hat{\bX}_i}_F^2.$
With $\hat{\bY} \in \R^{d\times cN}, \bD_i \in \R^{d\times k}$, each problem has complexity of $k(cN)(2d + L^2)$. Then the total complexity of these $c$ problems is:
    $T_{\text{DFDL}} \approx c^2kN(2d + L^2)$.
}\section{{Complexity of LC-KSVD}} 
\label{sec:complexity_of_lc_ksvdcite_zhuolin2013lcksvd}
{We consider LC-KSVD1 only (LC-KSVD2 has a higher complexity) whose optimization problem is written as \cite{Zhuolin2013LCKSVD}:
\begin{equation*}
    (\bD, \bA, \bX) = \arg\min_{\bD, \bA, \bX}\norm{\bY -\bD\bX}_F^2 + \alpha\norm{\bQ - \bA\bX}_F^2 \text{~s.t.~} \norm{\bx_i}_0 \leq L.
\end{equation*}
and it is rewritten in the K-SVD form:
\begin{equation}
  \label{eqn:lcksvd}
    (\bD, \bA, \bX) = \arg\min_{\bD, \bA, \bX}\norm{\bmt \bY \\ \sqrt{\alpha}\bQ \emt  -\bmt \bD \\ \sqrt{\alpha}\bA \emt\bX}_F^2 \text{~s.t.~} \norm{\bx_i}_0 \leq L.
\end{equation}
Since $\bQ \in \R^{ck \times cN}$ and $\bA \in \R^{ck \times ck}$, $\tilde{\bY} = \bmt \bY \\ \sqrt{\alpha}\bQ \emt \in \R^{(d + ck) \times cN}$ and $\tilde{\bD} = \bmt \bD \\ \sqrt{\alpha}\bA \emt \in \R^{(d + ck)\times ck} $. Omitting the computation of scalar multiplications, the complexity of (\ref{eqn:lcksvd}) is:
\begin{equation*}
    T_{\text{LC-KSVD}} \approx (ck)(cN)(2(d+ck)+L^2) = c^2kN(2d + 2ck + L^2).
\end{equation*}
}

\section{{Complexity of Nayak's}}    
\label{sec:complexity_of_nayak_scite_nandita2013}
{The optimization problem in Nayak's\cite{Nandita2013} is:
\begin{equation*}
    (\bD, \bX, \bW) = \arg\min_{\bD, \bX, \bW}\norm{\bY - \bD\bX}_F^2 + \lambda\norm{\bX}_1 + \norm{\bX - \bW\bY}_F^2.
\end{equation*}
$\bX$ is estimated via the gradient descent method that is an iterative method whose main computational task in each iteration is to calculate the gradient of $Q(\bX) = \norm{\bY - \bD\bX}_F^2 + \norm{\bX - \bW\bY}_F^2$ with respect to $\bX$. We have:
\begin{equation*}
  \label{eqn:gradnayak}
    \frac{\partial Q(\bX)}{\partial \bX} = 2\Big((\bD^T\bD + \mathbf{I})\bX - (\bD^T -\bW)\bY \Big).
\end{equation*}
where $\bD^T\bD + \mathbf{I}$, and $(\bD^T - \bW)\bY$ could be precomputed and at each step, only $(\bD^T\bD + \mathbf{I})\bX$ need to be recalculated after $\bX$ is updated.
With $\bD \in \R^{d\times ck}, \bX \in \R^{ck \times cN}, \bY \in \R^{d \times cN}, \bW \in \R^{ck \times d}$, the complexity of the sparse coding step can be estimated as:
\begin{eqnarray}
\label{eqn:compnayak}
     T_{\text{Nayak's}} &\approx& (ck)d(ck) + 2(ck)d(cN) + 2q(ck)^2cN, \\
     &= &c^2kN(2d + 2qck) + c^2dk^2.
\end{eqnarray}
with $q$ being the average number of iterations needed for convergence.
Here we have ignored matrix subtractions, additions and scalar multiplications and focused on matrix multiplications only. We have also used the approximation that complexity of $\bA\bB$ is $2mnp$ where $\bA \in \R^{m\times n}, \bB \in \R^{n \times p}$. The first term in (\ref{eqn:compnayak}) is of $\bD^T\bD + \mathbf{I}$ (note that this matrix is symmetric, then it needs only half of regular operations), the second term is of $(\bD^T -\bW)\bY$ and the last one comes from $q$ times complexity of calculating $(\bD^T\bD + \mathbf{I})\bX$.
}
\section{{Complexity of FDDL}} 
\label{sec:complexity_of_fddlcite_meng2011fddl}
{The sparse coding step in FDDL\cite{Meng2011FDDL} requires solving $c$ class-specific problems:
\begin{eqnarray*}
\label{eqn:findSFDDL}
    \bX_i = \arg\min_{\bX_i} \Big\{ \norm{\bY_i - \bD\bX_i}_F^2 + \norm{\bY_i - \bD_i\bX_i^i}_F^2 + \sum_{j = 1, j \neq i}^c\|\bD_j\bX_i^j\|_F^2 \\
    + \lambda_2\big\{\norm{\bX_i - \bM_i}_F^2 - \sum_{k=1}^c\|\bM_k - \bM\|_F^2 + \eta\|\bX_i\|_F^2 \big\}  + \lambda_1\norm{\bX_i}_1\Big\},
\end{eqnarray*}
with $\bD = [\bD_1, \dots, \bD_c], \bX_i^T = [(\bX_i^1)^T, \dots, (\bX_i^c)^T]$, and $\bM_k = [\bm_k, \dots, \bm_k] \in \R^{ck \times N}, \bM = [\bm, \dots, \bm] \in \R^{ck \times N}$ where $\bm_k$ and $\bm$ are the mean vector of $\bX_i$ and $\bX = [\bX_1, \dots, \bX_c]$ respectively. The algorithm for solving this problem uses Iterative Projective Method\cite{rosasco2009iterative} whose complexity depends on computing gradient of six Frobineous-involved terms in the above optimization problem at each iteration.}
\\
\\ {For the first three terms, the gradient could be computed as:
  \begin{equation}
      \label{eqn:fddl_123}
      2(\bD^T\bD)\bX_i - 2\bD^T\bY_i + \bmt
      2(\bD_1^T\bD_1)\bX_i^1 \\
      \vdots \\
      2(\bD_i^T\bD_i)\bX_i^i - \bD_i^T\bY_i \\
      \vdots \\
      2(\bD_c^T\bD_c)\bX_i^c
       \emt,
  \end{equation}
  where $\bD^T\bD, \and \bD^T\bY_i$ could be precomputed with the total cost of $(ck)d(ck) + 2(ck)dN = cdk(2N + ck)$; $\bD_i^T\bD_i, \and \bD_i^T\bY_i $ could be extracted from $\bD^T\bD, \and \bD^T\bY_i$ at no cost; at each iteration, cost of computing $(\bD^T\bD)\bX_i$ is $2(ck)^2N$, each of $(\bD_j^T\bD_j)\bX_i^j$ could be attained in the intermediate step of computing $(\bD^T\bD)\bX_i$. Therefore, with $q$ iterations, the computational cost of (\ref{eqn:findSFDDL}) is:
  \begin{equation}
  \label{eqn:cost123}
    cdk(2N + ck) + 2qc^2k^2N.         
  \end{equation}}
For the last three terms, we will prove that:
\begin{eqnarray}
\label{eqn:dev1}
  \frac{\partial}{\partial \bX_i}\|\bX_i - \bM_i\|_F^2 &=& 2(\bX_i - \bM_i),  \\
\label{eqn:dev2}
  \frac{\partial}{\partial \bX_i} \sum_{k=1}^c \|\bM_k - \bM\|_F^2 &=& 2(\bM_i - \bM),  \\
\label{eqn:dev3}
  \frac{\partial}{\partial \bX_i} \eta \|\bX_i\|_F^2 &=& 2\eta \bX_i.
\end{eqnarray}
Indeed, let $\bE_{m,n}$ be a all-one matrix in $\R^{m\times n}$, one could easily verify that:
\begin{equation*} 
    \bM_k = \frac{1}{N} \bX_k \bE_{N,N}; \quad \bM = \frac{1}{cN}\bX \bE_{cN,N} = \frac{1}{cN}\sum_{i=1}^c \bX_i \bE_{N,N};
\end{equation*}
\begin{equation*} 
    \bE_{m,n}\bE_{n,p} = n\bE_{m,p}; \quad (\mathbf{I} - \frac{1}{N} \bE_{N,N}) (\mathbf{I} - \frac{1}{N} \bE_{N,N}) ^T = (\mathbf{I} - \frac{1}{N} \bE_{N,N}).
\end{equation*}
Thus, (\ref{eqn:dev1}) can be obtained by:
\begin{multline*}
  \frac{\partial}{\partial \bX_i}\|\bX_i - \bM_i\|_F^2 = \frac{\partial}{\partial \bX_i}\|\bX_i - \frac{1}{N} \bX_i\bE_{N,N}\|_F^2  = \frac{\partial}{\partial \bX_i} \|\bX_i(\mathbf{I} - \frac{1}{N} \bE_{N,N})\|_F^2  \\= 2\bX_i(\mathbf{I} - \frac{1}{N} \bE_{N,N}) (\mathbf{I} - \frac{1}{N} \bE_{N,N}) ^T =  2\bX_i(\mathbf{I} - \frac{1}{N} \bE_{N,N}) = 2(\bX_i - \bM_i).
\end{multline*}
{For (\ref{eqn:dev2}), with simple algebra, we can prove that:
\begin{equation*}
    \frac{\partial}{\partial \bX_i} \|\bM_i - \bM\|_F^2 = \frac{2(c-1)}{cN}(\bM_i - \bM)\bE_{N,N} = \frac{2(c-1)}{c}(\bM_i - \bM).
\end{equation*}
\begin{equation*}
     \frac{\partial}{\partial \bX_i} \|\bM_k - \bM\|_F^2 = \frac{2}{cN}(\bM - \bM_k)\bE_{N,N} = \frac{2}{c} (\bM - \bM_k), (k \neq i).
\end{equation*}}
Compared to (\ref{eqn:fddl_123}), calculating (\ref{eqn:dev1}), (\ref{eqn:dev2}) and (\ref{eqn:dev3}) require much less computation. As a result, the total cost of solving $\bX_i$ approximately equals to (\ref{eqn:cost123}); and the total estimated cost of sparse coding step of FDDL is estimated as $c$ times cost of each class-specific problem and approximately equals to:
\begin{equation*}
T_{\text{FDDL}} \approx c^2dk(2N + ck) + 2qc^3k^2N = c^2kN(2d + 2qck) + c^3dk^2.
\end{equation*}
Final analyzed results of these four methods are reported in Table \ref{tab:complexity}.

\Appendix{Proof of Lemmas in Chapter~\ref{chapter:contrib_lrsdl}}
\section{Proof of Lemma \ref{lem:fddl_updateD}}
\label{apd:proof_fddl_updateD}

\def\bD{\mathbf{D}}
\def\bX{\mathbf{X}}
\def\bM{\mathbf{M}}
\def\bY{\mathbf{Y}}
\def\bx{\mathbf{x}}
\def\by{\mathbf{y}}
\def\bd{\mathbf{d}}
\def\bm{\mathbf{m}}

\def\M{\mathcal{M}}
\def\bG{\mathbf{G}}

\def\bbX{\lbar{\bX}}        
\def\bbx{\lbar{\bx}}        
\def\bbY{\lbar{\bY}}        
\def\bbD{\lbar{\bD}}

 Let $\bw_c \in \{0, 1\}^{K}$ is a binary vector whose $j$-th element is one if and only if the $j$-th columns of $\bD$ belong to $\bD_c$, and $\bW_c = \diag(\bw_c)$. We observe that $\bD_c\bX^c_i = \bD\bW_c\bX_i$. We can rewrite $f_{\bY, \bX}(\bD)$ as:
\small{ \begin{eqnarray*}
\nonumber
    &&\|\bY - \bD\bX\|_F^2 + \sum_{c=1}^C\big(\|\bY_c - \bD_c\bX^c_c\|_F^2 + \sum_{j\neq c} \|\bD_j\bX^j_c\|_F^2\big) \\
\nonumber
    &=& \|\bY - \bD\bX\|_F^2 + \sum_{c=1}^C\big(\|\bY_c - \bD\bW_c\bX_c\|_F^2 + \sum_{j\neq c} \|\bD\bW_j\bX_c\|_F^2\big) \\
    &=& \trace\left(\big( \bX\bX^T + \sum_{c=1}^C\sum_{j=1}^C\bW_j\bX_c\bX_c^T\bW_j^T\ \big)\bD^T\bD\right), \\
    && -2\trace\left(\big(\bY\bX^T + \sum_{c=1}^C \bY_c\bX_c^T\bW_c \big)\bD^T\right)+ \text{constant},\\
    &=& -2\trace(\bE\bD^T) + \trace(\Fb\bD^T\bD) + \text{constant}.
\end{eqnarray*}}
{ where we have defined:
\begin{eqnarray*}
    \bE &=& \bY\bX^T + \sum_{c=1}^C \bY_c\bX_c^T\bW_c =  \bY\bX^T + \bmt \bY_1(\bX_1^1)^T &\dots & \bY_C(\bX_C^C)^T\emt,  \\
    &=& \bY \left(\bX^T + \bmt
        (\bX_1^1)^T & \dots & \bzeros\\
        \bzeros  & \dots & \bzeros\\
        \dots    & \dots & \dots\\
        \bzeros  & \dots & (\bX_C^C)^T

        \emt\right) = \bY\M(\bX)^T,\\
    \Fb &=& \bX\bX^T + \sum_{c=1}^C\sum_{j=1}^C\bW_j\bX_c\bX_c^T\bW_j^T
    = \bX\bX^T + \sum_{j=1}^C \bW_j \left(\sum_{c=1}^C\bX_c\bX_c^T\right) \bW_j^T,\\
    &=& \bX\bX^T + \sum_{j=1}^C \bW_j\bX\bX^T\bW_j^T.
\end{eqnarray*}
Let:
\begin{equation*}
\bX\bX^T = \bA =
\bmt
        \bA_{11} &  \dots & \bA_{1j} & \dots & \bA_{1C}\\
        \dots    &  \dots & \dots    & \dots & \dots \\
        \bA_{21} &  \dots & \bA_{jj} & \dots & \bA_{2C}\\
        \dots    &  \dots & \dots    & \dots & \dots \\
        \bA_{C1} &  \dots & \bA_{Cj} & \dots & \bA_{CC}
        \emt.
\end{equation*}
From definition of $\bW_j$, we observe that `left-multiplying' a matrix by $\bW_j$ forces that matrix to be zero everywhere except the $j$-th block row. Similarly, `right-multiplying' a matrix by $\bW_j^T = \bW_j$ will keep its $j$-th block column only. Combining these two observations, we can obtain the result:
\begin{equation*}
    \bW_j\bA\bW_j^T =
    \bmt
        \bzeros &  \dots & \bzeros & \dots & \bzeros\\
        \dots    &  \dots & \dots    & \dots & \dots \\
        \bzeros &  \dots & \bA_{jj} & \dots & \bzeros\\
        \dots    &  \dots & \dots    & \dots & \dots \\
        \bzeros &  \dots & \bzeros & \dots & \bzeros
        \emt.
\end{equation*}
Then:
\begin{eqnarray*}
   \Fb &=&  \bX\bX^T + \sum_{j=1}^C \bW_j\bX\bX^T\bW_j^T, \\
   &=& \bA +  \bmt
        \bA_{11} & \dots & \bzeros\\
        \bzeros  & \dots & \bzeros\\
        \dots    & \dots & \dots\\
        \bzeros  & \dots & \bA_{CC}
        \emt = \M(\bA) = \M(\bX\bX^T).
\end{eqnarray*}

Lemma \ref{lem:fddl_updateD} has been proved. \hfill $\square$

\section{Proof of Lemma \ref{lem:fddl_updateX}}
\label{apd:proof_fddl_updateX}
We need to prove two parts:\\
\textit{For the gradient of} $f$, first we rewrite:
\begin{eqnarray}
\hspace{-0.1in}
\nonumber
  f(\bY, \bD, \bX) = \sum_{c=1}^C r(\bY_c, \bD, \bX_c) &=&
   \norm{\underbrace{\bmt
    \bY_1 & \bY_2  & \dots & \bY_C \\
    \bY_1 & \mathbf{0} & \dots & \mathbf{0} \\
    \mathbf{0} & \bY_2 & \dots & \mathbf{0} \\
    \dots & \dots & \dots & \dots \\
    \mathbf{0} & \mathbf{0} & \dots & \bY_C
    \emt}_{\widehat{\bY}}
   - \underbrace{\bmt
     \bD_1 & \bD_2  & \dots & \bD_C \\
     \bD_1 & \mathbf{0} & \dots & \mathbf{0} \\
     \mathbf{0} & \bD_2 & \dots & \mathbf{0} \\
     \dots & \dots & \dots & \dots \\
     \mathbf{0} & \mathbf{0} & \dots & \bD_C
     \emt}_{\widehat{\bD}}
  \bX
  }_F^2\\
  &=& \| \widehat{{\bY}}  - \widehat{\bD} \bX \|_F^2.
\end{eqnarray}
Then we obtain:}
{ \begin{eqnarray*}
\def\M{\mathcal{M}}
     \frac{\partial \frac{1}{2}f_{\bY, \bD}(\bX)}{\partial \bX}
     = \widehat{\bD}^T\widehat{\bD} - \widehat{\bD}^T\widehat{\bY} = \M(\bD^T\bD) \bX - \M(\bD^T\bY).
\end{eqnarray*}
\textit{For the gradient of} $g$, let $\bE_{p}^q$ be the all-one matrix in $\R^{p\times q}$. We can verify that:}
 \begin{eqnarray*}
    (\bE_p^q)^T = \bE_q^p,
    & &
    \bM_c = \bm_c\bE_{1}^{n_c} = \frac{1}{n_c}\bX_c\bE_{n_c}^{n_c},
    \\
    \bE_p^q\bE_q^r = q\bE_p^r,
    &&
    (\bI-\frac{1}{p}\bE_p^p)(\bI - \frac{1}{p}\bE_p^p)^T = (\bI - \frac{1}{p}\bE_p^p).
\end{eqnarray*}
 We have:\\
    \begin{equation*}
             \bX_c - \bM_c = \bX_c - \frac{1}{n_c}\bX_c\bE_{n_c}^{n_c} = \bX_c(\bI - \frac{1}{n_c}\bE_{n_c}^{n_c}),
    \end{equation*}
    \begin{eqnarray*}
          \imply \frac{\partial}{\partial \bX_c}\frac{1}{2}\|\bX_c - \bM_c\|_F^2 =   \bX_c (\bI - \frac{1}{n_c}\bE_{n_c}^{n_c})(\bI - \frac{1}{n_c}\bE_{n_c}^{n_c})^T
                                                                                = \bX_c (\bI - \frac{1}{n_c}\bE_{n_c}^{n_c}) = \bX_c - \bM_c.
      \end{eqnarray*}
Therefore we obtain:
    \begin{align}
        \label{eqn:dif_g_1}
        \frac{\partial \frac{1}{2}\sum_{c=1}^C \|\bX_c - \bM_c\|_F^2}{\partial \bX} &= [ \bX_1, \dots, \bX_C] - \underbrace{[ \bM_1, \dots, \bM_C]}_{\widehat{\bM}} 
        = \bX - \widehat{\bM}.
    \end{align}
    For $\bM_c - \bM$, first we write it in two ways:
{   \begin{align}
        \nonumber
        &\bM_c - \bM = \frac{1}{n_c} \bX_c \bE_{n_c}^{n_c} - \frac{1}{N}\bX\bE_{N}^{n_c}
         =  \frac{1}{n_c} \bX_c \bE_{n_c}^{n_c} - \frac{1}{N}\sum_{j=1}^C\bX_j\bE_{n_j}^{n_c}, \\
        \label{eqn:mcm_xc}
        & = \frac{N - n_c}{Nn_c}\bX_c \bE_{n_c}^{n_c} - \frac{1}{N}\sum_{j \neq c}\bX_j\bE_{n_j}^{n_c},\\
        \label{eqn:mcm_xj}
        & = \frac{1}{n_c} \bX_c \bE_{n_c}^{n_c} - \frac{1}{N}\bX_l\bE_{n_l}^{n_c} - \frac{1}{N}\sum_{j\neq l}\bX_j\bE_{n_j}^{n_c}  ~~ (l\neq c).
    \end{align}}
    Then we infer:
    { \begin{align*}
    \displaystyle
        \nonumber
            (\ref{eqn:mcm_xc}) \imply  \frac{\partial}{\partial \bX_c} \frac{1}{2} \|\bM_c - \bM\|_F^2 &= \left(\frac{1}{n_c} - \frac{1}{N}\right)(\bM_c - \bM)\bE_{n_c}^{n_c}
                 = (\bM_c - \bM) + \frac{1}{N} (\bM - \bM_c)\bE_{n_c}^{n_c}.\\
           (\ref{eqn:mcm_xj}) \imply  \frac{\partial}{\partial \bX_l} \frac{1}{2} \|\bM_c - \bM\|_F^2 &= {\frac{1}{N}(\bM - \bM_c)\bE_{n_c}^{n_l}} (l \neq c).
       \end{align*}}
        { $\displaystyle\imply\frac{\partial}{\partial \bX_l} \frac{1}{2}\sum_{c=1}^C \|\bM_c - \bM\|_F^2 = \bM_l - \bM + \frac{1}{N}\sum_{c=1}^C (\bM - \bM_c)\bE_{n_c}^{n_l}.$}\\
{    Now we prove that $\displaystyle\sum_{c=1}^C (\bM - \bM_c)\bE_{n_c}^{n_l} = \bzeros$. Indeed,
    \begin{align*}
        \sum_{c=1}^C (\bM - \bM_c)\bE_{n_c}^{n_l}&= \sum_{c=1}^C(\bm\bE_{1}^{n_c} - \bm_c\bE_{1}^{n_c})\bE_{n_c}^{n_l}, \\
        &= \sum_{c=1}^C(\bm - \bm_c)\bE_{1}^{n_c}\bE_{n_c}^{n_l}
        =\sum_{c=1}^Cn_c(\bm - \bm_c)\bE_{1}^{n_l,} \\
        &= \big(\sum_{c=1}^C n_c \bm - \sum_{c=1}^C n_c\bm_c\big)\bE_{1}^{n_l} =\bzeros~\left(\text{since~} \bm = \frac{\sum_{c=1}^C n_c \bm_c}{\sum_{c=1}^C n_c}\right).
    \end{align*}
    Then we obtain:}
 { \small   \begin{align}
        \label{eqn:dif_g_2}
        \frac{\partial}{\partial \bX}\frac{1}{2}\sum_{c=1}^C \|\bM_c - \bM\|_F^2 = [\bM_1, \dots, \bM_C] - \bM = \widehat{\bM} - \bM.
    \end{align}
    Combining (\ref{eqn:dif_g_1}), (\ref{eqn:dif_g_2}) and $\frac{\partial \frac{1}{2}\|\bX\|_F^2}{\partial \bX} = \bX$ , we have:}
{    \begin{equation*}
        \frac{\partial \frac{1}{2}g(\bX)}{\partial \bX} = 2\bX + \bM - 2\widehat{\bM.}
    \end{equation*}
    Lemma \ref{lem:fddl_updateX} has been proved. \hfill $\square$
\section{Proof of Lemma \ref{lem:lrsdld0x0}}
\vspace{-0.1in}
\label{apd:proof_lrsdl_updateD0X0}
When $\bY, \bD, \bX$ are fixed, we have:
\begin{align}
\label{eqn:sddl_solve_X0_ori}
    \nonumber
    &J_{\bY, \bD, \bX}(\bD_0, \bX^0) = \frac{1}{2}\|\bY - \bDc\bX^0 - \bD\bX\|_F^2 + \eta\|\bD_0\|_*+ \\
        &\sum_{c=1}^{C}\frac{1}{2}\|\bY_c - \bDc\bX_c^0 - \bD_c\bX_c^c\|_F^2
        +\lambda_1\|\bX^0\|_1  + \text{constant.}
\end{align}}
Let $\tilde{\bY} = \bY - \bD\bX$, $\hat{\bY}_c = \bY_c - \bD_c \bX_c^c$ and $\hat{\bY} =
    \bmt \hat{\bY}_1 & \hat{\bY}_2 & \dots & \hat{\bY}_C \emt $, we can rewrite (\ref{eqn:sddl_solve_X0_ori}) as:
\begin{eqnarray*}
\nonumber
 J_{\bY, \bD, \bX}(\bD_0, \bX^0) &=&\frac{1}{2}\|\tilde{\bY} - \bDc\bXc\|_F^2 + \frac{1}{2}\|\hat{\bY} - \bDc\bXc \|_F^2 
  +\lambda_1 \|\bXc\|_1 + \eta\|\bD_0\|_* + \text{constant}_1,\\
    &=& \norm{\frac{\tilde{\bY} + \hat{\bY}}{2} - \bDc\bXc}_F^2 + \lambda_1 \|\bXc\|_1 + \eta\|\bD_0\|_* + \text{constant}_2.
\end{eqnarray*}
We observe that:
 \begin{eqnarray*}
    \tilde{\bY} + \hat{\bY} &=& 2\bY - \bD\bX - \bmt \bD_1\bX_1^1 &\dots & \bD_C\bX_C^C\emt \\
    &=& 2\bY - \bD\mathcal{M}(\bX).
\end{eqnarray*}
Now, by letting
$\displaystyle\bV = \frac{\tilde{\bY} + \hat{\bY}}{2} $, Lemma \ref{lem:lrsdld0x0} has been proved. $~~~$ \hfill $\square$

}


   \begin{singlespace}
   \bibliographystyle{IEEEtran}
   \addcontentsline{toc}{chapter}{Bibliography}
   \bibliography{Biblio-Database}
\newpage

   \end{singlespace}

\thispagestyle{empty}
{\small

\textbf{Education:}

Pennsylvania State University, Ph.D. Electrical Engineering, Dec. 2018.
\vspace{-.08in}

Pennsylvania State University, M.Sc. Electrical Engineering, Apr. 2015.
\vspace{-.08in}

Hanoi University of Science and Technology, B.Sc. Electronics and Telecommunications, Aug. 2012.

\textbf{Experience:}

Research Assistant, Pennsylvania State University, Aug. 2013 - Jul. 2018.
\vspace{-.08in}

Research Intern, U.S. Army Research Laboratory, May - Sep 2016, May - Aug 2017. 
\vspace{-.08in}

Teaching Assistant, Hanoi University of Science and Technology, Jan - May 2011. 

\textbf{Personal Website:}

\url{https://www.linkedin.com/in/tiephuuvu/}

\textbf{Selected Publications related to Ph.D research:}

\begin{enumerate}

    \item \textbf{Tiep H. Vu}, L. Nguyen, V. Monga, ``Classifying Multi-channel UWB SAR Imagery via Tensor Sparsity Learning Techniques'', IEEE Transactions on Aerospace and Electronic Systems, 2018.

    \item \textbf{Tiep H. Vu}, L. Nguyen, T. Guo, Vishal Monga, ``Deep Network for Simultaneous Decomposition and Classification in UWB-SAR Imagery'', IEEE Radar Conference, 2018.

    \item \textbf{Tiep H. Vu}, V. Monga. ``Fast Low-rank Shared Dictionary Learning for Image Classification'', IEEE Transactions on Image Processing 26.11 (2017): 5160-5175.

    \item \textbf{Tiep H. Vu}, L. Nguyen, C. Le, V. Monga, ``Tensor Sparsity for Classifying Low-Frequency Ultra-Wideband (UWB) SAR Imagery'', IEEE Radar Conference, 2017.

    \textbf{Tiep H. Vu}, H.S. Mousavi, V. Monga, ``Adaptive matching pursuit for sparse signal recovery'', IEEE on International Conference on Acoustics, Speech, and Signal Processing (ICASSP), 2017. \textbf{Finalist for the Best Student Paper Award}.

    \item \textbf{Tiep H. Vu}, H.S. Mousavi, V. Monga, A. U. Rao and G. Rao, ``Histopathological Image Classification using Discriminative Feature-Oriented Dictionary Learning'', IEEE Transactions on Medical Imaging , volume 35, issue 3, pages 738-751, March 2016.


    \item \textbf{Tiep H. Vu}, Vishal Monga, ``Learning a low-rank shared dictionary for object classification'', IEEE International Conference on Image Processing (ICIP), pp. 4428-4432, 2016.

    \item \textbf{Tiep H. Vu}, H. S. Mousavi, V. Monga, UK A. Rao, G Rao, ``DFDL: Discriminative Feature-Oriented Dictionary Learning for Histopathological Image Classification'', IEEE International Symposium on Biomedical Imaging (ISBI), 2015.
\end{enumerate}
}

\backmatter

\end{document}